\documentclass[acmlarge]{acmart}

\usepackage[utf8]{inputenc} 
\usepackage[T1]{fontenc}    
\usepackage{hyperref}       
\usepackage{url}            
\usepackage{booktabs}
\usepackage{soul}
\usepackage{longtable}
\usepackage{amsfonts}       
\usepackage{nicefrac}       
\usepackage{microtype}      
\usepackage{xcolor}         
\usepackage{multirow}
\usepackage{subfigure}
\usepackage{colortbl}
\usepackage{amsmath}
\usepackage{amsthm}
\usepackage{booktabs}
\usepackage{algorithm}
\usepackage{algorithmic}
\usepackage{enumitem}
\usepackage{makecell}
\usepackage{wrapfig}
\usepackage{subcaption}
\usepackage{graphicx}
\usepackage{adjustbox}
\usepackage{float}
\usepackage{placeins}

\DeclareMathOperator*{\argmin}{arg\,min}
\AtBeginDocument{%
  }

\setcopyright{acmlicensed}
\copyrightyear{2018}
\acmYear{2018}
\acmDOI{XXXXXXX.XXXXXXX}
\acmConference[Conference acronym 'XX]{Make sure to enter the correct
  conference title from your rights confirmation email}{June 03--05,
  2018}{Woodstock, NY}
\acmISBN{978-1-4503-XXXX-X/2018/06}

\begin{document}

\title{CurvFed: Curvature-Aligned Federated Learning for Fairness without Demographics}
\author{Harshit Sharma\textsuperscript{*}, Shaily Roy\textsuperscript{*}, Asif Salekin}
\affiliation{%
  \institution{Arizona State University}
  \city{Tempe}
  \state{AZ}
  \country{USA}
}
\email{{hsharm62, shailyro,  asalekin}@asu.edu}

\thanks{\textsuperscript{*}\textbf{Equal Contribution}}
\renewcommand{\shortauthors}{Sharma et al.}


\begin{abstract}
Modern human-sensing applications often rely on data distributed across users and devices, where privacy concerns prevent centralized training. Federated Learning (FL) addresses this challenge by enabling collaborative model training without exposing raw data or attributes. However, achieving fairness in such settings remains difficult, as most human-sensing datasets lack demographic labels, and FL’s privacy guarantees limit the use of sensitive attributes. This paper introduces \textbf{\textit{CurvFed}}: Curvature-Aligned Federated Learning for Fairness without Demographics, a theoretically grounded framework that promotes fairness in FL without requiring any demographic or sensitive attribute information—a concept termed \textit{Fairness without Demographics (FWD)}—by optimizing the underlying loss-landscape curvature. Building on the theory that equivalent loss-landscape curvature corresponds to consistent model efficacy across sensitive attribute groups, \textbf{\textit{CurvFed}} regularizes the top eigenvalue of the Fisher Information Matrix (FIM) as an efficient proxy for loss-landscape curvature, both within and across clients. This alignment promotes uniform model behavior across diverse bias-inducing factors, offering an attribute-agnostic route to algorithmic fairness. \textbf{\textit{CurvFed}} is especially suitable for real-world human-sensing FL scenarios involving single or multi-user edge devices with unknown or multiple bias factors. We validated \textbf{\textit{CurvFed}} through theoretical and empirical justifications, as well as comprehensive evaluations using three real-world datasets and a deployment on a heterogeneous testbed of resource-constrained devices. Additionally, we conduct sensitivity analyses on local training data volume, client sampling, communication overhead, resource costs, and runtime performance to demonstrate its feasibility for practical FL edge device deployment.
\end{abstract}

\maketitle

\section{Introduction}\label{intro-label}
The integration of advanced machine learning (ML) with ubiquitous sensing \cite{li2021deep,puccinelli2005wireless} has enabled everyday devices like smartphones and wearables to collect rich data for human-centric applications, from gesture recognition \cite{zhang2021WIDAR3} to stress detection \cite{vos2023generalizable}.
Concurrently, edge and distributed computing—especially federated learning (FL)—have emerged as key frameworks for decentralized model training \cite{zhou2019edge, mcmahan2017communication}. 
In FL, each client (e.g., device or organization) retains local data and trains models privately \cite{li2024comprehensive}, reducing centralization risks \cite{arcuri2010centralization} and aligning with privacy regulations like GDPR, CCPA, and HIPAA \cite{zhang2020fairfl, abay2020mitigating}. This makes FL ideal for human-sensing applications like health monitoring and activity recognition \cite{epstein2020mapping, kim2016timeaware, shaik2022fedstack}.

\par 
Notably, recent studies have begun exploring fairness in human-sensing federated learning (FL) \cite{yfantidou2023beyond,xu2023globem,zhang2023framework}.
 However, state-of-the-art FL fairness approaches face a key challenge due to the trade-off between fairness and privacy \cite{chen2023privacy}, often requiring access to sensitive information at the client or server level, which contradicts \emph{FL’s privacy-preserving nature} \cite{zeng2021improving,papadaki2022minimax,djebrouni2024bias,zhang2020fairfl,ezzeldin2023fairfed}. This underscores the need for fairness-enhancing approaches that do not rely on sensitive or bias-inducing attributes. In human-sensing FL, this is especially important, as unlabeled factors can bias models —e.g., building materials can skew Wi-Fi CSI-based activity detection \cite{lee2019effects}.

These challenges motivate the pursuit of Fairness without Demographics (FWD) \cite{lahoti2020fairness} -refers to achieving equitable model performance across users or groups without relying on sensitive or demographic attributes such as sex, age, materials, among others, in FL, particularly since most human-sensing datasets lack labeled sensitive attributes \cite{yfantidou2023beyond,mehrabi2021survey,zhang2023framework}.
\textit{This also aligns with the 2025 Executive Order's \cite{house2025removing} vision} for bias-resistant, innovation-driven AI by promoting fairness through optimization principles—without relying on demographic or group attributes—supporting equitable systems free from engineered social agendas. While FWD has been studied in centralized AI training \cite{lahoti2020fairness,hashimoto2018fairness,chai2022fairness}, and only one FWD in FL work named FedSRCVaR \cite{papadaki2024federated} exists, FWD's integration into \emph{privacy-preserving} decentralized FL remains challenging in human-centric FL, as outlined below:

\begin{enumerate}[nosep]

\item Centralized FWD methods (Section \ref{FWD-rel-work}) depend on access to global data distributions to identify biases and adjust decision boundaries, which is unavailable in \emph{privacy-preserving} FL where clients only access local data \cite{abbas2023context}, making fairness enforcement more difficult.

\item Although FedSRCVaR \cite{papadaki2024federated} does not require sensitive attribute labels in data, it relies on the knowledge of the bounds for sensitive or worst-case group sizes, making it not entirely agnostic to sensitive attribute knowledge (see Section \ref{FWD-rel-work}).

\item Traditional group fairness methods—including FWD approaches (Section \ref{FWD-rel-work})—generally assume that all sensitive groups are represented during optimization \cite{rieke2020future}. However, this assumption is impractical in human-sensing FL, where each client often corresponds to a single user or a narrow demographic segment, resulting in an incomplete representation of population-level groups.


\item \textcolor{black}{In human-centric sensing, multiple latent biases (e.g., gender, body type, sensor placement) coexist \cite{jui2024fairness}, and mitigating one bias may inadvertently worsen another \cite{djebrouni2024bias}.}

\end{enumerate}

In summary, a practical FWD solution for privacy-preserving human-sensing FL must operate without the knowledge of the sensitive attributes, tolerate missing group representations, and ensure fairness across both single- and multi- user clients and multiple bias sources. 
To address these challenges, this paper takes a fundamentally different approach than the literature—examining the \emph{optimization loss-landscape geometry} of FL. It presents a theoretical justification (in Section \ref{theory-section}) that reducing disparities in loss-landscape curvature both across and within clients offers an attribute-agnostic route to algorithmic fairness.

Building on this insight, it introduces \textit{\textbf{Curv}ature-Aligned \textbf{Fed}erated Learning} (\textbf{\textit{CurvFed}}), 
a novel FL method that achieves Fairness Without Demographics (FWD) by promoting \emph{equitable and flatter} loss landscape curvatures across and within clients, without relying on sensitive attributes, bias factors, or global distribution knowledge, thus addressing the above-discussed first two and fourth challenges.

\textbf{\textit{CurvFed}} utilizes the Fisher Information Matrix (FIM) as a fast-computable proxy for loss-landscape sharpness \cite{lee2022masking,thomas_interplay_2020}, making it well-suited for resource-constrained edge devices (as FL clients), where it regularizes locally trained models to enhance fairness. 
Importantly, it introduces a sharpness-aware aggregation scheme, which aligns loss-landscape curvature across all clients by leveraging client-wise FIM data. This addresses the above-discussed third challenge, since FIM can be computed for both single-user and multi-user clients, allowing \textbf{\textit{CurvFed}} to operate without requiring full representation of sensitive attributes within each client. 


Recognizing the discussed challenges, our work introduces the first FWD for human-sensing FL, \textbf{\textit{CurvFed}} that provides a privacy-preserving, robust alternative to traditional FL bias mitigation for human-sensing applications through the following contributions:

\begin{itemize}[nosep]

\item \textit{Theoretically Grounded Novel Approach}: \textbf{\textit{CurvFed}}, being the first of its kind, provides a comprehensive theoretical justification establishing how having flatter curvature within the client through local training and equitable curvatures across clients through a novel sharpness-aware aggregation scheme promotes FWD (Section \ref{theory-section}). Its theoretical claims are further supported by empirical validation (Section \ref{empirical-verification}), making it a rigorous and novel solution.

\item \textit{Superior fairness–accuracy tradeoff}: Existing FL methods often struggle with the inherent tension between fairness and accuracy \cite{ezzeldin2023fairfed, gu2022privacy}. \textbf{\textit{CurvFed}} mitigates this by promoting flatter loss landscapes during client training, which inherently enhances generalizability \cite{dauphin2024neglected, jastrzkebski2018relation} and model inference confidence by pushing samples away from decision boundaries \cite{tran2022pruning}. It further optimizes both curvature smoothness and accuracy during aggregation, yielding a better balance of fairness and performance (see Sections \ref{approach-design-choice} and \ref{eval-results}, RQ-1).

\item \textit{Equitable performance across diverse FL setups:} \textbf{\textit{CurvFed}} handles varying client compositions from single-user to multi-user, without requiring full representation of sensitive groups, ensuring practical and equitable performance in real-world human-sensing FL (see Section \ref{eval-results}, RQ-2, RQ-3).

\item \textit{Improved multi-attribute fairness simultaneously:} Human-sensing data involves multiple sensitive attributes \cite{zhang2020fairfl,ezzeldin2023fairfed}, but fair FL methods struggle to address them simultaneously \cite{zhang2020fairfl,cui2021addressing}. \textbf{\textit{CurvFed}} promotes flatter client loss landscapes, improving fairness across all attributes without relying on attribute labels (see Section \ref{eval-results}, RQ-4).

\item \textit{Real-world feasibility and sensitivity analysis:} We validate \textbf{\textit{CurvFed}}’s practicality through extensive evaluations across data volume, client sampling, communication cost, resource usage, and runtime on diverse platforms. Additionally, a \textit{real-world testbed evaluation} on a heterogeneous testbed of six edge devices confirms its effectiveness under non-simulated conditions (see Sections \ref{Feasibility-Study} and \ref{Sensitivity-Analysis}).



\end{itemize}

We demonstrate the effectiveness of \textbf{\textit{CurvFed}} across \textit{three} real-world human-sensing datasets, each representing different sensing modalities, applications, and sensitive attributes, showing \textbf{\textit{CurvFed}}’s effectiveness in FWD. We adopt off-the-shelf models previously published for these datasets \cite{zhang2021WIDAR3,xiao2024reading,chereshnev2018hugadb} for FL setting. However, \textbf{\textit{CurvFed}} is architecture-agnostic. The framework is not tied to specific model choices and can be readily extended to diverse use cases and architectures.
\textit{The rest of the paper is \underline{organized} as follows}: Section \ref{rel_work} reviews prior work; Section \ref{theory-section} presents the theoretical foundation and assumptions of \textbf{\textit{CurvFed}}; Section \ref{problem_statement} defines the problem, gaps, and challenges. Section \ref{approach} details the proposed method, Section \ref{sec:results} presents in-depth results, Section \ref{empirical-verification} presents empirical verifications of our propositions, and Section \ref{broader-impact-limit} discusses broader impacts and limitations.


\section{Background and Related works}
\label{rel_work}

Fairness has emerged as a critical concern in machine learning (ML) research, with early efforts focusing on pre-processing strategies, such as removing sensitive attributes from training data \cite{dwork2012fairness}. However, such approaches often compromise model performance \cite{zhang2020fairfl}, prompting a shift toward in-processing methods that embed fairness constraints into model optimization \cite{zhang2019faht}. While effective in centralized ML, these methods face new challenges in federated learning (FL), where privacy constraints limit access to sensitive attributes \cite{pessach2022review,djebrouni2024bias}. This section discusses the recent fairness approaches in FL, highlights common evaluation metrics, and motivates the need for Fairness Without Demographics (FWD).


\subsection{Fairness in Federated Learning}\label{rel-fairness-in-FL}
Bias in FL systems can arise in multiple ways \cite{li2023hierarchical,zeng2021improving,djebrouni2024bias,cui2021addressing,zhao2018federated}, but a primary concern is \textbf{performance-based group fairness}, 
which aims to reduce disparities in model performance across demographic groups \cite{castelnovo2022clarification}. This is distinct from \textbf{contribution fairness}, which addresses disparities from client participation or data imbalance (e.g., non-IID data) \cite{li2023hierarchical,zhao2018federated}, 
and is beyond the scope of this work. 
In this study, we align with performance-based group fairness approaches and focus on \textit{ensuring that the global model does not disproportionately underperform for any groups}, a challenge exacerbated by the decentralized nature of data in FL.

Group fairness approaches in FL can be categorized by where the intervention occurs: at the client, at the server, or both.

\noindent\textbf{Client-Side and Hybrid Approaches:} 
Several methods enforce fairness during local training. For example, FedFB applies fair batch sampling on the client side \cite{zeng2021improving}, while others use min-max optimization to minimize worst-case group loss \cite{mohri2019agnostic, du2021fairness, papadaki2022minimax}. However, these approaches typically require access to sensitive attribute labels, making them unsuitable for Fairness Without Demographics (FWD). Notably, AFL \cite{mohri2019agnostic} targets fairness across clients, rather than demographic groups, whereas others assume the availability of knowledge of sensitive group labels and distributions.

Recent works that implicitly flatten the loss landscape offer a closer fit to our setting, as they can promote fairness without accessing sensitive data. For example, \cite{caldarola2022improving} applies Sharpness-Aware Minimization (SAM) at the client and Stochastic Weight Averaging (SWA) at the server. These methods, which do not rely on demographic attributes, serve as relevant baselines for our evaluations. Our approach, \textbf{\textit{CurvFed}}, explicitly leverages sharpness cues and follows a hybrid framework leveraging sharpness information from the loss surface (a manner distinct from SAM) and adapting SWA in one of the modules. Our ablation study in Section \ref{ablation-study} also demonstrates the impact of such adaptation in \textit{\textbf{CurvFed}}.


\noindent\textbf{Server-Side Approaches:} Some strategies address fairness during the server's aggregation phase. FairFed \cite{ezzeldin2023fairfed} adjusts model aggregation weights based on local fairness evaluations performed by clients. Similarly, Astral \cite{djebrouni2024bias} uses an evolutionary algorithm to select clients for aggregation to promote fairness. While these methods maintain data decentralization, they still require clients to compute fairness metrics using sensitive group information, creating potential privacy risks and conflicting with the goals of FWD.

\subsection{Fairness Without Demographics (FWD)}
\label{FWD-rel-work}
The challenge of achieving fairness without access to sensitive attributes has motivated the field of Fairness Without Demographics (FWD). In centralized ML, this has been explored through techniques like adversarial re-weighting to boost the importance of underperforming instances \cite{lahoti2020fairness}. However, in FL, each client operates with its own training instances, making it challenging to identify such instances and apply this approach directly. Another approach, knowledge distillation (KD), was used by \cite{chai2022fairness} to transfer knowledge from a complex teacher to a student model to improve fairness without accessing sensitive information. We adapt this KD concept to FL as a baseline. After initial few rounds of initial training via the FedAvg \cite{mcmahan2017communication} protocol, each client subsequently performs KD using the global model as the teacher, training a local student model on its own data with soft labels from the teacher before sharing their local model for aggregation.

To our knowledge, the only existing work applying FWD directly to FL is FedSRCVaR \cite{papadaki2024federated}. While both our work and FedSRCVaR target fairness without sensitive attributes, their problem formulations differ fundamentally. FedSRCVaR adopts a Rawlsian approach \cite{lahoti2020fairness}, optimizing for a worst-performing subset of individuals defined by a tunable threshold, $\rho$. Selecting an appropriate $\rho$ is non-trivial in heterogeneous FL settings and requires prior knowledge or extensive tuning \cite{guo2021towards}. Furthermore, its evaluation assumes each client represents a single sensitive group. In contrast, our setting assumes clients may have mixed or sparse group representations. Despite a lack of public implementation details, we adapted FedSRCVaR to our setting and report it as a key baseline for a comprehensive evaluation. Our method, \textbf{\textit{CurvFed}}, differs by not imposing constraints on group size, offering a more flexible FWD approach.

\subsection{\color{black} A Loss landscape Primer}
\label{loss-landscape-primer}


\textcolor{black}{The \textit{loss landscape} refers to the geometry of the objective function (or loss) mapped over the high-dimensional space of model parameters. The geometric properties of this landscape, particularly around the local minima found during training, are critical indicators of model performance \cite{li2018visualizing}.}

\textcolor{black}{\textbf{Curvature and Sharpness:} \textit{Curvature} describes the rate at which the loss function changes as model parameters are perturbed\cite{li2018visualizing}. Mathematically, this is characterized by the Hessian matrix (the matrix of second-order derivatives) of the loss function\cite{tran2022pruning,sankar2021deeper}. Regions with high curvature are often referred to as \textit{sharp} minima, where small changes in parameters lead to significant increases in loss. Conversely, \textit{flat} minima are regions with low curvature, where the loss remains relatively stable despite parameter perturbations.}

\textcolor{black}{\textbf{Connection to Model overfitting:} Prior research has established a strong link between the geometry of the loss landscape and model generalization. "Flat" minima are generally associated with better generalization to unseen data, as they indicate a solution that is robust to shifts in the data distribution or noisy parameter updates \cite{hochreiter1997flat, keskar2016large}. In contrast, sharp minima often lead to \textit{model overfitting} \cite{wen2018smoothout,keskar2016large}.}

\textcolor{black}{\textbf{Measuring Curvature:} While the Hessian provides a precise measure of curvature, computing it is computationally prohibitive for deep neural networks \cite{singh2020woodfisher}. In this work, we utilize the top eigenvalue of the Fisher Information Matrix (FIM) as a computationally efficient proxy for the Hessian's top eigenvalue (and thus, the sharpness) \cite{karakida2019universal, liang2019fisher}. This metric allows us to quantify and align the "flatness" of models across federated clients without incurring the heavy cost of full Hessian computation.}

\subsection{Evaluation Metrics}\label{eval-metrices}

We adopt established metrics to assess fairness and its trade-off with accuracy in FL
\cite{ezzeldin2023fairfed, djebrouni2022towards, xu2023fairadabn, chiu2024achieve, li2024bffn}.

\noindent\textbf{Fairness:} We use the \textbf{Equal Opportunity (EO) gap}
\cite{ezzeldin2023fairfed, djebrouni2022towards}, defined as:
\begin{equation}
\label{eq:eo_gap}
\Delta_{\text{EO}} =
P(\hat{Y}=1 \mid S=1, Y=1) -
P(\hat{Y}=1 \mid S=0, Y=1)
\end{equation}
\textcolor{black}{where $\hat{Y}$ denotes the model’s predicted label, $Y$ the ground-truth label,
and $S \in \{0,1\}$ the binary sensitive attribute (e.g., sex or hand).}
The EO gap measures disparities in true positive rates across sensitive groups;
a lower $\Delta_{\text{EO}}$ indicates greater fairness.
Sensitive attributes are used only during post-training evaluation.

\noindent\textbf{Fairness--Accuracy Trade-off:} To evaluate balance, we use the
\textbf{FATE} score
\cite{xu2023fairadabn, chiu2024achieve, li2024bffn}, defined as:
\begin{equation}
\label{eq:fate}
\text{FATE} =
\frac{\text{ACC}_m - \text{ACC}_b}{\text{ACC}_b}
-
\frac{\text{EO}_m - \text{EO}_b}{\text{EO}_b}
\end{equation}
where $\text{ACC}$ denotes classification accuracy,
$\text{EO}$ the Equal Opportunity gap,
subscript $m$ refers to the fair model,
and subscript $b$ denotes the baseline method (FedAvg~\cite{mcmahan2017communication}).
The FATE score jointly captures relative accuracy gains and fairness improvements;
higher values indicate a more favorable fairness--accuracy trade-off.

\section{Theoretical Justification of \textit{CurvFed}}\label{theory-section}
This section theoretically justifies and highlights \textit{CurvFed}'s novelty starting from the problem setting, literature gaps, and challenges. Finally, it outlines the study assumptions.
\subsection{Problem Context and Literature Gap}
\label{problem_statement}

\textbf{Fair FL Setting:}
In a human-sensing fair FL framework, a central server coordinates training across \( q \) distributed edge platforms, referred to as clients. Each client \( t \in \{1, 2, \dots, q\} \) holds its own local dataset \( \mathcal{D}_t \), which may contain data from \textit{one or more individuals}. Each client private dataset, \( \mathcal{D}_t = \{(x_k, y_k)\}_{k=1}^{n_t} \), where \( x_k \) is an input instance, \( y_k \) is the corresponding label, $n_t$ is the number of data instances for the client $t$.

We can consider $\mathcal{D}_t$ as a union of $m$ disjoint groups, $\mathcal{D}_t = \bigcup_{i=1}^{m} G_i$, where each group $G_i$ represents a subset of data distinguished by attributes such as, demographic characteristics (e.g., sex, age) or contextual factors (e.g., sensor placement or orientation). The goal of \emph{performance-based group fairness in this fair FL framework} is to ensure that all such groups—regardless of which client they belong to—receive similar model performance.

Each client $t$ trains a local model using $\mathcal{D}_t$ with parameters \( \theta^{*}_t \) \textcolor{black}{by minimizing an empirical risk function $J(\theta,\mathcal{D}_t)$, which represents the average empirical loss over the local dataset,}


\begin{equation}
\begin{split}
    \theta^{*}_t &= \argmin_{\theta} \quad J(\theta,\mathcal{D}_t) \\
    &= \argmin_{\theta} \quad \frac{1}{|\mathcal{D}_t|}\sum_{i}^{n_t} \ell(f_{\theta}(x_i),y_i)
\end{split}
\end{equation}

Here, $f_{\theta}$ is the predictive model parameterized by $\theta$, and $\ell(.)$ denotes a non-negative loss function that evaluates prediction error.

After local training, the server aggregates the client models \( \{\theta_1^*, \dots, \theta_q^*\} \) into a global model with parameter \( \theta^* \). This aggregation is typically done through a weighted average, as shown in Eq. \ref{aggregation-eq-1}, where each client is assigned a non-negative weight \( \beta_t \) such that \( \sum_{t=1}^{q} \beta_t = 1 \), with the goal of minimizing its \textit{excessive loss} ($R$) \cite{papadaki2024federated}, shown in Eq. \ref{excessive-loss-1}.

\begin{equation}
    \theta^{*} = \sum_{t=1}^{q} \beta_t \theta_t^*
    \label{aggregation-eq-1}
\end{equation}

\begin{equation}
R(t) := J(\theta^{*},\mathcal{D}_t) - J(\theta^{*}_t,\mathcal{D}_t) 
\label{excessive-loss-1}
\end{equation}

\textcolor{black}{Here, $\theta_t^{*}$ denotes the locally optimal model parameters obtained by client $t$, whereas $\theta^{*}$ represents the aggregated global model parameters produced by the central server.}
\textcolor{black}{The excessive loss $R(t)$ quantifies the performance degradation experienced by client $t$ when using the global model instead of its locally optimized model, evaluated on $\mathcal{D}_t$.}

A key fairness concern is that the global model may disproportionately favor some clients. To address this, prior work \cite{papadaki2024federated} introduces the concept of excessive loss in Eq. \ref{excessive-loss-1}—the performance gap between a client’s local and the global model—and promotes fairness by minimizing its variance across clients : 


\begin{equation}
    \argmin \; \text{Var}_t[R(t)] = \frac{1}{q} \sum_{t=1}^{q} \left( R(t) - \bar{R} \right)^2, \quad \text{where} \quad \bar{R} = \frac{1}{q} \sum_{t=1}^{q} R(t)
    \label{exss-loss-var-eq}
\end{equation}

\textcolor{black}{where, the operator $\text{Var}_t[\cdot]$ denotes variance computed across all clients indexed by $t \in \{1,\dots,q\}$.}

This objective ensures that model utility is distributed equitably across clients.\textcolor{black}{ Throughout this work, $\| \cdot \|$ denotes the Euclidean norm, vectors and matrices are represented by bold symbols, and sets are denoted by uppercase letters, unless stated otherwise.}


\textbf{Challenge of this paper's FWD problem setting:} 
Our FWD objective is to promote fairness in FL beyond the scope discussed above. 
We aim to minimize disparities in model performance across all groups of individuals—regardless of which client they belong to—even when sensitive attributes (e.g., sex, age) that constitute the groups or underlie the disparities are unknown, latent, or unavailable. This consideration is especially critical in human-sensing applications, where a single client may represent data from one or more individuals with diverse and often unobservable characteristics. Consequently, achieving uniform performance across clients does not guarantee equitable treatment for all underlying groups or individuals.
Our proposed solution builds on Eq. \ref{exss-loss-var-eq} to attain this FWD objective.

\subsection{FWD in FL: A Curvature-Based Roadmap}
\label{roadmap}
We begin by establishing an upper bound on a client’s excessive loss Eq. \ref{excessive-loss-1}, which helps isolate the disparity factors in terms of loss curvature within the client's data, \( \mathcal{D}_t \), and between the client's and the global model. The following theorem and justifications assume that the loss function $\ell(\cdot)$ is at least twice differentiable, as is common for standard objectives such as cross-entropy and mean squared error.

\textbf{Theorem 1} Through Taylor expansion, the upper bound of excessive loss $R(t)$ of the global model with parameters $\theta^{*}$ computed using client $t$'s private dataset, $\mathcal{D}_t$ is formalized as:

\begin{equation}
\label{theorem1}
    R(t) \leq \|g_{t}^{\ell} \| \times \| \theta^{*} - \theta^{*}_{t} \| + \frac{1}{2}  \lambda( \textbf{H}_{t}^{\ell}) \times \| \theta^{*} - \theta^{*}_{t} \|^{2} + O(\|\theta^{*} - \theta^{*}_{t} \|^3).
\end{equation}

Here, $g_t^{\ell}= \nabla_{\theta} \ell(\theta^{*}_{t},\mathcal{D}_{t})$ is the local models gradient vector, and $\textbf{H}_{t}^{\ell} = \nabla^2_{\theta}\ell(\theta^{*}_{t},\mathcal{D}_t)$ is the local model's Hessian at optimal parameter $\theta^{*}_t$ associated with the loss function $\ell$. The term $\lambda(\cdot)$ denotes the maximum eigenvalue (hereby referenced as the top eigenvalue) and \textcolor{black}{the big-oh term $O(\|\theta^{*} - \theta^{*}_{t} \|^3)$ represents all the higher order terms in the expansion.} Equation \ref{theorem1} reveals that the excessive loss is tightly linked to the client $t$'s (i) gradient vector corresponding to the loss, i.e., $g_{t}^{\ell}$ and (ii) the top Hessian eigenvalue, a proxy of the curvature (sharpness) of the local loss landscape.

\textit{\textbf{Sharpness, Curvature, and Fairness:}} High curvature in the loss landscape of model optimization, characterized by sharp minima, typically associated with large top Hessian eigenvalues—is known to impair generalization \cite{dauphin2024neglected, jastrzkebski2018relation}. Importantly, disparities in two groups (based on demography or other factors) in terms of their top Hessian eigenvalues have been shown to correlate with the model's performance disparities between those groups \cite{tran2022pruning}, suggesting that mitigating disparities among all groups' curvatures (top Hessian eigenvalues) will promote fairness.

However, \emph{there are several challenges to implementing this:}
\begin{enumerate}[nosep]
    \item Computing the full Hessian matrix is computationally costly \cite{dong2017learning}, thus, infeasible on resource-constrained edge devices.
    \item For us, the groups are unknown, because we are unaware of the factors (sensitive attributes) creating disparities.
    \item In our setting, some clients can be just one person, implying all data can belong to just one group (at least in terms of demography or individual attributes like left or right-handed).
\end{enumerate}

To address these challenges, we developed \textbf{\textit{CurvFed}} — a principled framework that achieves fairness without explicit sensitive attribute labels (or knowledge) by regulating sharpness at both local client-level and global levels during model aggregation. Our method is guided by the following three propositions:

\textbf{Proposition 1. Fisher Approximates Hessian}

To address the \emph{first challenge}, we leverage the empirical Fisher Information Matrix (FIM) as an efficient proxy for the Hessian:

\begin{equation}
\label{fisher}
\textbf{H} \approx \textbf{F} = \frac{1}{n_t} \sum\limits_{i=1}^{n_t} \nabla_{\theta}\ell(f_{\theta}(x_{i}),y_{i}) \nabla_{\theta}\ell(f_{\theta}(x_{i}),y_{i})^{T}.
\end{equation}
\textcolor{black}{For notational simplicity, $\mathbf{H}$ here denotes the client-specific Hessian $\mathbf{H}_t^\ell$ evaluated at $\theta_t^*$.}

Lee et. al. \cite{lee2022masking} describe the relationship between the empirical FIM ($\textbf{F}$) and the Hessian matrix ($\textbf{H}$) as shown in equation \ref{fisher}. Where $\nabla_{\theta}\ell(f_{\theta}(x_{i}),y_{i})$ is the the gradient vector corresponding to the loss of the $i^{th}$ data sample, for a model $f$ with parameters $\theta$ and the total number of data samples represented by $n_t$. The empirical FIM thus serves as a \emph{computationally efficient} approximation of the Hessian, reducing computational costs during training \cite{singh2020woodfisher,lee2022masking}.

\textbf{Proposition 2. Minimizing the FIM Top Eigenvalue Reduces Group-wise Curvature Disparity.}

Building on Proposition 1, which establishes the FIM as a tractable approximation of the Hessian, we now provide a theoretical argument that minimizing the top eigenvalue of the FIM (denoted as $\lambda(F_t))$ during client-level training reduces curvature disparity across all constituent groups $G_1, G_2, \ldots, G_m$ within the local dataset $\mathcal{D}_t = \bigcup_{i=1}^{m} G_i$.

The full-dataset FIM can be expressed as a convex combination of group FIMs:
\begin{equation}
F_t = \sum_{i=1}^m \alpha_i F_{G_i}, \quad \text{where } \alpha_i = \frac{|G_i|}{|\mathcal{D}_t|}, \quad \sum_i \alpha_i = 1.
\end{equation}

Using Jensen's and Weyl's inequalities \cite{boyd2004convex, horn2012matrix}, with an assumption of approximate dominant eigenvector alignment for the lower bound, we establish bounds: 
\begin{equation}
\label{group_wise_disparity}
\max_{i} \lambda(F_{G_i}) - \delta \leq \lambda(F_t) \leq \sum_{i=1}^m \alpha_i \lambda(F_{G_i})
\end{equation}
where $\delta \geq 0$ accounts for potential eigenvector misalignment. 


Our key insight is that during client training, penalizing $\lambda(F_t)$ implicitly drives the groups' top eigenvalues $\lambda(F_{G_i})$ towards uniformity. Notably, as shown in Eq. \ref{group_wise_disparity}, minimizing $\lambda(F_t)$ effectively reduces the top eigenvalue of the group with the highest curvature, i.e., highest top eigenvalue of FIM ($\max_{i} \lambda(F_{G_i}) - \delta$). Intuitively, in the loss landscape, continuing to lower the dominant curvature reduces the variance across all groups' curvatures, promoting uniformity. \emph{This approach is group-agnostic, relying purely on optimization geometry, thus addressing the second challenge discussed above.}
Notably, empirical evidence supporting this theoretical reduction in curvature disparity is presented in Section \ref{empirical-verification}.

\textbf{Proposition 3. Sharpness-Aware Aggregation Reduces Variance in Excessive Loss Across Clients.}

Building upon the understanding of excessive loss \(R(t)\) from Theorem 1 and the local curvature control from Proposition 2, we now posit that our sharpness-aware aggregation scheme directly contributes to reducing the variance of excessive loss, \(\text{Var}_t[R(t)]\), across clients-both in terms of performance and curvature. This promotes fairer performance distribution.

From Theorem 1 and Proposition 2, and assuming that the discrepancy \(\delta_t = \|\theta^* - \theta_t^*\|\) remains sufficiently small- since the local training across clients occurs for a few epochs before each aggregation (allowing us to neglect higher-order terms and its variances), the excessive loss for client \(t\) is approximately:
\[
R(t) \leq \|g_{t}^{\ell} \| \cdot \delta_t + \frac{1}{2} \lambda( \mathbf{F}_{t}^{\ell}) \cdot \delta_t^{2}
\]
Here, \textcolor{black}{The quantity $\delta_t$ measures the parameter-space deviation between the global model and client $t$’s locally optimal model.}
\(g_t^\ell\) is the local model's gradient at its optimum \(\theta_t^*\), and \(\lambda(\mathbf{F}_t^\ell)\) is the top eigenvalue of its FIM, serving as a proxy for Hessian's top eigenvalue (following Proposition 1). This indicates that for a client $t$, its excessive loss is impacted mostly by \(g_t^\ell\) and \(\lambda(\mathbf{F}_t^\ell)\), which may vary significantly across clients.
Our variance reduction strategy builds on this observation. During its local training, each client $t$ seeks to minimize its empirical risk $J(\theta, \mathcal{D}_t)$. Successful training results in a small local gradient norm $\|g_t^\ell\|$ at $\theta_t^*$. Furthermore, as established in Proposition 2, our framework encourages clients to find solutions $\theta_t^*$ that also exhibit low curvature, i.e., a minimized $\lambda(\mathbf{F}_t^\ell)$. That means an effective locally trained client will have lower \(g_t^\ell\) and \(\lambda(\mathbf{F}_t^\ell)\), thus lower excessive loss, and the opposite holds for ineffective locally trained ones.

\textbf{\textit{CurvFed}'s} \textit{Sharpness-Aware Aggregation} builds on standard fair FL aggregation (Eq. \ref{aggregation-eq-1}-\ref{exss-loss-var-eq}) with the novel incorporation of curvature information into the weighting scheme. Specifically, client weights $\beta_t$ in Eq. \ref{aggregation-eq-1} are computed based on excessive loss $R(t)$, which accounts for both the local loss (corresponding to \(g_t\)) and top FIM eigenvalue $\lambda(F_t)$ (detailed in Section \ref{global}).
Clients with lower $R(t)$, i.e., lower $g_t$ and $\lambda(F_t)$ receive higher weights. This serves two goals: First, it \emph{prioritizes effectively trained clients}—those with low loss (indicated by small \(g_t\)) and low curvature $\lambda(F_t^\ell)$), indicating locally fairer. These clients contribute more to the global model, resulting in lower and more uniform excessive loss $R(t)$ across them. And Second, the scheme effectively \textit{mitigates "outlier" influence}: conversely, clients that exhibit high $R(t)$, which would lead to a higher $\text{Var}_t[R(t)]$, are assigned lower weights, curtailing their ability to inflate the $\text{Var}_t[R(t)]$ in Eq. \ref{exss-loss-var-eq}.

Because the aggregation is driven by excessive loss, \textit{CurvFed} is \emph{agnostic to demographic (i.e., sensitive attributes) knowledge and the number of individuals belonging to clients, addressing challenges 2 and 3 discussed above.}
In summary, by emphasizing well-trained, low-curvature clients, CurvFed reduces \(\text{Var}_t[R(t)]\) in Eq. \ref{exss-loss-var-eq}, guiding the global model toward fairer convergence, promoting group fairness without explicit demographic (i.e., sensitive attributes) knowledge.

\subsection{\textcolor{black}{Synthesis of Theoretical Insights: Curvature Alignment as a Proxy for Fairness Without Demographics (FWD) in Human-centric federated learning (FL)}}\label{Summary-Problem-Formulation}

{\color{black}
The above theoretical results motivate curvature alignment as a principled and operational mechanism for achieving fairness under the challenging setting of FWD in human-centric FL.

As discussed in Section~\ref{rel-fairness-in-FL}, conventional fairness-attaining approaches rely on explicit group-based metrics (e.g., the Equal Opportunity gap), which require access to sensitive or bias-inducing attribute labels. In our FWD setting, such labels are unavailable during training, and FL privacy constraints prohibit their use. Therefore, standard group-equalization strategies are not implementable. Moreover, human-sensing systems often involve multiple simultaneous bias-inducing factors (e.g., sex, body type, sensor placement)~\cite{jui2024fairness}, and mitigating one explicitly may exacerbate another~\cite{djebrouni2024bias}. Therefore, this paper's challenge becomes more complex: ensuring fairness under unknown and potentially multi-attribute biases, rather than across predefined groups with known sensitive attributes.

Under these constraints, we pursue an alternative pathway grounded in recent ML optimization theories linking loss-landscape geometry to group-wise performance disparities. As Section \ref{theory-section}, establishes that a client’s local model’s loss landscape curvature being flatter reduces disparities in generalization performance across groups within that client’s local data (within client disparity reduction established through Theorem 1 and Proposition 1), and the loss landscape curvatures being similar across clients’ local models reduce disparities in generalization performance across clients’ local data (inter-client established in Proposition 2). Thus, loss landscape curvature is not treated as fairness itself, but as a measurable, optimization-level quantity that upper bounds performance disparity in the absence of information about demographic or bias-creating factors.

In summary, by promoting (1) Flatter curvature within clients, we reduce intra-client (even unknown) group-wise disparity, and (2) Similar curvature across clients, we reduce inter-client local data performance disparities. These together provide an attribute-agnostic pathway toward fairness across groups, whether a single or multiple sensitive or demographic factors are creating the disparity.

Importantly, curvature information (via the FIM top eigenvalue) does not reveal demographic information and can be computed locally, satisfying the constraints of privacy-preserving fairness without demographic (FWD) scope of FL. This makes curvature geometric consistency a feasible operational mechanism to attain FWD, whereas conventional fairness mechanisms are not implementable under the same constraints.
}

\subsection{Assumptions}\label{assumtion}

This study operates under the following assumptions:

\begin{enumerate}[nosep]
    \item We consider a centralized FL setup where a server coordinates training without access to clients' local datasets \( \mathcal{D}_t \). No external public datasets are assumed to approximate the overall data distribution.
    
    \item Our goal is to promote \textbf{performance-based group fairness} by ensuring the global model performs equitably across groups defined by sensitive attributes (e.g., sex, age, sensor location). For a sensitive attribute \( a \), let \( G_c^a = \{(x_i, y_i) \mid x_i \text{ belongs to the category } c\} \) with \( c \in \{1, \dots, C_a\} \)-categories of \( a \). Fairness entails minimizing performance disparities across these groups, such as ensuring similar accuracy for \( G_{\text{male}}^{\text{sex}} \) and \( G_{\text{female}}^{\text{sex}} \).

    \item We assume the central server is not resource-constrained and, consistent with standard FL approaches (e.g., FedAvg\cite{mcmahan2017communication}), can perform model aggregation efficiently.

    \item \textcolor{black}{We assume that clients permit sharing loss-landscape curvature statistics (e.g., the top FIM eigenvalue) during FL aggregation. These statistics capture aggregate optimization geometry and do not involve sharing raw data, gradients, or sensitive attributes. To the best of our knowledge, there is no evidence that such curvature summaries can be used to reconstruct demographic or sensitive attribute information; thus, satisfying the FWD setting.}
\end{enumerate}

\section{\textit{CurvFed}: Design Choices and Approach}
\label{approach}

\begin{figure}[h!]
    \centering
    \includegraphics[width=0.6\linewidth]{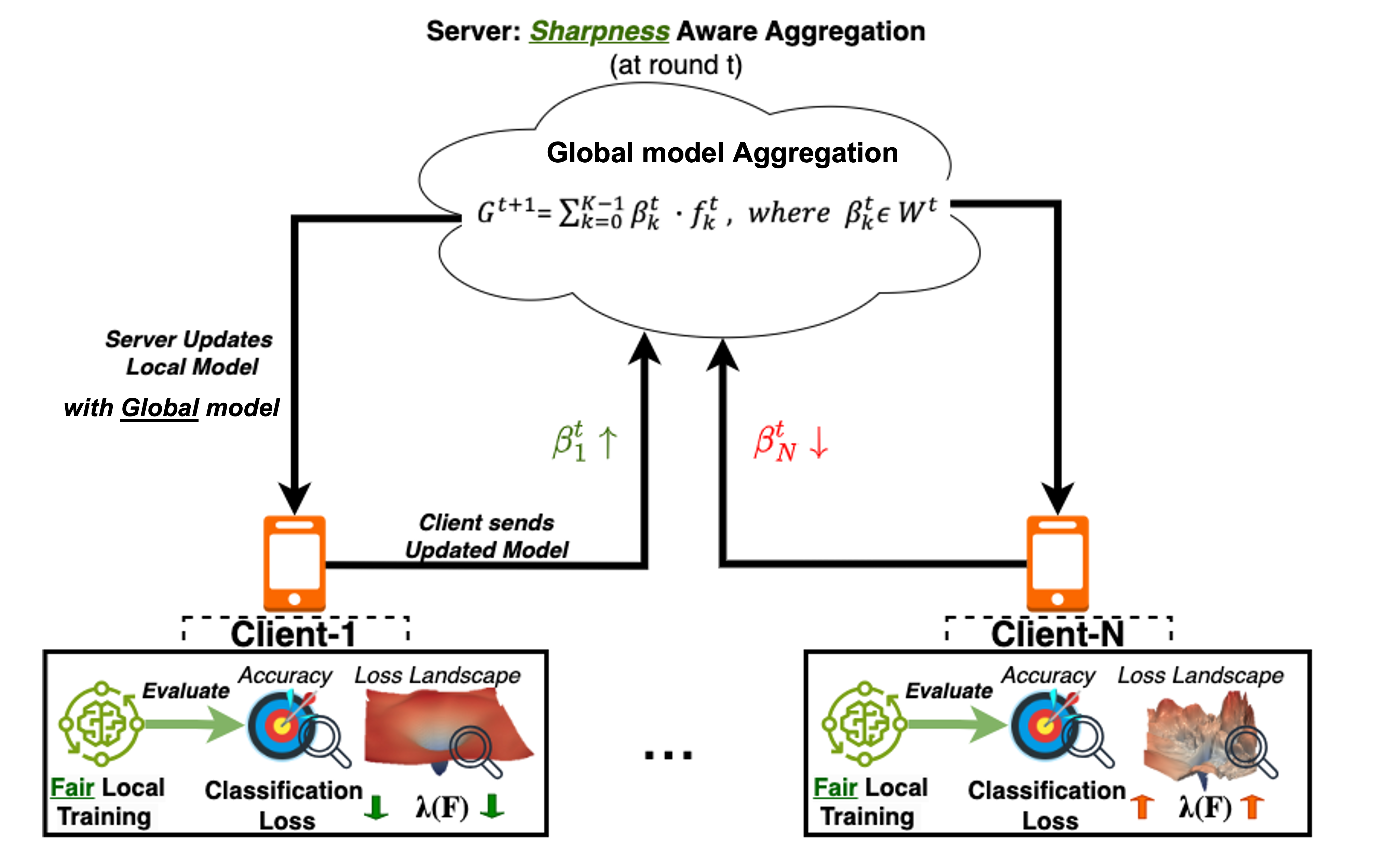}
    \caption{\color{black}{Overview of \textit{CurvFed}, illustrating client-side training, sharpness-aware aggregation,
and global model updates in federated learning.}}
    \label{fig:approach}
\end{figure}

Fig. \ref{fig:approach} provides an overview of \textbf{\textit{CurvFed}}. In each FL round, clients use their own individuals’ data to train local models, aiming to optimize accuracy and promote local fairness. The server then aggregates these models using our \textit{Sharpness-aware aggregation} strategy to enhance fairness and accuracy. The global model is redistributed to clients for the next round. Details of \textbf{\textit{CurvFed}} is below.

\subsection{Design Choices}\label{approach-design-choice}
Our FWD attaining design choices leverage loss-landscape curvature at both client and global aggregation levels, grounded in the theoretical justifications in Section \ref{theory-section}, are presented below:


\subsubsection{Fair Local Training Objective: }
\label{fair_local_training}
In the FL context, each client learns a model, denoted as \(f_{\text{client}}\). We have two objectives on the client side: (1) maximizing accuracy and (2) promoting fairness, i.e., mitigating bias within the client's own data distribution.

We introduce a curvature regularization term in the client's loss function to enhance group fairness. The learning objective at the client-level is formulated as:  

\begin{equation}
\label{local_training_eq}
\begin{aligned}
\arg\min \; \text{Loss}_{\text{client}} &= \alpha \cdot \ell_{\text{client}}(f_{\text{client}}(x_i)) \\
&\quad + (1-\alpha) \cdot \frac{1}{N_{\text{correct}}} \cdot \lambda \left( F^{\ell_{\text{client}}} \right)_{\text{correct}}
\end{aligned}
\end{equation}  

Here, $N_{\text{correct}}$ represents the number of correctly classified instances in a batch, and $x_i$ is the input in the $i^{th}$ batch. The term $\ell_{\text{client}}(f_{\text{client}}(x_i))$ ensures accuracy maximization by minimizing classification loss.  The regularization term penalizes the top eigenvalue of the FIM, i.e., \(\lambda(F^{\ell_{\text{client}}})_{correct}\), associated with the loss for correctly classified samples, promoting group fairness as outlined in \emph{Proposition 2} in Section \ref{theory-section}.

\par 
In equation \ref{local_training_eq}, we introduce $\alpha$, a hyperparameter that acts as a balancing factor between the two objectives. $\alpha$ is identified using a grid search in this work. However, popular hyperparameter optimization frameworks like Optuna \cite{akiba2019optuna} can be used to find the optimal $\alpha$. \textcolor{black}{In our evaluations, the $\alpha$ values identified via grid search were effective across all three datasets, with one dataset exhibiting an optimal value within a similar range. A detailed discussion of the sensitivity to the weighting factor is provided in Section~\ref{alpha}.} This consistency suggests that the local optimization strategy transfers reasonably well across heterogeneous human-sensing datasets. We will use $\lambda(F_{client})$ instead of $\lambda(F^{\ell_{\text{client}}})_{\text{correct}}$ for simplicity for the rest of the section.


\subsubsection{Sharpness-aware Aggregation to Promote Fairness: }
\label{global}

After local training, each client has a distinct accuracy and top eigenvalue of FIM ($\lambda(F_{client})$), as depicted in Fig. \ref{fig:approach}, leading to varying excessive losses. As outlined in \emph{Proposition 3} in Section \ref{theory-section}, clients with lower excessive losses should contribute more to deriving the aggregated global model.

This aggregation strategy draws inspiration from Pessimistic Weighted Aggregation (P-W) in FL \cite{djebrouni2022towards}, which excludes clients with known biases during aggregation. However, in the FWD setting, such bias information is unavailable. Instead, our \textit{Sharpness-aware Aggregation} implicitly promotes fairness by assigning higher weights to clients with lower excessive loss, i.e., having lower top eigenvalue of FIM and lower classification error (according to \emph{Proposition 3}), thus supporting fairer and more performant global updates without relying on explicit sensitive attributes.

To compute the aggregation weights for $N$ clients: \( W^r = \{ \beta^r_n \mid \forall n \in \{1, \dots, N\} \} \) at each round \( r \), the server collects each client's evaluation loss \( \text{loss}_n^{\text{eval}} \) and the top FIM eigenvalue \( \lambda(F_n^{\text{eval}}) \). These are then used to compute the aggregation weights as described in Equation~\ref{agg-combine-weight_eqn}.

\begin{equation}
W^{r} = S(S(L) \cdot S(T))
\label{agg-combine-weight_eqn}
\end{equation}

Here, we use the \textit{Softmax operator} denoted as ${S}$ for simplicity.
Notably, $L = \left\{ \epsilon + \frac{1}{\text{loss}_{n}^{\text{eval}}} \right\}_{n=1}^{N}$ and $T = \left\{ \epsilon + \frac{1}{\lambda(F_{n}^{\text{eval}})} \right\}_{n=1}^{N}$ and \(\epsilon\) is a small constant to avoid division by zero in weight calculations.


\subsection{Approach}

Algorithm \ref{alg:algo-1} outlines \textbf{\textit{CurvFed}} approach, which begins with the following inputs: a randomly initialized global model denoted as $G^{0}$, the number of clients participating in FL process represented by $\text{N}$, the total number of rounds for the FL process indicated by $\text{R}$, the number of epochs $\text{E}$ for conducting local training on each client's side, the cycle length $\text{c}$ for implementing the stochastic weight averaging (SWA) protocol \cite{caldarola2022improving}, the weighting factor $\alpha$ used to balance the client loss, the batch size $\text{B}$ utilized by the clients during training, client learning rate $\gamma$ and learning rate scheduler $\gamma()$.
Different modules of \textbf{\textit{CurvFed}} are detailed below:

\begin{algorithm}[h!]
    \caption{Curvature-Aligned Federated Learning (\textbf{\textit{CurvFed}})}
    \small
    \label{alg:algo-1}
\begin{algorithmic}[1]
   \STATE {\bfseries Input:}:
    $~\diamond~ G^{0}$:  Initial random model,
    $~\diamond~ \text{N}$: Number of Clients, 
    $~\diamond~ \text{R}$: Total rounds,
    $~\diamond~ \text{E}$: local training epochs, 
    $~\diamond~ \text{c}$: cycle length,
     $~\diamond~ \eta$: SWA starting threshold,
    $~\diamond~ \alpha$: client loss weighting factor,
    $~\diamond~ \text{B}$: Client batch size,
    $~\diamond~ \gamma$: Learning rate, 
    $~\diamond~ \gamma()$: Learning rate scheduler\\ 
    
    \STATE {\bfseries{Parameter}}:
    $~\diamond~ \text{L[]}$: Array to store classification loss of clients, 
    $~\diamond~ \text{T[]}$: Array to store top eigenvalue of FIM of clients,
    $~\diamond~ S$: Softmax operator\\
   \STATE {\bfseries{Returns}}: $G_{\text{SWA}}$: Global model
        \FOR{each round r=0 to R-1} 
            \IF {$ r= \eta \times R$ } 
                \STATE $G_{\text{SWA}} \leftarrow G^{r}$
            \ENDIF
            \IF {$n \geq \eta \times R$} 
                \STATE $\gamma = \gamma (r) $
            \ENDIF
            \FOR{each $Client_{\text{n}}$ n=0 to N-1}

                \STATE $params \leftarrow G^r,Client_n,E,B,\alpha,\gamma $
                \STATE $f_{\text{n}}^{r},loss_{\text{n}}^{r},\lambda(F_{\text{n}}^{r}) \leftarrow TrainClient(params)$
                \STATE L[n] $\leftarrow \epsilon + \frac{1}{loss_{\text{n}}^{r}}$
                \STATE T[k] $\leftarrow \epsilon +\frac{1}{\lambda(F_{\text{n}}^{r})}$
            \ENDFOR
            \STATE $\text{W}^{r}=S(S(L) \cdot S(T) )$
            \STATE $G^{r+1} = \sum_{n=1}^{N} \beta_{n}^{r} \cdot f_{n}^{r} \quad \forall \beta_{n}^{r} \in \text{W}^{r}$

            \IF{$r \geq \eta \times R$ and $mod(r,c) = 0$}
                \STATE $ n_{models} \leftarrow n/c $
                \STATE $ G_{\text{SWA}} \leftarrow \frac{n_{model} \times G_{\text{SWA}}+G^r}{n_{models} + 1} $
            \ENDIF
        \ENDFOR
        \RETURN{ $G_{\text{SWA}}$} 
    \end{algorithmic}
\end{algorithm}

\subsubsection{Local Client Training: } \label{Approach-local-client-training}
Following standard FL protocol \cite{mcmahan2017communication}, the server initializes by distributing a randomly initialized global model $G^{0}$ and other training parameters to the clients for local training through the $TrainClient$ module (line 12-13). 

\textbf{TrainClient Module}, as shown in Algorithm \ref{alg:Client-training}, receives the global model, along with training configurations such as the number of local training epochs $E$, batch size $B$, learning rate $\gamma$, and the client loss weighting factor $\alpha$ from the server. On the client side, data is divided into a training set $\mathcal{D}_{train}$ and an evaluation set $\mathcal{D}_{eval}$. The model is trained on $\mathcal{D}_{train}$ (line-9), where we utilized the Sharpness-Aware Minimization (SAM) optimizer for model training\cite{foret2021sharpnessaware}. The choice of the model optimizer is influenced by prior works in FL, which claim that using SAM optimizer at the client level helps with convergence and promotes local flatness \cite{caldarola2022improving}, thus lowering curvature and further promoting fairness. 

\begin{algorithm}[h!]
\centering
\caption{$TrainClient (G^{t},Client_k,E,B,\alpha,\gamma )$}
\label{alg:Client-training}
\small
\begin{algorithmic}[1]
    \STATE {\bfseries{Parameter}}:
    $~\diamond~ G^{t}$:   Initial model sent from Server side,
    $~\diamond~ \text{E}$: local training epochs at Client side, 
    $~\diamond~ Client_k$: Client-ID,
    $~\diamond~ \text{B}$: Client Batch size,
    $~\diamond~ \alpha$: client loss weighting factor,
    $~\diamond~ \gamma$: client learning rate 
    
\STATE {\bfseries{Returns}}: $f_{\text{n}}^{r},\text{Loss}_{eval,n}^{r},\lambda(F_{\text{eval},n}^{r})$
    
        \STATE \textit{Initialize:} $f_{\text{n}}^{r} \leftarrow G^r$ \COMMENT{Server-side global model}
        \STATE \textit{Initialize:} $l_{\text{n}}$ \COMMENT{Client Classification loss}
        \STATE \textit{Initialize:} SAM optimizer
        \STATE \textit{Split:} $Client_n$ data $\mathcal{D}_n$ into $\mathcal{D}_{train}$ and $\mathcal{D}_{eval}$
        
        \FOR{each epoch e=0 to E-1}
            \FOR{each batch i=0 to $\text{B}$ in $\mathcal{D}_{train}$ }
                    \STATE $argmin \text{Loss}_{\text{n}}$ 
            \ENDFOR

                \STATE \textit{Evaluate} $ f_{\text{n}}$ on $D_{eval}$ get: $\text{eval\_}loss_{\text{n}}, \text{eval\_}\lambda(F_{n})$

                \STATE $\text{Loss}_{eval,n}^{r} \leftarrow \text{eval\_}loss_{\text{n}}$
                \STATE $\lambda(F_{\text{eval},n}^{r}) \leftarrow \text{eval\_}\lambda(F_{n})$\\

        \ENDFOR
    
        \RETURN {$f_{\text{n}}^{r},\text{Loss}_{eval,n}^{r},\lambda(F_{\text{eval},n}^{r})$}
        
    \end{algorithmic}
    \vspace{-2pt}
\end{algorithm}

The client-local training follows the dual objective discussed in Section \ref{fair_local_training}, as per Equation \ref{local_training_eq}. The model performance is assessed on the disjoint $\mathcal{D}_{eval}$ (line-11) to identify the best-performing model. For this model, the evaluation loss $\text{eval\_}loss_{n}$ and the top eigenvalue of the FIM $\text{eval\_}\lambda(F_{n})$ are calculated. The optimal model's evaluation loss and FIM top eigenvalue are stored in $\text{Loss}_{\text{eval},n}^{r}$ and $\lambda(F_{\text{eval},n}^{r})$, respectively (Lines 12,13). These metrics and the weights of the selected model are then sent back to the server (line-15).

\subsubsection{Sharpness-aware Aggregation: }\label{Algo-aggregation}
Upon receipt of the clients' local training returns, as per Algorithm \ref{alg:algo-1}, the server starts the model aggregation phase. Our \textit{Sharpness-aware aggregation} method calculates weights for each client based on their classification loss and the top eigenvalue of the FIM related to their loss (lines 14,15). Clients with lower classification loss and lower top eigenvalue of FIM are assigned higher weights (line 17). 
\par
\textbf{Stochastic Weight Averaging (SWA): }
On the server side, we also use SWA \cite{izmailov2018averaging}, a technique shown to enhance generalization in FL systems \cite{caldarola2022improving}. The significance of generalization in fair models is underscored by research \cite{ferry2023improving,cotter2018training} focusing on improving their robustness and ability to perform fairly across unseen data. 
This addresses the challenge of fairness constraint overfitting \cite{cotter2018training}, where models fair on training data may still exhibit unfairness on unseen data. In human-sensing FL, ensuring fairness and robustness across unseen instances is critical; hence, the adoption of SWA is crucial in \textbf{\textit{CurvFed}}.
SWA averages the global models at regular intervals, i.e., cycles $c$ (line 19), and adjusts the learning rate throughout the FL process (line 9) to achieve flatter minima, thereby preventing overfitting as the process advances. The SWA process begins after a threshold ($\eta$) number of rounds are complete (lines 4-7); for our evaluations, this threshold is set to $\eta = 20\%$. The early start threshold was chosen due to the limited data size, which allows for quicker convergence in our setting.

\section{Experimental Evaluation}
\label{sec:results}
This section provides a comprehensive evaluation of \textbf{\textit{CurvFed}}, covering datasets and models (Section~\ref{dataset}), followed by an analysis addressing four key research questions on CurvFed’s effectiveness (Section~\ref{eval-results}).  \textcolor{black}{In this work, we compared \textbf{\textit{CurvFed}} against standard benign FL models—FedAvg\cite{mcmahan2017communication}, FedSAM \cite{qu2022generalized}, and FedSWA \cite{caldarola2022improving}—and two recent fairness-without-disparity (FWD) approaches: KD-FedAvg \cite{chai2022fairness} (adopted for FL settings) and FedSRCVaR \cite{papadaki2024federated}.} 


\subsection{\textcolor{black}{Evaluation Datasets}}

\label{dataset}

\begin{table}[h]
\color{black}
\centering
\resizebox{0.99\linewidth}{!}{
\begin{tabular}{c|c|c|c|c}
\hline
\textbf{Dataset} & \textbf{Sensing Modality} & \makecell{\textbf{\# of Participants}\\ \textbf{($N$)}} & \textbf{Application Task} & \textbf{Sensitive Attributes} \\
\hline
\textbf{WIDAR}~\cite{zhang2021WIDAR3} & WiFi (CSI Velocity Profiles) & $N=12$ &Gesture Recognition & Sex, Sensor Orientation \\
\textbf{HugaDB}~\cite{chereshnev2018hugadb} & Inertial (Accel + Gyro) &$N=18$ &Human Activity Recognition & Sex \\
\textbf{Stress Sensing}~\cite{xiao2024reading} & Physiological (EDA + Accel) & $N=48$ & Stress Detection & Sex, Watch Hand \\
\textbf{PERCEPT-R}~\cite{benway2022percept} & Audio (Speech) & $N=76$ &Clinical Speech Assessment & Sex \\
\hline
\end{tabular}
}
\caption{\color{black} Summary of datasets used for evaluation. The diversity in sensing modalities and tasks demonstrates the generalizability of \textbf{\textit{CurvFed}} across the human-sensing spectrum.}
\label{tab:dataset_summary}
\vspace{-2em}
\end{table}

{\color{black}
We evaluate \textbf{\textit{CurvFed}} on \textit{four} heterogeneous human-sensing datasets, summarized in Table~\ref{tab:dataset_summary}. These datasets were selected to test generalizability across diverse sensing modalities (WiFi, inertial, physiological, and audio) and distinct sensing tasks (HAR, stress detection, and speech therapy). Crucially, each dataset contains different sensitive attributes as biological sex, sensor placement, that challenge the fairness of standard FL algorithms \cite{yfantidou2023beyond,djebrouni2024bias}. In all our evaluation, the sensitive attributes were used solely for \textit{\underline{post-hoc fairness assessment}}.

\textbf{WIDAR (WiFi-based HAR):} This dataset leverages Channel State Information (CSI) derived from WiFi signals \cite{ratnam2024optimal} to capture body-coordinate velocity profiles during human activities \cite{zhang2021WIDAR3}. Data were collected from $16$ participants performing a set of gestures under varying conditions. To evaluate the robustness of \textbf{\textit{CurvFed}}, we consider biological \textit{sex} and receiver \textit{orientation} as sources of natural variability across participants.

\textbf{HugaDB (Inertial HAR):} HugaDB \cite{chereshnev2018hugadb} focuses on wearable activity recognition and comprises accelerometer and gyroscope data collected from body-worn sensors on $18$ participants. The dataset provides a testbed for assessing whether \textbf{\textit{CurvFed}} can achieve equitable performance across individuals with differing physiological characteristics, without relying on demographic information during training. For fairness evaluation purposes, biological \textit{sex} is designated as the sensitive attribute.

\textbf{Stress Sensing (Physiological):} To evaluate multi-modal robustness, we utilize the Stress Sensing dataset \cite{xiao2024reading} ($48$ participants). It combines electrodermal activity (EDA) and accelerometer readings from an Empatica E4 wristband \cite{mccarthy2016validation} to detect stress events. Sensitive attributes include biological \textit{sex} and the \textit{hand} (left vs. right) on which the device was worn.

\textbf{PERCEPT-R (Clinical Audio):} We extend our evaluation to the digital health domain using a publicly available subset of the PERCEPT corpus (v2.2.2) \cite{benway2022percept}. The dataset consists of single-word speech recordings from children with speech sound disorders involving the rhotic /\rotatebox[origin=c]{180}{r}/ sound. In collaboration with clinical experts, we curated a cohort of $76$ participants and used expert-provided rhoticity annotations (0 = derhotic, 1 = fully rhotic). \textit{Sex} is treated as the sensitive attribute to assess the equity of the clinical classification model across demographic groups. Further details on the datasets and FL client distributions for FL evaluations can be found in Appendix~\ref{Appendix:data_label}.
}

\subsection{\textcolor{black}{Model Architectures and Hyper-parameters}}
\label{sec:models_hyperparams}

{\color{black}
We employ task-specific benchmark model architectures tailored to the computational constraints of edge devices and the specific modalities of each dataset.

\textbf{HAR and Stress Sensing (MLPs):} For the WIDAR, HugaDB, and Stress Sensing datasets, we adapt established benchmark architectures~\cite{yang2023benchmark,sharma2022psychophysiological,pirttikangas2006feature}. These are implemented as Multi-Layer Perceptrons (MLPs) with varying depth (2--3 layers) and regularization techniques (Dropout, Batch/Instance Normalization) optimized for their respective feature sets.

\textbf{Clinical Audio (CNN):} For the PERCEPT-R dataset, we utilized a Convolutional Neural Network (CNN) model \cite{benway2022percept} to process the temporal speech sequences. The network comprises two 1D convolutional layers (kernel size 5) that extract temporal features from the 5-channel input. The output is flattened and passed through a three-layer MLP classifier employing Hardswish activations to predict rhoticity.

\textbf{Hyperparameters:} We maintain consistent training budgets i.e. total number of FL rounds (R) and Local client training epochs (E) across all methods to ensure fair comparison. Table~\ref{tab:hyperparameters} details the specific hyperparameter settings for all four datasets. For the standard FedAvg, KD-Fedavg and FedSAM  baselines, we utilize the "Base Learning Rate'' and constant averaging. For \textbf{\textit{CurvFed}} and FedSWA, we introduce the SWA components (SWA LR, $\alpha$, Cycle). Notably, the PERCEPT-R dataset requires a distinct configuration (lower $\alpha$, longer training) due to the higher complexity of the audio modality. Further details on grid search  are provided in Appendix~\ref{appendix:reproducibility}.

}

\begin{table}[h!]
\centering
\color{black}
\resizebox{0.7\linewidth}{!}{
\begin{tabular}{l|c|c|c|c}
\hline
\textbf{Hyperparameter} & \textbf{WIDAR} & \textbf{Stress Sensing} & \textbf{HugaDB} & \textbf{PERCEPT-R} \\
\hline
\multicolumn{5}{l}{\textit{Training Budget}} \\
\hline
Local Epochs ($E$) & 3 & 3 & 3 & 5 \\
Total Rounds ($R$) & 80 & 80 & 80 & 100 \\
\hline
\multicolumn{5}{l}{\textit{Optimization (CurvFed / Baseline)}} \\
\hline
Base Learning Rate ($\eta$) & 0.01 / 0.1 & 0.01 / 0.1 & 0.001 / 0.001 & 0.001 / 0.001 \\
SWA Learning Rate ($\eta_{swa}$) & 0.1 & 0.001 & 0.001 & 0.001 \\
\hline
\multicolumn{5}{l}{\textit{Curvature Constraints (CurvFed)}} \\
\hline
Curvature Penalty ($\alpha$) & 0.92 & 0.92 & 0.92 & 0.8 \\
Cycle Length ($c$) & 5 & 5 & 3 & 5 \\
SWA Start Round & 16 & 16 & 16 & 20 \\
Stability Constant ($\epsilon$) (Eq \ref{agg-combine-weight_eqn}) & 0.005 & 0.005 & 0.005 & 0.005 \\
\hline
\end{tabular}
}
\caption{\textcolor{black}{Hyperparameter configuration for main experiments. \textbf{\textit{CurvFed}} introduces specific parameters (SWA, $\alpha$) to manage curvature. All methods share the same epoch and round budget to ensure fair comparison.}}
\label{tab:hyperparameters}
\vspace{-2em}
\end{table}


\subsection{Evaluation on \textbf{\textit{CurvFed}}'s Effectiveness}\label{eval-results}
\label{sec:rq1}

\textcolor{black}{To establish the effectiveness and stability of \textbf{\textit{CurvFed}}, we conducted a comprehensive evaluation using \textbf{\textit{stratified 5-fold cross-validation}} (maintaining an 80:20 train-test split per client), repeated across \textit{three} distinct random seeds. To maintain clarity, we present the mean and standard deviation for both the F1 score and the EO gap, averaged across folds for a representative seed. Furthermore, we report the FATE score, which is calculated using the mean F1 and mean EO values as defined in Equation~\ref{eq:fate}. Complete results for the additional seeds are provided in Appendix~\ref{appendix:reproducibility}. Our analysis is structured around the following research questions (RQs), each targeting a different aspect of fairness and efficacy in FL:}

\begin{itemize}
    \item \textbf{RQ-1:} Could \textbf{\textit{CurvFed}} improve the fairness--efficacy balance compared to contemporary benign (i.e., non-fairness-aware) FL approaches or FWD FL baselines?
    \item \textbf{RQ-2:} Can fairness be achieved in FL settings that include both \textit{single-person} and \textit{multi-person} clients?
    \item \textbf{RQ-3:} Can \textbf{\textit{CurvFed}} enhance fairness specifically for \textit{single-person} clients?
    \item \textbf{RQ-4:} Could \textbf{\textit{CurvFed}} improve fairness across \textit{multiple sensitive attributes simultaneously?}
\end{itemize}
The following sections present a detailed analysis of each question.

\subsubsection{RQ-1: Could \textbf{\textit{CurvFed}} improve the fairness-efficacy balance compared to contemporary benign FL approaches or FWD FL baselines?}

\textcolor{black}{We evaluated whether \textbf{\textit{CurvFed}} improves the trade-off between predictive performance and fairness relative to existing federated learning baselines. Fairness is measured using the FATE score and the Equal Opportunity (EO) gap with respect to \textit{sex}.}

\begin{table}[h!]
\centering
\color{black}
\resizebox{0.7\linewidth}{!}{
\begin{tabular}{|c|c|c|c|c|}
\hline
\textbf{Dataset} & \textbf{Model} &
\makecell{\textbf{F1 Score}\\ \textbf{Mean $\pm$ Std} ($\uparrow$)} &
\makecell{\textbf{EO Gap}\\ \textbf{Mean $\pm$ Std} ($\downarrow$)} &
\makecell{\textbf{FATE $\times 10^{-1}$}\\ \textbf{Mean Score} ($\uparrow$)} \\ 
\hline

\multirow{6}{*}{WIDAR}            & FedAvg     & $\textbf{0.862} \pm 0.0006$ & $0.412 \pm 0.0154$ & $0.0000$ \\ \cline{2-5}
                 & FedSAM            & $0.843 \pm 0.0108$ & $0.401 \pm 0.0230$ & $0.056$ \\ 
                 & FedSWA           & $0.859 \pm 0.0030$ & $0.405 \pm 0.0142$ & $0.147$           \\ 
                 & FedSRCVaR           & $0.724 \pm 0.0187$ & $0.345 \pm 0.0534$ & $0.019$          \\ 
                 & KD-Fedavg                 & $0.849 \pm 0.0123$ & $0.392 \pm 0.0299$ & $0.327$            \\ 
                 & CurvFed             & $0.806 \pm 0.0458$ & $\textbf{0.354} \pm 0.0099$ & $\textbf{0.741}$             \\ \hline
Stress sensing    & FedAvg          & $0.767 \pm 0.0146$ & $0.3240 \pm 0.014$           & 0.000             \\ \cline{2-5} 
                 & FedSAM            & $0.784 \pm 0.0094$ & $0.3135 \pm 0.0103$ & $0.548$            \\ 
                 & FedSWA            & $0.743 \pm 0.0164$ & $0.305 \pm 0.0034$ & $0.257$           \\ 
                 & FedSRCVaR           & $0.694 \pm 0.0253$ & $0.278 \pm 0.0231$ & $0.458$            \\ 
                 & KD-Fedavg                & $0.717 \pm 0.0157$ & $0.259 \pm 0.0095$ & $1.344$            \\ 
                 & CurvFed             & $\textbf{0.789} \pm 0.0086$ & $\textbf{0.259} \pm 0.0045$ & $\textbf{2.292}$              \\ \hline
HugaDB              & FedAvg           & $0.852 \pm 0.0063$ & $0.482 \pm 0.0294$ & 0.0000             \\ \cline{2-5} 
                 & FedSAM     & $0.786 \pm 0.0034$ & $0.438 \pm 0.0042$ & $0.014$             \\ 
                 & FedSWA            & $0.809 \pm 0.0033$ & $0.439 \pm 0.0144$ & $0.040$           \\ 
                 & FedSRCVaR           & $0.862 \pm 0.0032$ & $0.410 \pm 0.0103$ & $0.163$             \\ 
                 & KD-Fedavg                 & $0.776 \pm 0.0061$ & $0.444 \pm 0.0161$ & $-0.008$            \\ 
                 & CurvFed              & $\mathbf{0.892} \pm 0.0072$ & $\mathbf{0.406} \pm 0.0042$ & $\mathbf{0.206}$             \\ \hline

Percept-R              & FedAvg            & \textbf{0.7774} $\pm$ 0.002          & 0.162 $\pm$ 0.005           & 0.0000             \\ \cline{2-5} 
                 & FedSAM            & 0.777 $\pm$0.002            & 0.136 $\pm$ 0.01           & 1.72             \\ 
                 & FedSWA            & 0.777 $\pm$ 0.002          & 0.127 $\pm$ 0.01 & 1.66              \\ 
                 & FedSRCVaR            & 0.771 $\pm$ 0.004            & 0.130 $\pm$ 0.01           & 1.73            \\ 
                 & KD-Fedavg                 & 0.777 $\pm$ 0.001           & 0.150 $\pm$ 0.01           & 0.76             \\ 
                 & CurvFed             & 0.764 $\pm$ 0.003            & \textbf{0.106} $\pm$ 0.01           & \textbf{3.29}             \\ \hline
\end{tabular}
}
\caption{\color{black} Fairness--efficacy balance comparisons for RQ-1. Results are averaged over 5-fold cross-validation. Metrics with $\uparrow$ indicate higher is better, while metrics with $\downarrow$ indicate lower is better. CurvFed attains the highest FATE score, reflecting the overall fairness--utility balance}

\label{tab:model_performance_main}
\vspace{-15pt}
\end{table}

FedAvg is the fairness-unaware baseline, while FedSAM and FedSWA are benign baselines that promote flatter loss landscapes via client-side SAM and server-side SWA \cite{caldarola2022improving}. 


\emph{FL Client composition reflecting real-world disparities:} WIDAR and HugaDB clients typically include four males and one female (with one HugaDB client having two males and one female), while Stress Sensing shows greater variation (e.g., 2:4 to 3:6 male:female ratios). Unless otherwise noted, this distribution is used throughout. 

\emph{Evaluation discussion: }\textcolor{black}{As shown in Table~\ref{tab:model_performance_main}, results are reported as the mean $\pm$ standard deviation over 5-fold cross-validation. \textbf{\textit{CurvFed}} consistently improves the fairness–efficacy trade-off across all datasets. In WIDAR, it reduces the EO gap from $0.412 \pm 0.0154$ (FedAvg) to $0.354 \pm 0.0099$, with a moderate F1 reduction ($0.806 \pm 0.0458$ vs. $0.862 \pm 0.0006$). In Stress Sensing, it improves both F1 ($0.789 \pm 0.0086$ vs. $0.767 \pm 0.0146$) and EO gap ($0.259 \pm 0.0045$ vs. $0.324 \pm 0.014$). HugaDB shows similar trends, with a higher F1 ($0.892 \pm 0.0072$) and reduced EO gap ($0.406 \pm 0.0042$) compared to FedAvg ($0.852 \pm 0.0063$, $0.482 \pm 0.0294$). In Percept-R, where baseline models exhibit comparable F1 performance, \textbf{\textit{CurvFed}} achieves the lowest EO gap ($0.106 \pm 0.01$) with only a small reduction in F1 ($0.764 \pm 0.003$ vs. $0.777 \pm 0.002$ for FedAvg).}

\textcolor{black}{\textbf{\textit{CurvFed}} consistently outperforms KD-FedAvg and FedSRCVaR in FATE scores across all datasets—Stress Sensing ($2.292$ vs. $1.344$ and $0.458$), WIDAR ($0.741$ vs. $0.327$ and $0.019$), HugaDB ($0.206$ vs. $-0.008$ and $0.163$), and Percept-R ($3.29$ vs. $0.76$ and $1.73$). While KD-based methods excel in centralized settings with diverse group-wise data \cite{chai2022fairness}, their local fairness improvements do not reliably carry over after model aggregation in FL \cite{hamman2023demystifying}. This limits their effectiveness in FL settings.}

\textcolor{black}{FedSRCVaR, relies on a constraint $\rho$ to bound worst-case group sizes \cite{papadaki2024federated}, assumes known group distributions—often unavailable in human-centric data with missing or highly variable groups across clients. This results in unstable optimization and poor fairness–utility trade-offs, as reflected by the reduced F1 score ($0.694 \pm 0.0253$) on the Stress Sensing dataset. In contrast, \textbf{\textit{CurvFed}} makes no such assumptions and achieves better fairness–performance balance.}


\subsubsection{RQ-2: Can fairness be achieved in FL settings that include both \textit{single-person} and \textit{multi-person} clients?}

Traditional fairness-aware FL methods assume each client contains data from all sensitive groups. However, as noted in \textcolor{black}{Section \ref{intro-label} challenge 3}, edge devices may contain data from a single user or multiple users. Thus, this assumption often fails in real-world deployments. Notably, \textbf{\textit{CurvFed}} removes this assumption by promoting fairness through loss landscape regularization, making it impactful for practical deployment in human-sensing applications.
\textcolor{black}{ Table~\ref{tab:combined_model_performance_client_single_multi} presents \textbf{\textit{CurvFed}}’s performance across all four datasets, under a mix of single- and multi-person clients with 1–2 males and 0–1 females per client (See Appendix \ref{appendix:client} for detailed client distributions). Despite the absence of full group representation, \textbf{\textit{CurvFed}} consistently outperforms other FL methods, achieving strong FATE scores of $0.479 \times 10^{-1}$ on WIDAR, $4.032 \times 10^{-1}$ on Stress Sensing, $0.590 \times 10^{-1}$ on HugaDB, and $2.03 \times 10^{-1}$ on Percept-R, demonstrating balanced fairness and performance under practical constraints.}



\subsubsection{RQ-3: Can \textbf{\textit{CurvFed}} enhance fairness specifically for \textit{single-person} clients?}

Only single-person clients are common in human-centric FL but pose challenges due to the lack of intra-client diversity, making global fairness and performance difficult to achieve. Prior studies \cite{lu2023fedcom,yao2024rethinking} show that simple aggregation methods like FedAvg often degrade global performance in these cases \cite{lu2023fedcom}.

\FloatBarrier
\begin{table}[H]
\color{black}
\centering
\resizebox{0.8\linewidth}{!}{
\begin{tabular}{|c|c|c|c|c|c|}
\hline
\textbf{Dataset} & \textbf{Client Type} & \textbf{Model} &
\makecell{\textbf{F1 Score}\\ \textbf{Mean $\pm$ Std} ($\uparrow$)} &
\makecell{\textbf{EO Gap}\\ \textbf{Mean $\pm$ Std} ($\downarrow$)} &
\makecell{\textbf{FATE $\times 10^{-1}$}\\ \textbf{Mean} ($\uparrow$)} \\
\hline

\multirow{12}{*}{WIDAR}
& \multirow{6}{*}{\shortstack{Single \& \\Multi-Person}}
& FedAvg     & $0.836 \pm 0.0132$ & $0.429 \pm 0.0081$ & $0.000$ \\ \cline{3-6}
&  & FedSAM     & $0.822 \pm 0.0143$ & $0.411 \pm 0.0032$ & $0.238$ \\
&  & FedSWA     & $\mathbf{0.852} \pm 0.0323$ & $0.431 \pm 0.0034$ & $0.140$ \\
&  & KD-FedAvg  & $0.781 \pm 0.0030$ & $0.389 \pm 0.0013$ & $0.270$ \\
&  & FedSRCVaR  & $0.712 \pm 0.0048$ & $0.405 \pm 0.0112$ & $-0.929$ \\
&  & CurvFed    & $0.791 \pm 0.0089$ & $\mathbf{0.385} \pm 0.0089$ & $\mathbf{0.479}$ \\ \cline{2-6}

& \multirow{6}{*}{\shortstack{Single \\Person Only}}
& FedAvg     & $\mathbf{0.843} \pm 0.0115$ & $0.441 \pm 0.0143$ & $0.000$ \\ \cline{3-6}
&  & FedSAM     & $0.839 \pm 0.0053$ & $0.423 \pm 0.0098$ & $0.366$ \\
&  & FedSWA     & $0.842 \pm 0.0093$ & $0.424 \pm 0.0034$ & $0.383$ \\
&  & FedSRCVaR  & $0.706 \pm 0.0124$ & $0.418 \pm 0.0075$ & $-1.094$ \\
&  & KD-FedAvg  & $0.771 \pm 0.0119$ & $0.402 \pm 0.0082$ & $0.030$ \\
&  & CurvFed    & $0.793 \pm 0.0092$ & $\mathbf{0.382} \pm 0.0045$ & $\mathbf{0.754}$ \\
\hline

\multirow{12}{*}{Stress Sensing}
& \multirow{6}{*}{\shortstack{Single \& \\Multi-Person}}
& FedAvg     & $0.761 \pm 0.0143$ & $0.376 \pm 0.0023$ & $0.000$ \\ \cline{3-6}
&  & FedSAM     & $0.749 \pm 0.0034$ & $0.382 \pm 0.0120$ & $-0.313$ \\
&  & FedSWA     & $\mathbf{0.841} \pm 0.0145$ & $0.310 \pm 0.0114$ & $2.822$ \\
&  & KD-FedAvg  & $0.745 \pm 0.0129$ & $0.248 \pm 0.0024$ & $3.193$ \\
&  & FedSRCVaR  & $0.701 \pm 0.0083$ & $0.283 \pm 0.0215$ & $1.683$ \\
&  & CurvFed    & $0.829 \pm 0.0129$ & $\mathbf{0.258} \pm 0.0134$ & $\mathbf{4.032}$ \\ \cline{2-6}

& \multirow{6}{*}{\shortstack{Single \\Person Only}}
& FedAvg     & $0.762 \pm 0.0014$ & $0.382 \pm 0.0034$ & $0.000$ \\ \cline{3-6}
&  & FedSAM     & $0.769 \pm 0.0093$ & $0.338 \pm 0.0093$ & $1.235$ \\
&  & FedSWA     & $0.779 \pm 0.0182$ & $0.368 \pm 0.0076$ & $0.589$ \\
&  & FedSRCVaR  & $0.675 \pm 0.0214$ & $0.218 \pm 0.0142$ & $3.143$ \\
&  & KD-FedAvg  & $0.680 \pm 0.0313$ & $0.257 \pm 0.0117$ & $2.186$ \\
&  & CurvFed    & $\mathbf{0.731} \pm 0.0362$ & $\mathbf{0.235} \pm 0.0189$ & $\mathbf{3.447}$ \\
\hline
\multirow{12}{*}{HugaDB}
& \multirow{6}{*}{Single \& Multi-Person}
& FedAvg     & $\mathbf{0.856} \pm 0.0113$ & $0.441 \pm 0.0042$ & $0.000$ \\ \cline{3-6}
&  & FedSAM     & $0.847 \pm 0.0118$ & $0.408 \pm 0.0032$ & $0.636$ \\
&  & FedSWA     & $0.842 \pm 0.0092$ & $0.413 \pm 0.0092$ & $0.461$ \\
&  & KD-FedAvg  & $0.789 \pm 0.0068$ & $0.409 \pm 0.0045$ & $-0.048$ \\
&  & FedSRCVaR  & $0.827 \pm 0.0182$ & $0.403 \pm 0.0142$ & $0.524$ \\
&  & CurvFed    & $0.816 \pm 0.0340$ & $\mathbf{0.395} \pm 0.0083$ & $\mathbf{0.590}$ \\ \cline{2-6}

& \multirow{6}{*}{Single Person Only}
& FedAvg     & $\mathbf{0.871} \pm 0.0013$ & $0.454 \pm 0.0034$ & $0.000$ \\ \cline{3-6}
&  & FedSAM     & $0.856 \pm 0.0052$ & $0.502 \pm 0.0182$ & $-1.230$ \\
&  & FedSWA     & $0.783 \pm 0.0432$ & $0.432 \pm 0.0231$ & $-0.525$ \\
&  & FedSRCVaR  & $0.792 \pm 0.0092$ & $0.411 \pm 0.0092$ & $0.038$ \\
&  & KD-FedAvg  & $0.769 \pm 0.0082$ & $\mathbf{0.409} \pm 0.0048$ & $-0.180$ \\
&  & CurvFed    & $0.824 \pm 0.0054$ & $0.418 \pm 0.0092$ & $\mathbf{0.253}$ \\
\hline

 \multirow{8}{*}{Percept-R}  
 & \multirow{4}{*}{\shortstack{Single \& \\Multi-Person}} & FedAvg  & \textbf{0.785} $\pm$ 0.006  & 0.132 $\pm$ 0.01 & 0.0000 \\ \cline{3-6} 
     &  & FedSAM      & 0.781 $\pm$ 0.004  & 0.113 $\pm$ 0.01 & 1.31 \\
  &  & FedSWA      & 0.780 $\pm$ 0.004  & 0.117 $\pm$ 0.01 & 1.02 \\
 &  & KD-FedAvg      & 0.784 $\pm$ 0.006  & 0.132 $\pm$ 0.01 & -0.06 \\  
 &  & FedSRCVaR      & 0.783 $\pm$ 0.006  & 0.131 $\pm$ 0.009 & 0.03 \\ 
 &  & CurvFed          & 0.772 $\pm$ 0.003  & \textbf{0.102} $\pm$ 0.01  & \textbf{2.03} \\ \cline{2-6}
 & \multirow{4}{*}{\shortstack{Single \\Person Only}} & FedAvg  & \textbf{0.770} $\pm$ 0.005   & 0.108 $\pm$ 0.01 & 0.0000 \\ \cline{3-6} 
    &  & FedSAM      & 0.768 $\pm$ 0.006  & 0.115 $\pm$ 0.008  & -0.71 \\
  &  & FedSWA      & 0.765 $\pm$ 0.007  & 0.112 $\pm$ 0.008 & -0.45 \\
 &  & FedSRCVaR      & 0.770 $\pm$ 0.005  & 0.102 $\pm$ 0.01 & 0.5 \\ 
 &  & KD-FedAvg      & 0.770 $\pm$ 0.005  & 0.106 $\pm$ 0.01 & 0.17\\ 
 &  & CurvFed          & 0.764 $\pm$ 0.007  & \textbf{0.054} $\pm$ 0.004  & \textbf{4.85}  \\ \hline

\end{tabular}
}
\caption{\color{black} Performance comparison across different client setups: single \& multi-person clients (RQ2) and single-person clients (RQ3). Results are averaged over 5-fold cross-validation. Metrics with $\uparrow$ indicate higher is better, while metrics with $\downarrow$ indicate lower is better. \footnotesize\textit{ Note: While some methods perform better on individual F1 or EO Gap metrics, CurvFed attains the highest FATE score, reflecting the overall fairness--utility balance}
}
\label{tab:combined_model_performance_client_single_multi}
\vspace{-2em}
\end{table}

\textcolor{black}{We evaluated \textbf{\textit{CurvFed}} against other FWD baselines in only single-client scenarios. As shown in Table~\ref{tab:combined_model_performance_client_single_multi}, CurvFed consistently achieves the highest FATE scores across all datasets in this setting, including $0.754 \times 10^{-1}$ for WIDAR, $3.447 \times 10^{-1}$ for Stress Sensing, $0.253 \times 10^{-1}$ for HugaDB, and $4.85 \times 10^{-1}$ for Percept-R —demonstrating its effectiveness even without client-level sensitive attribute diversity.}



\subsubsection{RQ-4: Could \textbf{\textit{CurvFed}} improve fairness across \textit{multiple sensitive attributes simultaneously?}}

In FL-based human sensing, data often simultaneously involve multiple sensitive attributes (e.g., sex, age, body type, sensor placement). To assess \textbf{\textit{CurvFed}} under such conditions, we evaluated fairness with respect to additional sensitive attributes—hands’ (watch-hand) and orientation’ (sensor placement)—in the Stress Sensing and WIDAR datasets, using the same models as in Table~\ref{tab:model_performance_main}. For WIDAR, we induced disparity by subsampling 50\% of data from orientations 4 and 5, grouping orientations 1–3 as "major" and the rest as "minor." In Stress Sensing, only 13 of 48 participants had both-hand data, while others had left-hand only. HugaDB and Percept\_R, with only sex as a sensitive attribute, were excluded.

\textcolor{black}{As shown in Table~\ref{tab:metrics_sensitive_multiple}, \textbf{\textit{CurvFed}} consistently achieves the strongest fairness–utility trade-off across both sensitive attributes. Although other methods may attain higher F1 scores or lower EO gaps individually, \textbf{\textit{CurvFed}} maintains competitive classification performance while simultaneously reducing EO gaps, resulting in the highest FATE scores on both WIDAR and Stress Sensing. This indicates that \textbf{\textit{CurvFed}} does not optimize a single objective in isolation but instead preserves a balanced trade-off between fairness and accuracy when multiple sensitive attributes are present, thereby addressing Challenge~4 outlined in Section~\ref{intro-label}.}

\FloatBarrier
\begin{table}[H]
\centering
\color{black}
\resizebox{0.8\linewidth}{!}{
\begin{tabular}{|c|c|c|c|c|c|}
\hline
\textbf{Dataset} &
\makecell{\textbf{Sensitive}\\ \textbf{Attribute}} &
\textbf{Model} &
\makecell{\textbf{F1 Score}\\ \textbf{Mean $\pm$ Std}\\ ($\uparrow$)} &
\makecell{\textbf{EO Gap}\\ \textbf{Mean $\pm$ Std}\\ ($\downarrow$)} &
\makecell{\textbf{FATE $\times 10^{-1}$}\\ \textbf{Mean}\\ ($\uparrow$)} \\
\hline

\multirow{6}{*}{WIDAR}
& \multirow{6}{*}{Orientation}
& FedAvg     & $\mathbf{0.862} \pm 0.0006$ & $0.506 \pm 0.0161$ & $0.000$ \\ \cline{3-6}
&  & FedSAM     & $0.843 \pm 0.0108$ & $0.481 \pm 0.0146$ & $0.284$ \\
&  & FedSWA     & $0.859 \pm 0.0030$ & $0.499 \pm 0.0499$ & $0.102$ \\
&  & FedSRCVaR  & $0.725 \pm 0.0187$ & $0.422 \pm 0.0535$ & $0.060$ \\
&  & KD-Fedavg  & $0.849 \pm 0.0123$ & $0.458 \pm 0.0343$ & $0.793$ \\
&  & CurvFed    & $0.806 \pm 0.0458$ & $\mathbf{0.395} \pm 0.0507$ & $\mathbf{1.550}$ \\
\hline

\multirow{6}{*}{Stress Sensing}
& \multirow{6}{*}{Hand}
& FedAvg     & $0.767 \pm 0.0146$ & $0.464 \pm 0.0124$ & $0.000$ \\ \cline{3-6}
&  & FedSAM     & $0.785 \pm 0.0094$ & $0.463 \pm 0.0830$ & $0.233$ \\
&  & FedSWA     & $0.744 \pm 0.0164$ & $0.441 \pm 0.0234$ & $0.173$ \\
&  & FedSRCVaR  & $0.695 \pm 0.0253$ & $\mathbf{0.394} \pm 0.0211$ & $0.550$ \\
&  & KD-Fedavg  & $0.718 \pm 0.0157$ & $0.406 \pm 0.0193$ & $0.587$ \\
&  & CurvFed    & $\mathbf{0.789} \pm 0.0086$ & $0.433 \pm 0.0023$ & $\mathbf{0.945}$ \\
\hline

\end{tabular}
}
\caption{\color{black} Fairness--efficacy balance under different sensitive attributes for RQ-4.Results are averaged over 5-fold cross-validation. Metrics with $\uparrow$ indicate higher is better, while metrics with $\downarrow$ indicate lower is better. \footnotesize\textit{ Note: While some methods perform better on individual F1 or EO Gap metrics, CurvFed attains the highest FATE score, reflecting the overall fairness--utility balance}
}
\label{tab:metrics_sensitive_multiple}
\vspace{-2em}
\end{table}

\section{\textcolor{black}{Feasibility Study}}
\label{Feasibility-Study}
\textcolor{black}{In this section, we conduct a rigorous empirical study of \textit{CurvFed} in real-world settings. We systematically evaluate its deployment feasibility (Section \ref{real-world-user-study}) with additional runtime benchmarking/overhead analysis detailed in Appendix \ref{benchmarking_old}. Furthermore, we discuss sensitivity to key factors (Sections \ref{Sensitivity-Analysis} and \ref{alpha}), and conclude with an ablation study that substantiates the effectiveness of the core design choices (Section \ref{ablation-study}).
}
{\color{black}
\subsection{Assessing Real-World Deployment Feasibility of CurvFed's Modules}
\label{real-world-user-study}

\begin{figure}[!ht]
\centering
\includegraphics[width=0.7\linewidth]{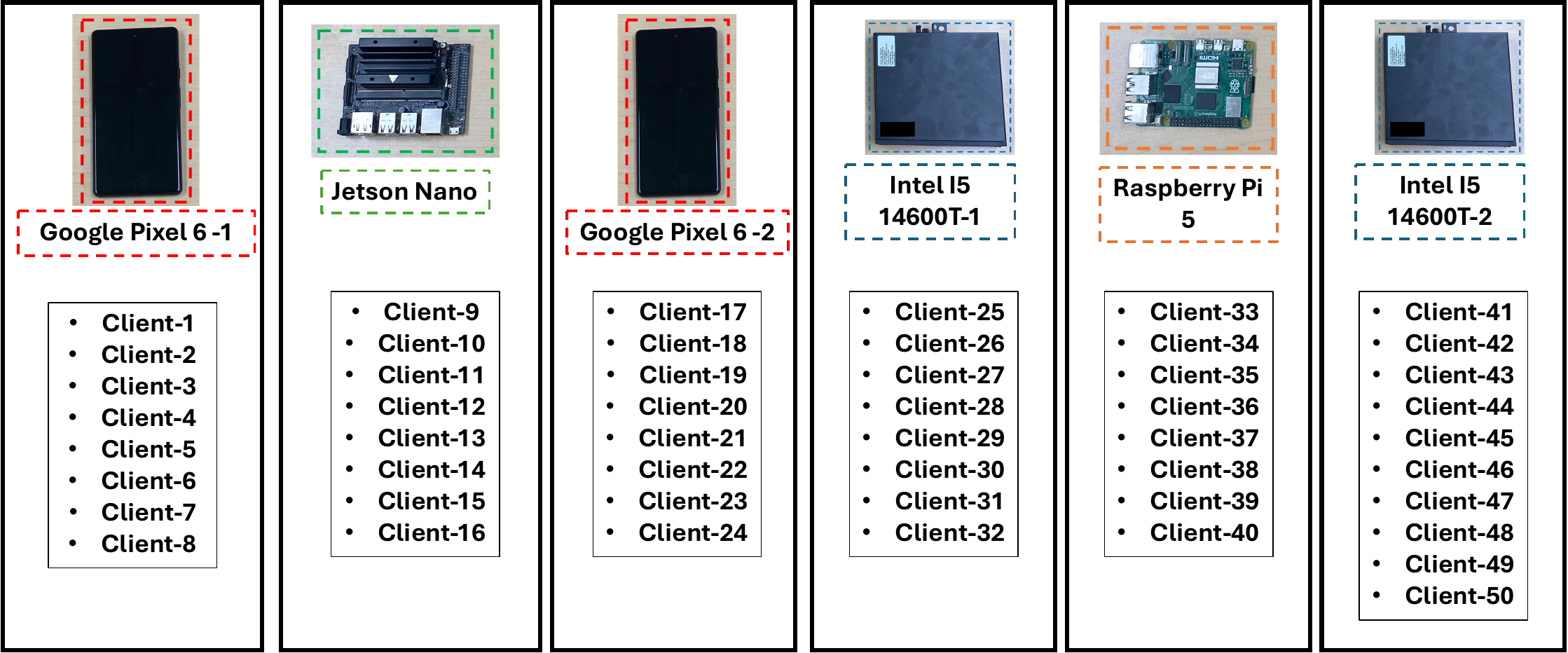}
\caption{\color{black} Experimental setup and client distribution for real-world FL evaluation using heterogeneous edge devices.}
\label{fig:fl_setup}
\end{figure}

To validate \textbf{\textit{CurvFed}}'s feasibility on heterogeneous edge hardware, we conducted a comprehensive runtime evaluation using a physical FL testbed. Our setup mapped the 48 participants from the Stress Sensing dataset (Section~\ref{dataset})—augmented with two randomized duplicates to standardize the cohort size at $N=50$ clients—to a heterogeneous cluster of six edge devices. This cluster included two Google Pixel 6 smartphones, a Raspberry Pi 5, an NVIDIA Jetson Nano, and two Intel i5-14600T desktops (Fig.~\ref{fig:fl_setup}). Each device executed distinct, isolated client workloads to simulate realistic deployment conditions, training local models and exchanging updates with a central server via Google Cloud APIs authenticated through PyDrive. The following subsections analyze the computational overhead incurred by this specific 50-client configuration, first examining the client-side training costs and subsequently evaluating the server-side aggregation performance.

\subsubsection{Client-Side Computational Overhead Analysis:} We focus first on the local training costs incurred by the heterogeneous edge devices. Table~\ref{tab:realtime_stats} breaks down the computational overhead of our approach compared to standard (\textit{Benign}) FL training and an ablation variation without the SAM optimizer, just using Equation \ref{local_training_eq}. The results confirm that \textbf{CurvFed} imposes minimal overhead, remaining well within the operational limits of modern edge platforms.

\begin{itemize}
    \item \textbf{Training Latency:} The results from Table~\ref{tab:realtime_stats} show that, although curvature regularization ($\lambda(F_{\text{client}})$ penalty) and SAM introduce additional gradient computations, the resulting latency increase is modest. Even on resource-constrained devices such as the Jetson Nano, average per-client training time remains below $0.8$s. On more capable hardware, our complete \textbf{\textit{CurvFed}} module shows average-per-client latency of about ($\approx 0.2$s). The overhead of \textbf{\textit{CurvFed}} compared to the benign variant across devices remains small $\approx 0.2-0.4s$ and about $\approx 0.1-0.3s$ when compared to the ablation variant without SAM, rendering it insignificant relative to typical network communication delays (Appendix Table \ref{tab:network_latency})
    
    \item \textbf{Resource Utilization:} Memory usage remained stable, with \textit{CurvFed} incurring $<1\%$ additional RAM overhead compared to the others. CPU utilization varied by architecture (higher on Raspberry Pi, lower on Desktop) but stayed within safe operating ranges, avoiding utilization spikes that could degrade device stability or battery life.
    
    \item \textbf{Communication Stability:} Upload latencies were consistent across all methods ($\approx 1.6\text{--}2.9$s), confirming that our optimization objective does not inflate model checkpoints or hinder transmission reliability.
\end{itemize}

\begin{table}[h!]
    \centering
    \color{black}
    \resizebox{0.95\linewidth}{!}{%
    \begin{tabular}{|c|c|c|c|c|c|c|}
        \hline
        \textbf{Client Assignments} & \textbf{Device} & \textbf{Method} & \makecell{\textbf{Avg Training} \\ \textbf{Time (s)} $\pm$ std}   &  \makecell{\textbf{Avg. Memory}  \\ \textbf{(MB)}} & \makecell{\textbf{Avg. Train} \\ \textbf{Usage (\%)}} & \makecell{\textbf{Avg. Upload} \\ \textbf{Time (s)}}  \\
        \hline
        \multirow{3}{*}{Client 1-8} & \multirow{3}{*}{Google Pixel 6} & Benign Training & 0.28 $\pm$ 0.02 & 292.78 & 27.38 & 2.6402 \\ \cline{3-7}
                                   &                                 & Eq \ref{local_training_eq} & 0.42 $\pm$ 0.16 & 296.19 & 26.67 & 2.3620 \\ \cline{3-7}
                                   &                                 & CurvFed & 0.43 $\pm$ 0.05 & 295.64 & 27.40 & 2.3574 \\
        \hline
        \multirow{3}{*}{Client 9-16} & \multirow{3}{*}{Jetson Nano} & Benign Training & 0.25 $\pm$ 0.10 & 321.07 & 3.20 & 1.76 \\ \cline{3-7}
                                   &                                 & Eq \ref{local_training_eq} & 0.35 $\pm$ 0.12 & 321.59 & 3.90 & 1.8248 \\ \cline{3-7}
                                   &                                 & CurvFed & 0.77 $\pm$ 0.21 & 321.08 & 5.45 & 1.6292 \\
        \hline
        \multirow{3}{*}{Client 17-24} & \multirow{3}{*}{Google Pixel 6} & Benign Training & 0.30 $\pm$ 0.02 & 291.99 & 27.48 & 2.2517 \\ \cline{3-7}
                                   &                                 & Eq \ref{local_training_eq} & 0.38 $\pm$ 0.05 & 296.49 & 27.22 & 2.2923 \\ \cline{3-7}
                                   &                                 & CurvFed & 0.47 $\pm$ 0.06 & 296.09 & 27.66 & 2.9414 \\
        \hline
        \multirow{3}{*}{Client 25-32} & \multirow{3}{*}{Intel i5-14600T} & Benign Training & 0.12 $\pm$0.02 & 285.64 & 2.64 & 2.0143 \\ \cline{3-7}
                                   &                                 & Eq \ref{local_training_eq} & 0.24 $\pm$ 0.02 & 286.55 & 2.73 & 1.9582 \\ \cline{3-7}
                                   &                                 & CurvFed & 0.27 $\pm$ 0.02 & 286.96 & 3.02 & 2.0142 \\
        \hline
        \multirow{3}{*}{Client 33-40} & \multirow{3}{*}{Raspberry Pi} & Benign Training & 0.24 $\pm$ 0.04 & 312.65 & 42.7 & 1.9864 \\ \cline{3-7}
                                   &                                 & Eq \ref{local_training_eq} & 0.39 $\pm$ 0.09 & 314.57 & 42.50 & 1.9634 \\ \cline{3-7}
                                   &                                 & CurvFed & 0.37 $\pm$ 0.02 & 314.65 & 42.69 & 2.0028 \\
        \hline
        \multirow{3}{*}{Client 40-50} & \multirow{3}{*}{Intel i5-14600T} & Benign Training & 0.15 $\pm$ 0.06 & 285.65 & 2.64 & 1.937 \\ \cline{3-7}
                                   &                                 & Eq \ref{local_training_eq} & 0.30 $\pm$ 0.06 & 286.55 & 3.29 & 1.9204 \\ \cline{3-7}
                                   &                                 & CurvFed & 0.26 $\pm$ 0.02 & 286.95 & 2.71 & 1.9586 \\
        \hline
    \end{tabular}%
    }
    \caption{\color{black} Client side training statistics (averaged) from the real-world FL deployment. Metrics include training time (mean $\pm$ std), Process RSS memory (MB), average CPU utilization (\%), and model upload latency (s).}
    \label{tab:realtime_stats}
    \vspace{-6pt}
\end{table}

\subsubsection{Server-Side Computational Overhead Analysis:} To validate the deployment feasibility of \textbf{CurvFed} on standard aggregation servers \cite{mcmahan2017communication}, we profiled the computational cost of our server level design choices, i.e.,  Stochastic Weight Averaging (SWA) update step relative to the standard Federated Averaging (FedAvg) aggregation. It is important to note that \textbf{\textit{CurvFed}}’s sharpness-aware weighted aggregation scheme (Equation~\ref{agg-combine-weight_eqn}) follows the same aggregation procedure as FedAvg, differing only in the choice of weighting coefficients. These weights encode client curvature and utility information ($W^{r}$); consequently, the server-side runtime overhead remains identical to FedAvg.
We utilized a high-performance workstation equipped with an AMD Ryzen 9 7950X CPU and an NVIDIA GeForce RTX 4090 GPU to simulate a realistic server environment handling 50 concurrent client updates. Table~\ref{tab:agg_perf} summarizes the Time, Memory, and Hardware Utilization for aggregating 50 client models (FedAvg) versus updating the global SWA model.

Our profiling reveals that the SWA update step incurs \textit{negligible computational overhead} compared to the FedAvg aggregation.

\begin{itemize}
    \item \textbf{Aggregation (FedAvg):} The standard FedAvg aggregation of 50 client models requires iterating through all client parameters, resulting in a processing time of approximately $1\text{ms}$ on CPU and $5\text{ms}$ on GPU. The slightly higher latency on the GPU is attributed to the PCIe data transfer\cite{van2014performance} overhead dominating the arithmetic operations for these model sizes~\cite{mcmahan2017communication}.

    \item \textbf{SWA Update:} In contrast, the SWA update involves a simple weighted moving average between only two sets of parameters. This operation is mathematically $O(1)$ (Line $21$ in Algorithm \ref{alg:algo-1}), whereas FedAvg is $O(K)$ relative to the number of clients $K$ (Line $18$ in Algorithm \ref{alg:algo-1}). Consequently, the SWA update time was unmeasurable ($<1\text{ms}$). The increased GPU utilization (32.0\%) confirms better efficiency: unlike FedAvg, which frequently leaves the GPU idle to wait for loading client model, SWA performs continuous computation on data already in memory~\cite{williams2009roofline}.
    
\end{itemize}

These results demonstrate that integrating SWA into the server-side workflow does not introduce \textit{any latency overhead}. The computational bottleneck remains the aggregation of client updates and network communication, not the SWA optimization. This justifies our design choice to leverage SWA for improving generalization~\cite{izmailov2018averaging} without compromising the real-time constraints of ubiquitous computing environments.

}

\begin{table}[h!]
    \centering
    \color{black}
    \resizebox{0.6\linewidth}{!}{%
    \begin{tabular}{|c|c|c|c|c|c|}
        \hline
        \textbf{Platform} & \textbf{Method} & \textbf{Time (s)} & \textbf{Peak Mem (MB)} & \textbf{CPU (\%)} & \textbf{GPU (\%)} \\
        \hline
        \multirow{2}{*}{CPU} & FedAvg & 0.001 & 511.477 & 3.873 & -- \\
                                   & SWA Update & $\approx$ 0.000 & 512.477 & 5.300 & -- \\ \hline
        \multirow{2}{*}{GPU} & FedAvg & 0.005 & 638.742 & 4.205 & 26.892 \\
                                   & SWA Update & $\approx$ 0.000 & 638.742 & 2.600 & 32.000 \\ \hline
    \end{tabular}%
    }
    \caption{\color{black} Aggregation Performance Comparison (Avg 50 Clients). \textbf{CPU:} AMD Ryzen 9 7950X 16-Core Processor, \textbf{GPU:} NVIDIA GeForce RTX 4090}
    \label{tab:agg_perf}
\vspace{-2em}
\end{table}


\subsubsection{\textbf{Communication Overhead: }}
\label{comm_overhead}
We compare \textbf{\textit{CurvFed}}’s communication cost with FL baselines: Fedavg, KD-Fedavg and FedSWA approaches exchange 34.38MB (WIDAR) and 0.042MB (Stress Sensing) per round. As overhead, \textbf{\textit{CurvFed}} adds only 8 bytes for sharpness-aware aggregation, while FedSRCVaR adds 4 bytes for a risk threshold—both negligible relative to their fairness benefits.

\subsection{\textbf{Sensitivity Analysis}}  
\label{Sensitivity-Analysis}
In this section, we explore the performance and practicality of \textbf{\textit{CurvFed}} under the following real-world scenarios:

\begin{enumerate}
    \item \textbf{Effect of Ordered Sampling on FL Performance}
    \item \textbf{Effect of Subsampling Local Labeled Data and Clients on Global Fairness}
    \item \textbf{Impact of Random Client Participation on Fairness Across Models}
\end{enumerate}

Here, we focus on the WIDAR and Stress Sensing datasets using \textit{sex} as the sensitive attribute. They were chosen for their contrasting modalities: WIDAR is uni-modal, while Stress Sensing is multi-modal. While FedSAM and FedSWA were previously compared for benign generalization, they are excluded here to emphasize fairness-oriented baselines: FedAvg serves as the primary benign baseline, and KD-FedAvg and FedSRCVaR serve as FWD FL baselines.

\subsubsection{\textbf{Effect of Ordered Sampling on FL Performance:}}  
Human-sensing data often contains strong temporal dependencies \cite{lymberopoulos2009methodology,li2025survey}, making it important to preserve data order during local training \cite{liu2024personalized,mashhadi2022auditing}. It also mimics real-world FL deployments \cite{wu2024spatio}. To reflect this, we evaluate FL performance using an ordered 80:20 train-test split without shuffling.
As shown in Table~\ref{tab:performance_comparison_8020}, \textbf{\textit{CurvFed}} achieves the best fairness–utility tradeoff, reducing the EO gap and achieving the highest F1 and FATE scores, demonstrating its effectiveness in temporally structured FL settings.

\begin{table}[h!]
\centering
\resizebox{0.6\linewidth}{!}{
\begin{tabular}{|c|c|c|c|c|}
\hline
\textbf{Dataset} & \textbf{Model} & \textbf{F1 Score $\uparrow$} & \textbf{EO Gap $\downarrow$} & \textbf{FATE Score $\times 10^{-1}$ $\uparrow$} \\
\hline
\multirow{4}{*}{WIDAR} 
    & FedAvg      & \textbf{0.8570} & 0.4629 & 0.0000 \\ \cline{2-5}
    & FedSRCVaR   & 0.7676 & 0.4002 & 0.311 \\
    & KD-FedAvg   & 0.8135 & 0.4069 & 0.701 \\
    & CurvFed     & 0.7488 & \textbf{0.3452} & \textbf{1.280} \\
\hline
\multirow{4}{*}{Stress Sensing} 
    & FedAvg      & 0.7482 & 0.4000 & 0.0000 \\ \cline{2-5}
    & FedSRCVaR   & 0.6989 & 0.3633 & 0.259 \\
    & KD-FedAvg   & 0.7326 & 0.3633 & 0.709 \\
    & CurvFed     & \textbf{0.7693} & \textbf{0.2938} & \textbf{2.935} \\
\hline
\end{tabular}
}
\caption{\color{black}{
Performance comparison under ordered 80:20 train--test splits. CurvFed achieves the lowest EO Gap and highest FATE score, indicating improved fairness and a superior fairness–utility trade-off, while maintaining competitive F1 performance.
}}

\label{tab:performance_comparison_8020}
\vspace{-15pt}
\end{table}

 \subsubsection{\textbf{Effect of Subsampling of Local Labeled Data and Clients on Global Fairness:}}
In real-world edge deployments, clients often operate under resource constraints, leading to limited local data and sporadic client availability \cite{li2021fedmask}. To simulate this, we evaluate \textbf{\textit{CurvFed}} under (i) varying local training data availability (20\%, 50\%, 90\%), and (ii) fluctuating client participation per round (3, 5, 7 clients), with a fixed 20\% test set.

\begin{figure}[h!]
    \centering
    \begin{minipage}[b]{0.7\linewidth}
        \centering
        \includegraphics[width=\textwidth]{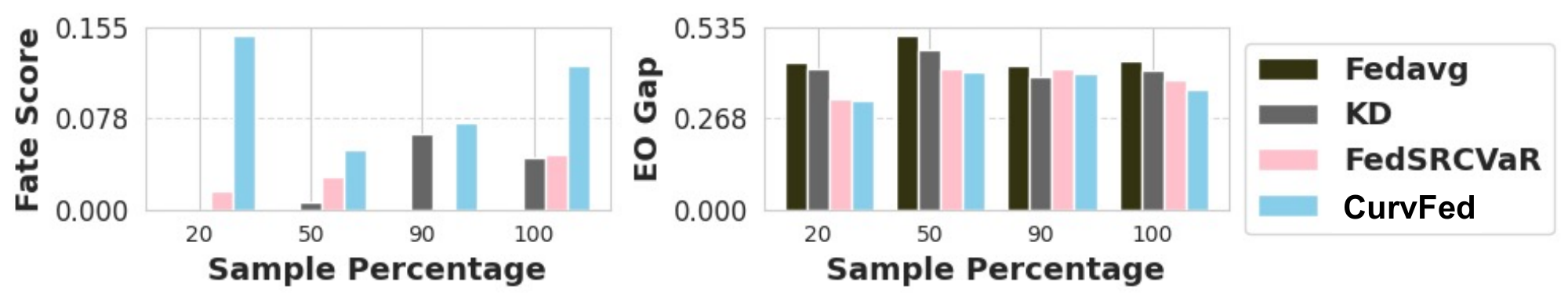}
        \caption*{(a) \color{black} WIDAR Dataset}
        \label{fig:WIDAR_samples}
    \end{minipage}
    
    \begin{minipage}[b]{0.7\linewidth}
        \centering
        \includegraphics[width=\textwidth]{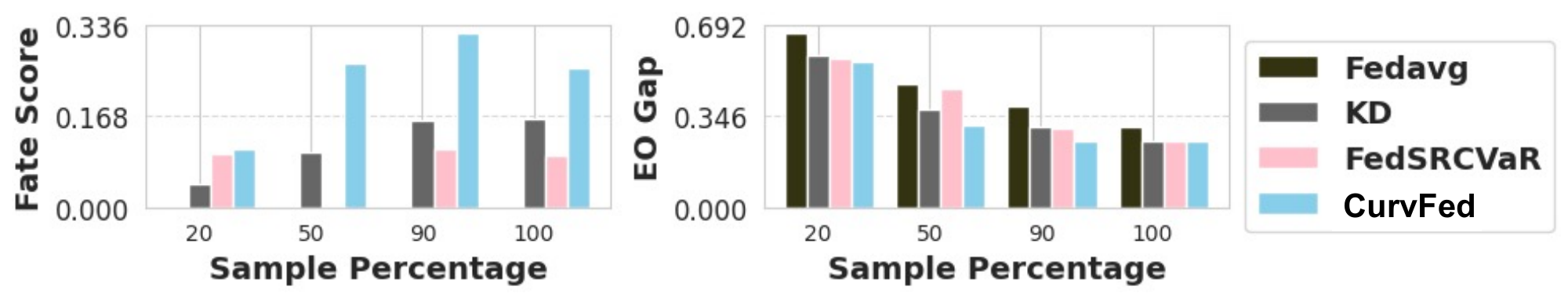}
        \caption*{(b) \color{black} Stress sensing Dataset}
        \label{fig:eda_samples}
    \end{minipage}
\vspace{-5pt}
    \caption{\color{black}{
Impact of limited local data availability (20\%, 50\%, 90\%) on fairness and utility in FL. The \textbf{left} subplots report the FATE score (higher is better), while the \textbf{right} subplots report the EO gap (lower is better). Results are shown for the (a) WIDAR dataset (top row) and (b) Stress Sensing dataset (bottom row). CurvFed consistently achieves higher FATE scores and lower EO gaps than baseline methods, demonstrating a superior fairness--utility trade-off under limited local data.}
}
\vspace{-15pt}
    \label{fig:fate_eo_gap}
\end{figure}

\begin{figure}[h!]
\vspace{-5pt}
    \centering
    \begin{minipage}[b]{0.7\linewidth}
        \centering
        \includegraphics[width=\textwidth]{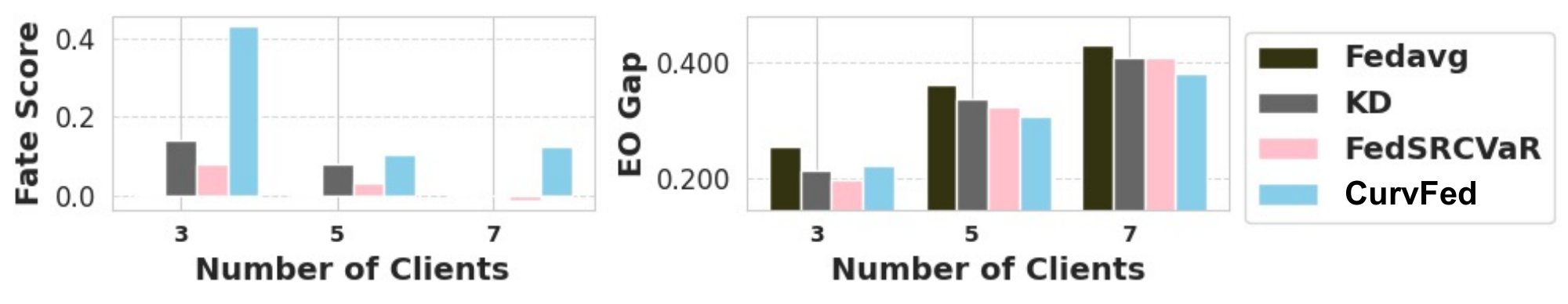}
        \caption*{(a) WIDAR Dataset}
        \label{fig:WIDAR_clients}
    \end{minipage}
    
    \vspace{0cm} 
    
    \begin{minipage}[b]{0.7\linewidth}
        \centering
        \includegraphics[width=\textwidth]{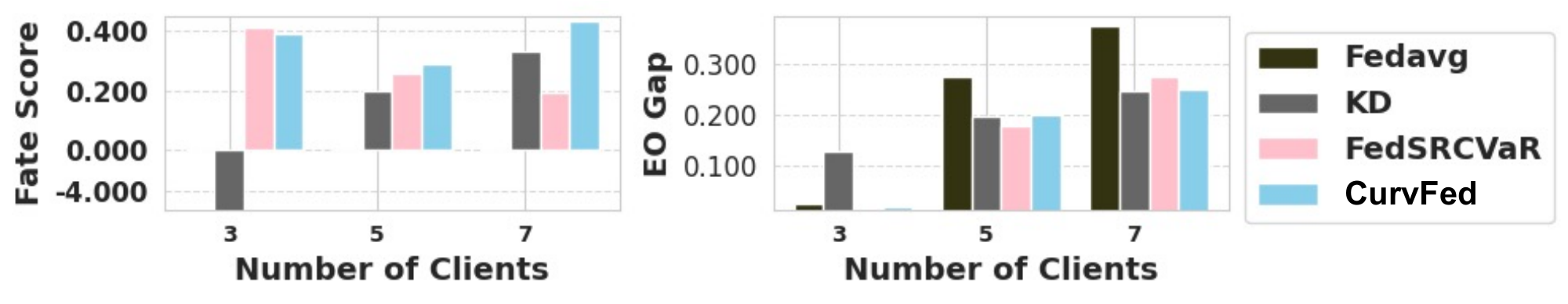}
        \caption*{(b) Stress sensing Dataset}
        \label{fig:eda_clients}
    \end{minipage}
      \vspace{-5pt}
    \caption{ \color{black}{Impact of client participation subsampling (3 or 5 or 7 clients per round) on fairness–utility trade-off. The \textbf{left} subplots report the FATE score (higher is better), while the \textbf{right} subplots report the EO gap (lower is better). Results are shown for the (a) WIDAR dataset (top row) and (b) Stress Sensing dataset (bottom row). CurvFed consistently achieves the highest FATE scores and lower EO gaps than baseline methods, demonstrating improved fairness–utility trade-offs under client subsampling.}}
    \label{fig:cliensample_eo_gap_fate}
\vspace{-10pt}
\end{figure}

As shown in Figs.~\ref{fig:fate_eo_gap} and~\ref{fig:cliensample_eo_gap_fate}, \textbf{\textit{CurvFed}} consistently demonstrates competitive or superior performance in terms of both fairness (lower EO gap) and utility (higher FATE score) across the WIDAR and Stress Sensing datasets. Notably, in low-data regimes (e.g., 20\%), \textbf{\textit{CurvFed}} maintains a clear advantage over FedAvg and KD, while FedSRCVaR occasionally achieves lower EO gaps but with significant sacrifices in FATE scores—especially evident in the WIDAR dataset with 3 and 5 clients. Furthermore, CurvFed's performance stabilizes with $\ge$ 50\% data, suggesting robustness even with moderate training data. In client subsampling scenarios, CurvFed maintains the best fairness-utility balance across different client counts, indicating its suitability for dynamic, resource-constrained FL deployments.

\subsubsection{\textbf{Impact of Random Client Participation on Fairness:}}
\label{random-client-eval}

In real-world FL deployments, especially in human-sensing applications, client availability is often inconsistent due to network instability and device constraints \cite{lim2020federated}. Unlike idealized settings with fixed clients per round, edge devices may randomly join or drop out during training. We simulate this, where clients were independently sampled using Bernoulli sampling with each client having a random probability (ranging from $0.1$ to $0.9$) of getting selected, introducing realistic variations in both the number and identity of participating clients.

\begin{figure}[!ht]
    \centering
    \includegraphics[width=0.7\linewidth]{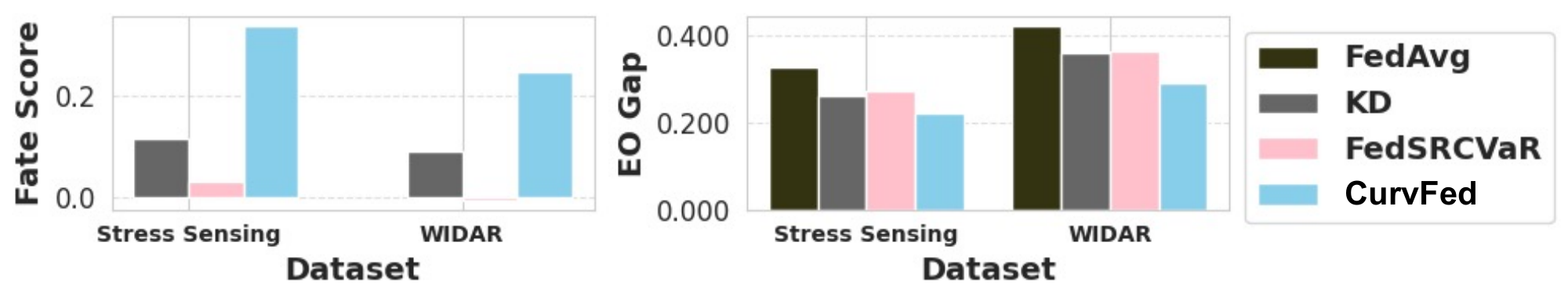}
    \vspace{-5pt}
    \caption{\color{black}{Experiment on real-world FL evaluation simulating an inconsistent number of clients in each round. The \textbf{left} subplots report the FATE score (higher is better), while the \textbf{right} subplots report the EO gap (lower is better). CurvFed consistently attains higher FATE scores and lower EO gaps across datasets, indicating improved fairness–utility trade-offs.}}
    \label{fig:fl_bernulli}
\vspace{-15pt}
\end{figure}

Despite this variability, \textbf{\textit{CurvFed}} consistently achieves the best fairness–accuracy tradeoff, achieving the highest FATE scores—while also attaining the lowest EO gaps, as shown in Fig. \ref{fig:fl_bernulli}. While FedSRCVaR sometimes matches fairness, it sacrifices accuracy, and KD-FedAvg shows unstable results. These results demonstrate \textbf{\textit{CurvFed}}'s robustness to client variability in practical FL deployments.

\subsubsection{\textcolor{black}{\textbf{Sensitivity to the Weighting Parameter $\alpha$}}}
\label{alpha}

\textcolor{black}{We study the sensitivity of \textbf{\textit{CurvFed}} to the weighting parameter $\alpha$ by examining its effect on classification performance (F1 score), fairness (EO Gap), and their combined trade-off measured by the FATE score. The parameter $\alpha$ controls the relative weighting of the components in the client-side optimization objective (Equation \ref{local_training_eq}), and its impact is therefore evaluated empirically.}

\begin{figure}[h!]
    \centering
    \begin{minipage}[t]{0.99\linewidth}
        \centering
        \includegraphics[width=\textwidth]{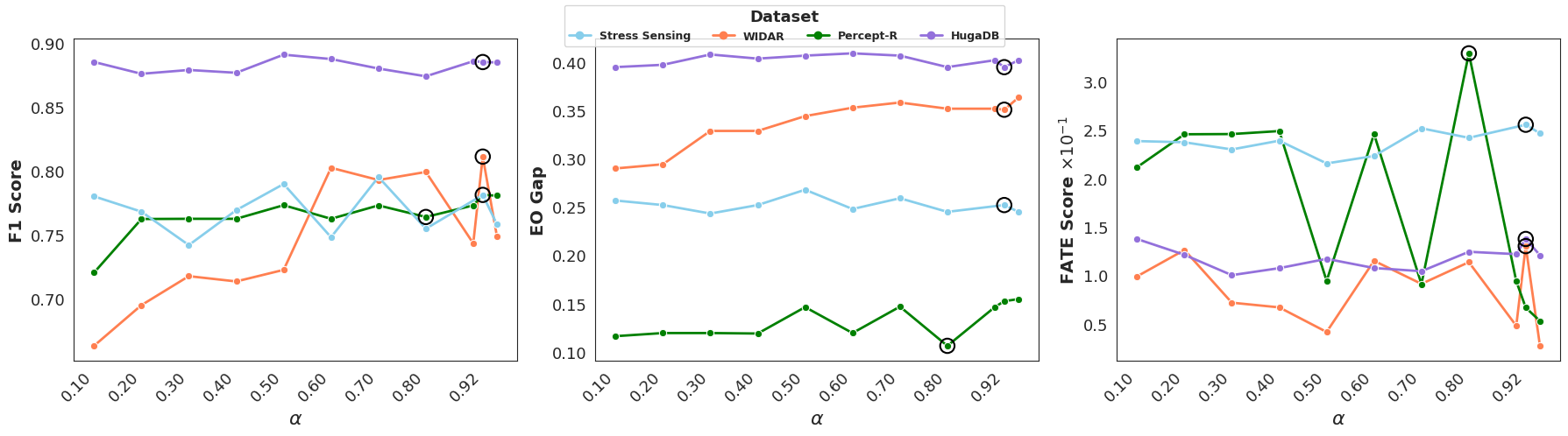}

    \end{minipage}
    \caption{\textcolor{black}{
    Impact of the weighting parameter $\alpha$ on (left) F1 score, (middle) EO gap, and (right) FATE score across datasets. Highlighted points denote the selected $\alpha$ values obtained via grid search, which balance higher F1 score, lower EO gap, and higher FATE score.
    }}
\label{fig:alpha}
\end{figure}

\textcolor{black}{We performed a grid search over $\alpha \in \{0.1, 0.2, \ldots, 0.8, 0.92, 0.95\}$ and report the corresponding F1 score, EO Gap, and FATE score. As shown in Figure~\ref{fig:alpha}, varying $\alpha$ leads to different performance--fairness profiles across datasets. We selected dataset-specific values of $\alpha$ corresponding to regions where the FATE score is comparatively high and both F1 score and EO Gap remain stable. In particular, $\alpha=0.8$ is selected for Percept-R, whereas $\alpha=0.92$ is used for Stress Sensing, WIDAR, and HugaDB datasets. These choices reflect operating points that provide a balanced trade-off between utility and fairness as observed in the empirical results.}

\textcolor{black}{\subsection{Ablation Study}\label{ablation-study}}

{\color{black}
To rigorously evaluate the architectural decisions underpinning \textbf{\textit{CurvFed}}, we conducted a component-wise ablation study across all datasets using 5-fold cross-validation. 
The goal was to disentangle both the individual and joint contributions of local curvature regularization and global sharpness-aware aggregation to achieving Fairness Without Demographics (FWD), as well as to clarify the roles of SAM in local client training and SWA in server-side aggregation.
Table~\ref{tab:model_performance_aba_percept} reports the cross-validated performance across twelve distinct configurations for the PERCEPT-R dataset, enabling an analysis of the trade-offs between model utility (F1 score) and equity (EO Gap). Results for the remaining datasets are provided in Appendix~\ref{model_performance_abalation}.

\begin{table}[h]
\color{black}
\centering
\resizebox{0.99\linewidth}{!}{
\begin{tabular}{|c|c|c|c|c|c|c|}
\hline
\textbf{\#} & \textbf{Dataset} &
\makecell{\textbf{Client-Side}\\ \textbf{Training}} &
\makecell{\textbf{Aggregation}\\ \textbf{Strategy}} &
\makecell{\textbf{F1 Score}\\ \textbf{Mean $\pm$ Std}\\ ($\uparrow$)} &
\makecell{\textbf{EO Gap}\\ \textbf{Mean $\pm$ Std}\\ ($\downarrow$)} &
\makecell{\textbf{FATE $\times 10^{-1}$}\\ \textbf{Mean Score}\\ ($\uparrow$)} \\
\hline

1 & \multirow{12}{*}{PERCEPT-R} & Benign & Fedavg & 0.7774 $\pm$ 0.0029 & 0.1626 $\pm$ 0.0060 & 0.000 \\
2 & & SAM & Fedavg & 0.7784 $\pm$ 0.0030 & 0.1509 $\pm$ 0.0125 & 0.732 \\
3 & & SAM & Fedswa & 0.7726 $\pm$ 0.0026 & 0.1305 $\pm$ 0.0139 & 1.914 \\
4 & & SAM+(Eq11) & Fedswa & 0.7727 $\pm$ 0.0025 & 0.1312 $\pm$ 0.0143 & 1.871 \\
5 & &  (Eq11) & Fedavg & 0.7474 $\pm$ 0.0023 & 0.1291 $\pm$ 0.0050 & 1.734 \\
6 & &  (Eq11) & (Eq12) & \textbf{0.7866} $\pm$ 0.0022 & 0.1589 $\pm$ 0.0131 & 0.287 \\
7 & & SAM+(Eq11) & (Eq12) & 0.7793 $\pm$ 0.0026 & 0.1483 $\pm$ 0.0119 & 0.905 \\
8 & & (Eq11) & Sharpness Aware aggregation & 0.7736 $\pm$ 0.0025 & 0.1514 $\pm$ 0.0125 & 0.642 \\
9 & & SAM & Fedswa+S(L) (Eq12* partial) & 0.7664 $\pm$ 0.0100 & 0.1324 $\pm$ 0.0280 & 1.717 \\
10 & & SAM & Fedswa+S(T) (Eq12* partial) & 0.7699 $\pm$ 0.0037 & 0.1288 $\pm$ 0.0207 & 1.987 \\
11 & & SAM & Sharpness Aware aggregation & 0.7662 $\pm$ 0.0101 & 0.1318 $\pm$ 0.0293 & 1.753 \\
12 & & SAM+(Eq11) & Sharpness Aware aggregation & 0.7642 $\pm$ 0.0030 & \textbf{0.1062} $\pm$ 0.0119 & \textbf{3.296} \\
\hline
\end{tabular}
}
\caption{\textcolor{black}{Ablation of \textbf{\textit{CurvFed}}'s design choices on the PERCEPT-R dataset. Results are averaged over 5-fold cross-validation. \textbf{Client-Side Training} indicates the optimization or regularization applied locally at each client. \textbf{Aggregation Strategy} specifies the server-side model aggregation rule .
Unless explicitly stated, each row applies \emph{only} the listed client-side or aggregation component, with no additional modifications. Metrics with $\uparrow$ indicate higher is better, while $\downarrow$ indicates lower is better. Best values for F1, EO Gap, and FATE are bolded. Rows are numbered for reference in the text.}}
\label{tab:model_performance_aba_percept}
\vspace{-2em}
\end{table}

This study systematically isolates and evaluates the independent contributions of three key components: (1) the \textit{Local curvature constraints} during client-side training, (2) \textbf{\textit{CurvFed}}'s \textit{sharpness-aware aggregation strategy} (the sharpness-aware weighting scheme $W^r$ in Equation~\ref{agg-combine-weight_eqn} combined with SWA), 
and (3) synergistic effect arising from \textit{integrating all components} within the full \textbf{\textit{CurvFed}} framework.

\subsubsection{\textbf{The Efficacy of Local Curvature Constraints}}
Our approach is based on the idea that local model sharpness correlates with biased decision boundaries. We validated this by applying the FIM penalty ($\lambda(F_{client})$) during local training, i.e., following Equation \ref{local_training_eq}, while maintaining a standard FedAvg aggregation (Row~5). Results from this isolation demonstrates that \textit{restricting the local search space to flatter regions inherently mitigates bias}: the EO Gap decreased by \textbf{20.60\%} compared to the benign baseline (Row~1). While this constraint induced a modest drop in F1 score ($\approx 3\%$), it provides empirical evidence that individual clients can contribute to global fairness through geometric regularization alone, even without server-side coordination.

To further disentangle the contribution of client-side curvature constraints from server-side sharpness-aware aggregation within the \textbf{\textit{CurvFed}} design, we evaluated an additional variant in which the sharpness-aware weighting scheme $W^r$ (Equation~\ref{agg-combine-weight_eqn}) was removed, while all other \textbf{\textit{CurvFed}} components were preserved (Row~4). This configuration achieved a lower EO Gap (\textbf{19.31}\% reduction) and comparable F1 score relative to the benign baseline (Row~1), but underperformed compared to the full \textbf{\textit{CurvFed}} framework (Row~12). 


Overall, these findings confirm that local curvature constraints independently promote fairness gains; however, their full potential is realized only when combined with server-side sharpness-aware aggregation, highlighting the complementary and synergistic design of the complete \textbf{\textit{CurvFed}} framework.


\subsubsection{\textbf{Sharpness-Awareness at the Aggregator}}
We next examined the role of the server in mitigating biased client updates. To isolate the effect of our Sharpness-Aware Aggregation (the sharpness-aware weighting scheme $W^r$ in Equation~\ref{agg-combine-weight_eqn} combined with SWA), we removed the local curvature constraint (Equation~\ref{local_training_eq}) from the \textbf{\textit{CurvFed}} framework (Row~11) and compared it against clients trained with SAM while the server used standard FedAvg (Row~2)—keeping client-side training consistent across settings.
Isolated Sharpness-Aware Aggregation of \textbf{\textit{CurvFed}} reduced the EO Gap by \textbf{12.65\%} relative to the baseline (Row~2). This suggests that simply averaging locally flat models (SAM+FedAvg; where SAM promotes flatter minima) is insufficient; the aggregation mechanism itself must account for the sharpness of the incoming updates to prevent the re-introduction of sharp, biased minima during the global update step.


\subsubsection{\textbf{Navigating the Fairness-Utility Tension}}
The ablation results illustrate the distinct roles of aggregation and local optimization in balancing the accuracy-fairness trade-off. Notably, Row~6—which combines applying the FIM penalty ($\lambda(F_{client})$) during local training, i.e., following Equation \ref{local_training_eq}, with the sharpness-aware weighting scheme $W^r$ in Equation~\ref{agg-combine-weight_eqn}
—achieved the \textit{highest absolute F1 score} (0.7866) of all configurations.
This result confirms that our curvature-aware aggregation strategy is highly effective at identifying and promoting stable, high-performing models. However, this configuration yielded a modest FATE score (0.287) with a small EO Gap reduction than the benign baseline (Row~1). 
In contrast, the full \textbf{\textit{CurvFed}} integration (Row~12) accepts a minor utility trade-off (1.32\% lower F1 than benign baseline) to secure a dramatic \textbf{34.66\% reduction} in EO Gap compared to Benign Fedavg (Row~1). Consequently, \textbf{\textit{CurvFed}} achieves the highest FATE score (3.296) among all methods.
These findings suggest that applying curvature constraints alone—via FIM regularization at both the client and server levels—can make optimization more challenging. In contrast, incorporating SAM and SWA at the client and server, respectively, facilitates smoother optimization dynamics, enabling the model to reach a more favorable fairness–utility trade-off and effectively address the FWD objective.






}

\section{Empirical Justification}
\label{empirical-verification}

While Section \ref{sec:results} confirms the effectiveness of \textbf{\textit{CurvFed}}, this section offers an empirical justification of \textit{why and how} it works. Specifically, we empirically evaluate whether the observed outcomes support the propositions introduced in Section~\ref{roadmap}, which hypothesize the mechanisms through which CurvFed improves fairness and performance in the absence of sensitive attributes.
Each of the following subsections corresponds to one of the three propositions and presents analysis and visualizations using the WIDAR and stress sensing datasets.

\begin{figure}[h!]
    \centering
        \begin{minipage}[t]{0.7\linewidth}
        \centering
        \includegraphics[width=\textwidth]{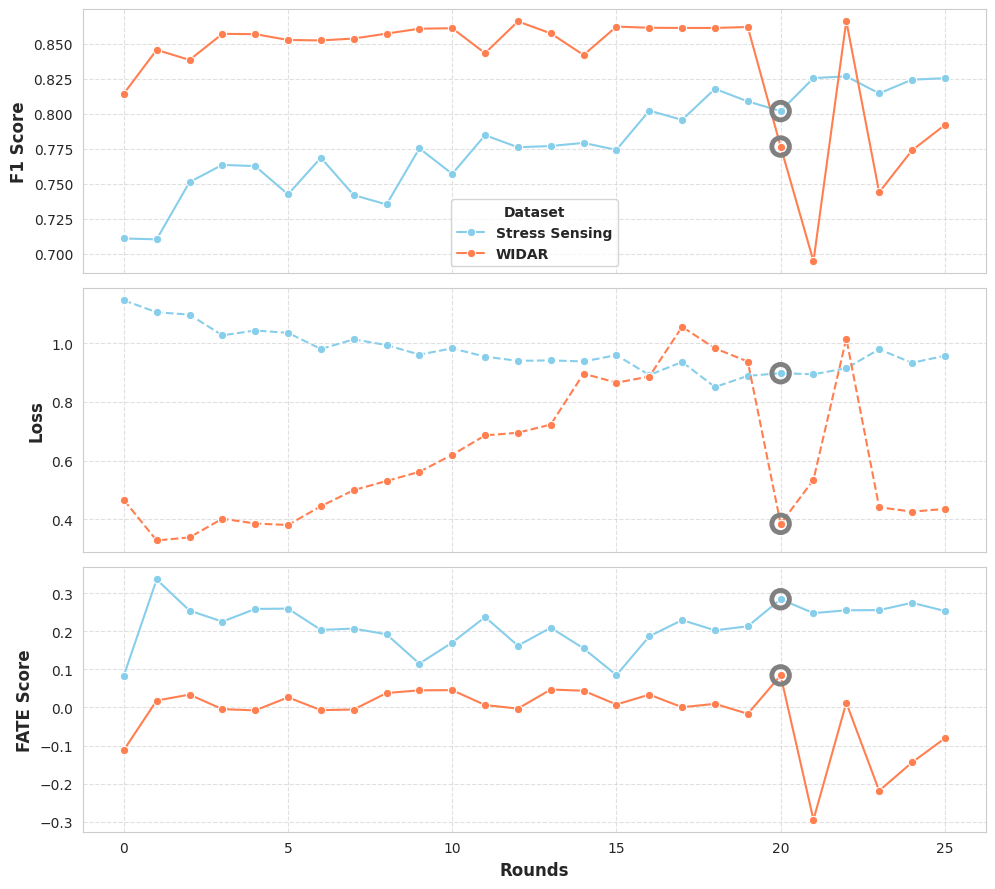}
        \label{fig:communication_f1_loss}
    \end{minipage}
    \vspace{-25pt}
\caption{\textcolor{black}{
Evolution of F1 score (top), loss (middle), and FATE score (bottom) across communication rounds for CurvFed on the Stress Sensing and WIDAR datasets. Both datasets exhibit similar convergence trends, with F1 and FATE scores stabilizing as loss decreases. The highlighted round corresponds to the point where both loss and F1 converge, and it also aligns with a stable or improved FATE score, indicating a balanced trade-off
}}
    \label{fig:communication}
\vspace{-5pt}
\end{figure}

\textbf{(Validation of P1) Alignment between FIM and Hessian top eigenvalue:} Fig. \ref{fig:communication} illustrates the relationship between communication rounds and test-set model performance in terms of the FATE Score, F1 Score, and Loss Landscape, for the Stress Sensing dataset and WIDAR dataset. We utilized FATE score as our selection criterion for the best global model highlighted in yellow, corresponding to round 20.

\begin{figure}[h!]
    \centering
    \begin{minipage}[b]{0.7\linewidth}
        \centering
        \includegraphics[width=\textwidth]{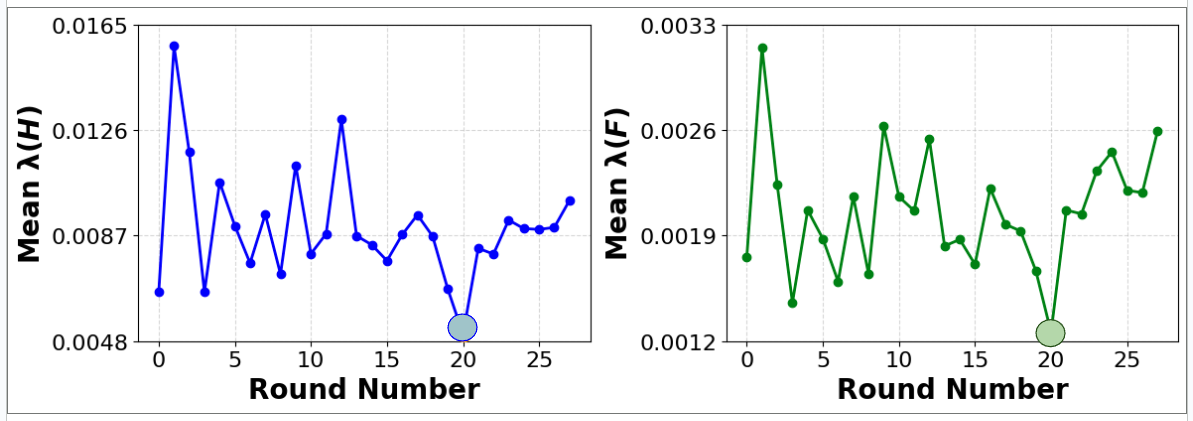}
        \caption*{(a) Stress-sensing Dataset}
        \label{fig:eda_hessian_fisher}
    \end{minipage}
    
    \begin{minipage}[b]{0.7\linewidth}
        \centering
        \includegraphics[width=\textwidth]{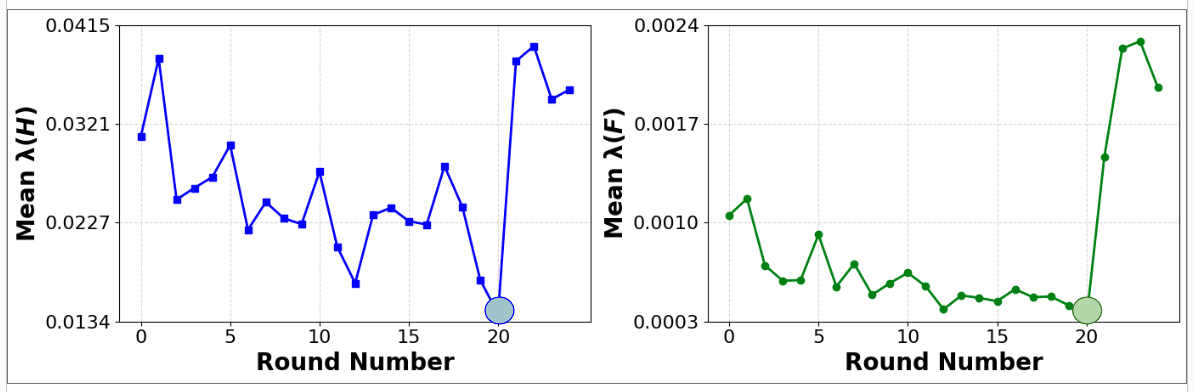}
        \caption*{(b) WIDAR Dataset}
        \label{fig:widar_hessian_fisher}
    \end{minipage}
    \vspace{-10pt}
\caption{\color{black}{
Mean person-wise top eigenvalues of the Hessian (left) and Fisher (right) matrices across federated learning rounds for the Stress Sensing (top row) and WIDAR (bottom row) datasets. Both datasets exhibit similar decreasing and stabilizing trends, indicating consistent optimization behavior across tasks. Lower eigenvalues correspond to flatter and more stable solutions. The highlighted rounds, selected based on F1 and loss convergence, aligning with minimum eigenvalues.
}}

\label{fig:hessian_fisher_comparison}
\vspace{-5pt}
\end{figure}

Fig. \ref{fig:hessian_fisher_comparison} shows the mean per-person top eigenvalues of the Hessian and FIM for the Stress Sensing in Fig. \ref{fig:hessian_fisher_comparison} (a) and WIDAR dataset in Fig. \ref{fig:hessian_fisher_comparison} (b). Notably, the graph traits for both FIM and Hessian are similar, and lowest in our optimal selected round ($20$), providing empirical justification for \emph{proposition 1}.

\textbf{(Validation of P2) Minimizing the FIM Top Eigenvalue Reduces group Curvature Disparity:} To evaluate whether FL training aligns with proposition 2 discussed in Section \ref{roadmap}, we analyze the difference in curvature metrics across known sensitive attribute groups. Table \ref{tab:summary_metrics} reports the difference in the top eigenvalues ($\lambda$) of the FIM and Hessian matrix for the sensitive attribute \textit{sex} across two distributions, $D_{\text{male}}^{\text{sex}}$ and $D_{\text{female}}^{\text{sex}}$ along with the FATE scores 
for the WIDAR and Stress sensing datasets.

\emph{Global Trends:} FedSAM and FedSWA reduce curvature disparity via implicit sharpness minimization at the client and server levels, respectively. While this results in moderate gains in both curvature alignment and FATE scores, \textbf{CurvFed} consistently outperforms them in curvature alignment and FATE scores, confirming the superior effectiveness of explicit FIM-based sharpness regularization.


\begin{table}[h!]
    \centering
    \resizebox{0.6\linewidth}{!}{
    \begin{tabular}{|c|c|c|c|c|}
        \hline
        \textbf{Dataset} & \textbf{Model} & $\Delta \mathrm{\lambda(\textbf{F}})$ $\downarrow$ & $\Delta \mathrm{\lambda(\textbf{H}} )$ $\downarrow$  & \textbf{FATE Score $\times 10^{-1}$ $\uparrow$ } \\
        \hline
        \multirow{3}{*}{\color{black} WIDAR} & Fedavg  & 0.009 & 0.061 & 0.0000 \\ \cline{2-5}
                                & FedSAM & 0.009 & 0.0365  & -0.56 \\
                                & FedSWA & 0.002 & 0.019  & 0.33 \\
                                & FedSRCVaR  & 0.0015 & 0.0310 & 0.160 \\ 
                                & KD-Fedavg & 0.0022& 0.0377 & 0.122 \\
                                & CurvFed & \textbf{0.0007} & \textbf{0.0017}& \textbf{1.233}\\  
        \hline
        \multirow{3}{*}{Stress Sensing}   & Fedavg  & 0.03 & 0.0330  & 0.0000 \\ \cline{2-5}
                                & FedSAM  & 0.019 & 0.0347  & 0.812 \\
                                & FedSWA  &0.020 & 0.0278  & 0.38 \\
                                & FedSRCVaR  & 0.019& 0.1446& 0.319\\ 
                                & KD-Fedavg & 0.021 &0.6089 & 1.645\\                 
                                & CurvFed&  \textbf{0.004} & \textbf{0.0097} & \textbf{2.559}\\
        \hline
    \end{tabular}
    }
\caption{\textcolor{black}{Disparity based on `Sex' Across Datasets for Different Models. Metrics with $\uparrow$ indicate higher is better, while metrics with $\downarrow$ indicate lower is better.CurvFed achieves the lowest disparity measures and the highest FATE score across both datasets.}}
\label{tab:summary_metrics}
\vspace{-15pt}
\end{table}

\begin{figure}[!ht]
    \centering
    \includegraphics[width=0.7\linewidth]{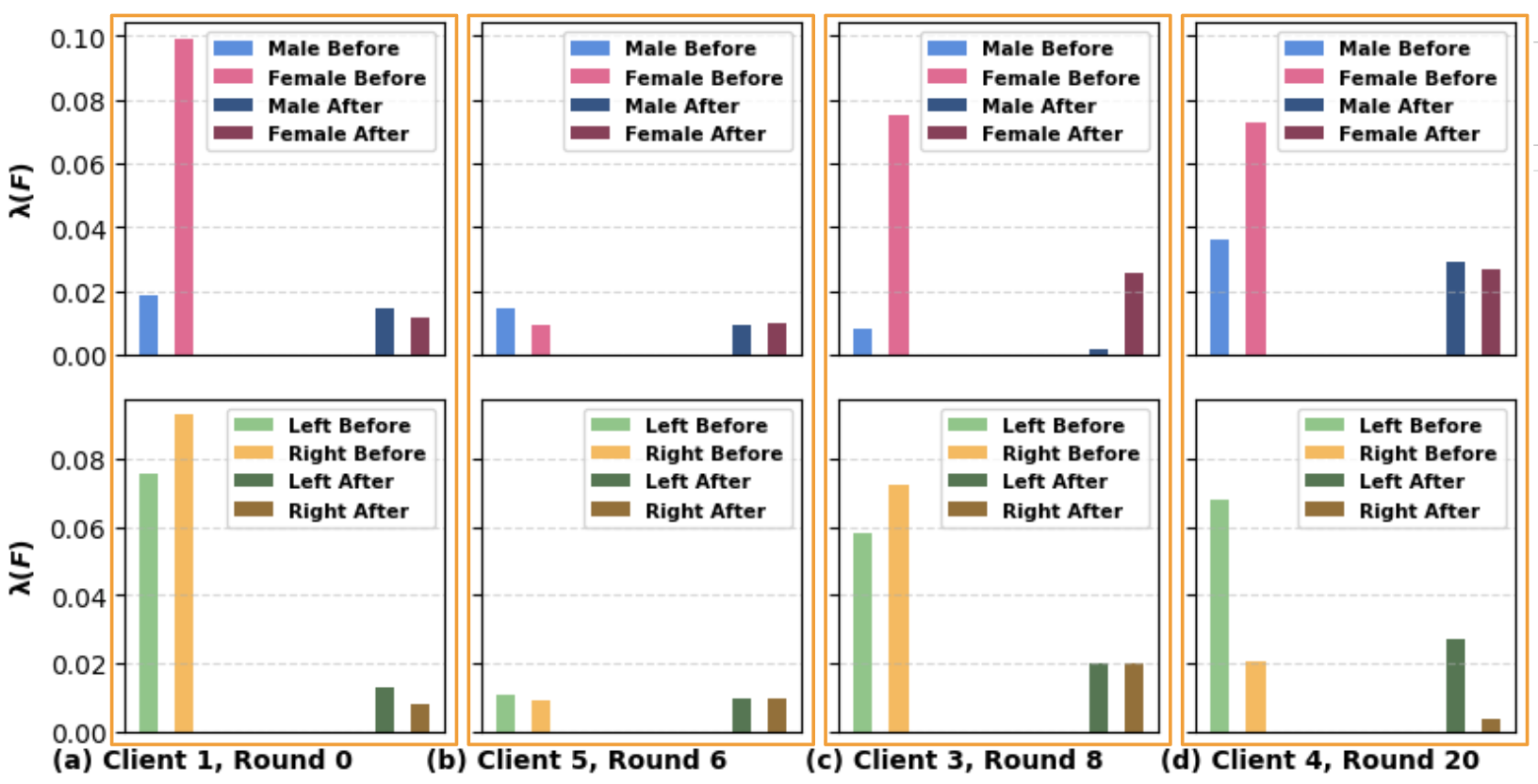}
    \vspace{-5pt}
    \caption{\color{black}{{\color{black}{Top Fisher eigenvalues by sex (top) and hand (bottom) across clients and rounds in the Stress Sensing dataset. Bars labeled ``Before'' and ``After'' denote values prior to and following the application of the fairness-aware training. Reduced gaps between groups and lower $\lambda(F)$ and $\lambda(H)$ values after training indicate decreased group-level disparity and improved optimization stability.}}
    }}
    \label{fig:fisher_all_minipage}
\vspace{-10pt}
\end{figure}

\textit{Client-Level Trends: Does CurvFed Reduce group-wise Disparity Locally?} To examine whether global fairness trends extend to individual clients, we visualize group-wise FIM eigenvalues for randomly selected clients from the Stress Sensing dataset (see Fig. \ref{fig:fisher_all_minipage}). We analyze changes across sensitive attributes such as \textit{sex} and \textit{handedness} before and after local training.

The visualizations show that reducing the top eigenvalue of the FIM not only lowers overall curvature but also reduces disparity between groups at the client level—supporting our \emph{proposition 2}.

\begin{figure}[h!]
    \centering
    \setcounter{subfigure}{0}

    \subfigure[Stress Sensing Dataset]{%
        \begin{minipage}[b]{0.78\linewidth}
            \centering
            \includegraphics[width=\textwidth]{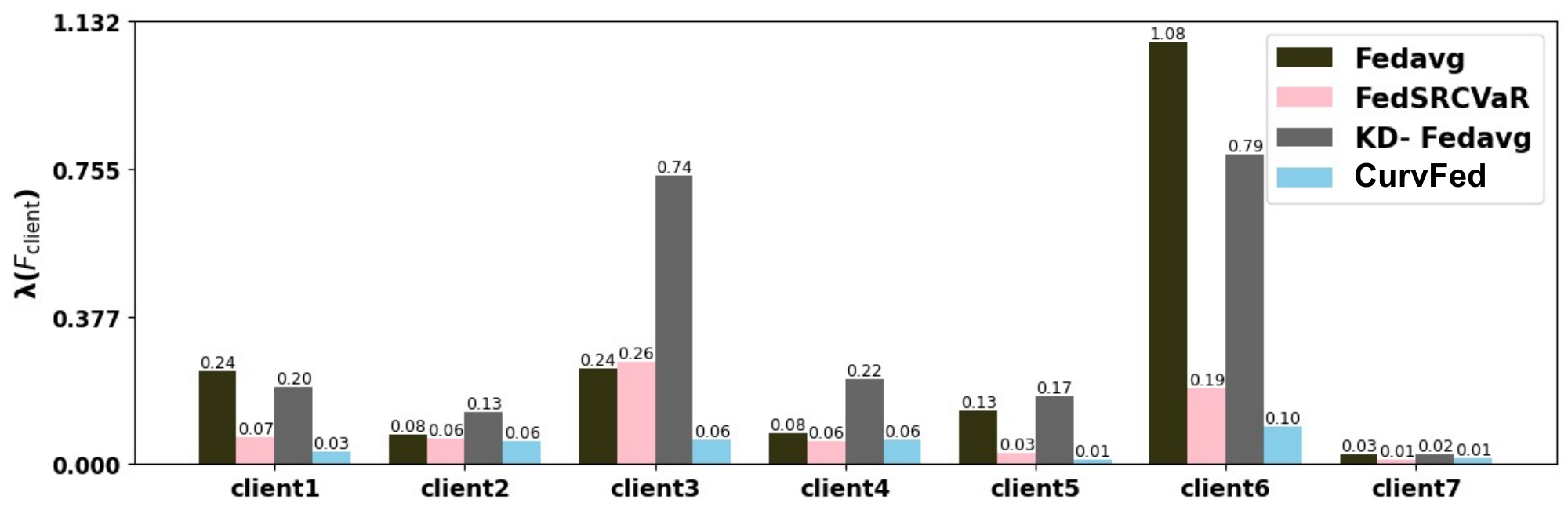} \\
            \includegraphics[width=\textwidth]{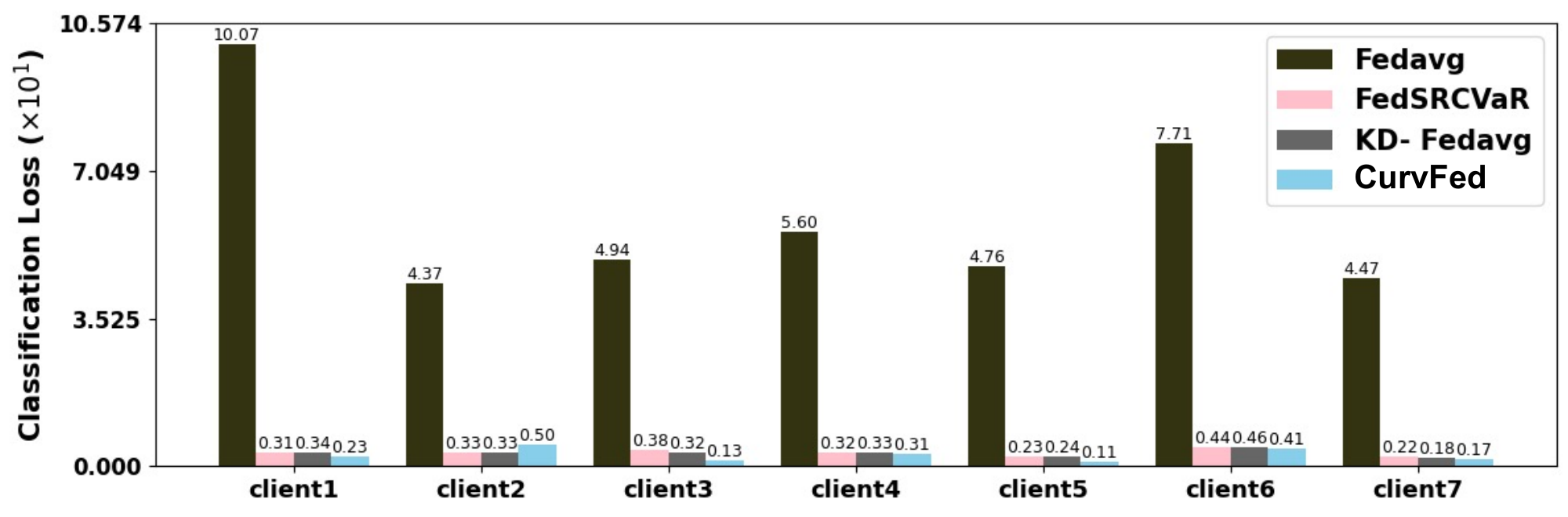}
        \end{minipage}
    } \\

    \subfigure[WIDAR Dataset]{%
        \begin{minipage}[b]{0.7\linewidth}
            \centering
            \includegraphics[width=\textwidth]{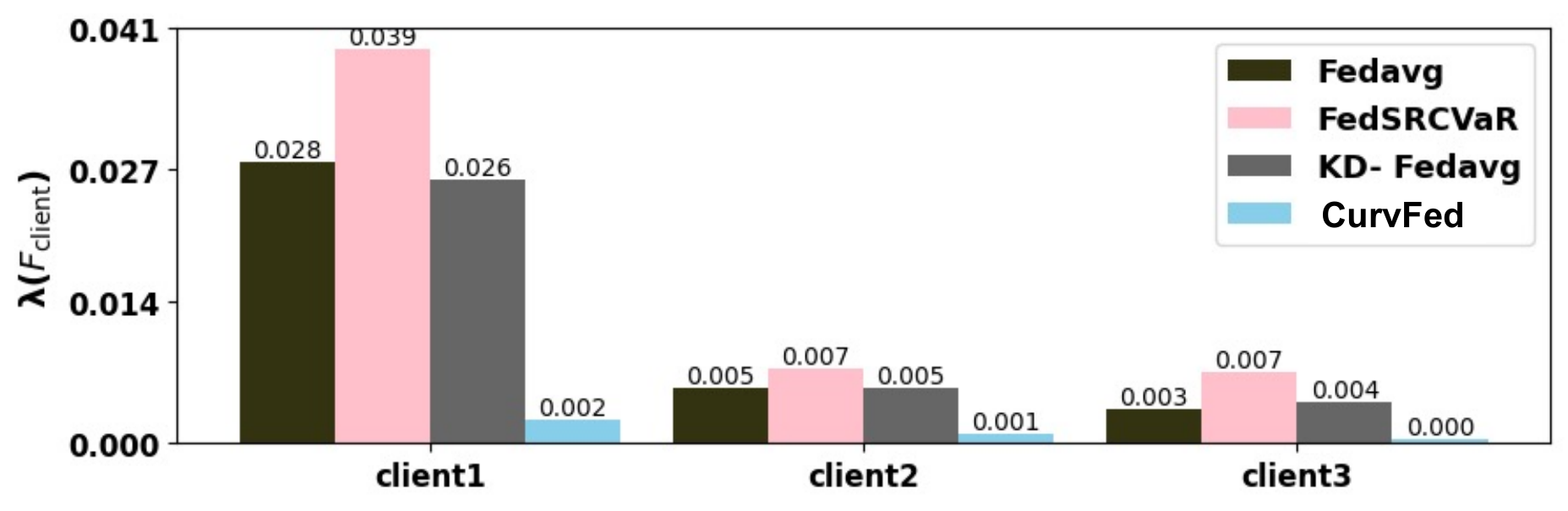} \\
            \includegraphics[width=\textwidth]{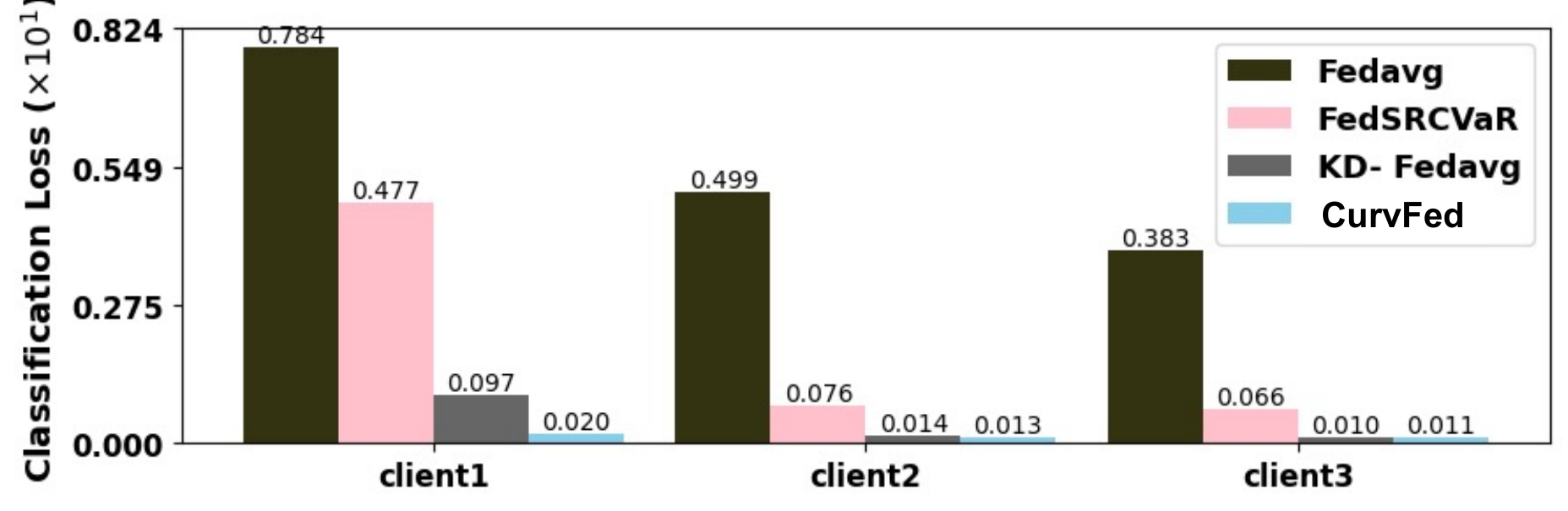}
        \end{minipage}
    }
 \vspace{-15pt}
\caption{\color{black}{
Client-wise comparison of the maximum Fisher eigenvalue $\lambda(F_{\text{client}})$ (top) and classification loss (bottom) across federated learning models on the Stress Sensing (top row) and WIDAR (bottom row) datasets. Lower values in both metrics indicate more stable client updates and reduced client-level disparity. CurvFed consistently achieves lower Fisher eigenvalues and lower classification loss across clients, demonstrating improved optimization stability and reduced fairness-related client heterogeneity}}

    \label{fig:losses_benign_proposed1}
    \vspace{-10pt}
\end{figure}

\textbf{(Validation of P3) Sharpness-Aware Aggregation Reduces Variance in Excessive Loss Across Clients:} To verify whether \textbf{\textit{CurvFed}} indeed upholds this proposition, we visualized the $\lambda(F_{client}))$ and classification loss among single-person and multi-person clients in FL systems for WIDAR and Stress-Sensing datasets, as shown in Fig. \ref{fig:losses_benign_proposed1}. \textbf{\textit{CurvFed}} outperforms other FWD baselines by consistently reducing variance across the excessive loss in the form of `reducing variances across the top eigenvalue values $\lambda(F)$ and loss' across clients. \textit{This visualization confirms proposition 3.}


\section{Broader Impact and Limitation}
\label{broader-impact-limit}
This first-of-its-kind paper bridges the gap between fairness and federated learning in human-sensing applications, paving the way for a more equitable and responsible future in decentralized human-sensing, ensuring fair access to technological advancements for all.
\emph{Some limitations and future research scopes are discussed below:}
\begin{itemize}

\item Achieving fairness in FL systems requires attaining a balance between fairness and model performance \cite{gu2022privacy}. The quest to ensure fairness for individual clients often results in reduced performance across the board, an issue highlighted in literature \cite{ezzeldin2023fairfed,zhang2020fairfl,al2014crowd,wang2019age}. Although \textbf{\textit{CurvFed}} demonstrates significant effectiveness in achieving this balance, as discussed in Section \ref{eval-results}, \textbf{RQ-3}, it exhibits minor performance degradation in the WIDAR and Stress sensing dataset for single-person clients. This could be attributed to our aggregation scheme, outlined in Section \ref{approach-design-choice} Equation \ref{agg-combine-weight_eqn}, which equally weighs low error rates and lower maximal uncertainty, thus limiting the ability to balance accuracy and fairness when dealing with single-person clients. The issue could be resolved by introducing a hyperparameter as the weighting factor.

\item Recent studies have explored the relationship between fairness and privacy in machine learning, often treating them as competing metrics \cite{tran2023interplay,zhang2024unraveling}. While this work has focused on fairness in FL, extending the investigation to explore the interplay between fairness and privacy in FL would be an important direction for future research.

\item A limitation of our study is the experimental scale, constrained by the lack of large-scale human-sensing datasets with sensitive attributes. While prior (non-human-centric) FL fairness works often evaluate on a larger number of clients, such datasets remain scarce in the human sensing domain \cite{stateofalgobias}. Even recent large-scale datasets like GLOBEM \cite{xu2023globem} do not release demographic or other sensitive information needed for fairness evaluation. This gap has also been discussed in recent survey studies within the ubiquitous sensing community \cite{yfantidou2023beyond, uncoveringbias}.
\item Like many Rawlsian or min–max fairness approaches \cite{lahoti2020fairness,hashimoto2018fairness}, \textbf{\textit{CurvFed}} promotes equity by aligning loss landscape curvatures across clients, which prioritizes worst-case (or underrepresented) groups. This strategy can occasionally come at the expense of performance in majority groups, a trade-off that has also been reported in other fairness-driven FL methods \cite{mohri2019agnostic,papadaki2022minimax,du2021fairness}. While this reflects a broader property of parity-based optimization \cite{foulds2020parity,padh2021addressing} rather than a limitation of \textbf{\textit{CurvFed}} itself, it highlights an important avenue for future research: designing FWD mechanisms that uplift the worst-off groups without diminishing performance for others, thereby moving toward Pareto-improving fairness \cite{zanna2024enhancing,nagpal2025optimizing}.

\end{itemize}
\section{Conclusion}
This paper presents \textbf{\textit{CurvFed}}, a novel fairness-enhancing strategy for FL that introduces sharpness-aware regularization during local training and aggregation. Unlike existing methods, \textbf{\textit{CurvFed}} is theoretically grounded and operates without access to sensitive or bias-inducing attributes, thereby preserving FL’s core privacy principles. Extensive evaluations—including a live real-world deployment on heterogeneous, resource-constrained edge devices—demonstrate \textbf{\textit{CurvFed}}’s effectiveness in achieving a strong fairness–performance trade-off across diverse human-sensing applications, even under single- or multi-user scenarios with multiple unknown bias sources. Ablation studies and empirical justifications further validate that leveraging loss-landscape sharpness is key to improving fairness. Overall, \textbf{\textit{CurvFed}} offers a practical and privacy-respecting path forward for equitable FL in real-world human-sensing systems.

\bibliographystyle{ACM-Reference-Format}

\bibliography{sample-base}
\newpage
\appendix
\label{appendix}
\section{Methodological Transparency \& Reproducibility Appendix (META)}

\subsection{Dataset and Model Description}

\label{Appendix:data_label}
This section provides a detailed overview of the dataset and the models we utilized in our evaluations.
\subsubsection{Dataset Description}
\label{appen:data}
\textbf{WIDAR Dataset:} The WIDAR3.0 dataset \cite{zhang2021WIDAR3} contains WIFI channel state information (CSI) \cite{kaurwireless} which represents the variation in the Wifi channel portrayed by commodity wifi devices \cite{kaurwireless,zhang2021WIDAR3}. The WIFI CSI contains fine-grained information about variations in the Wifi channel based on the Doppler phenomenon \cite{zhang2021WIDAR3}. These signals have been used by prior researchers for gesture detection \cite{zhang2021WIDAR3}, presence detection \cite{sathya2023systematic}, pose estimation \cite{wang2018csi}, human identification and activity detection \cite{shahbazian2023human}. We have used it as one of the HAR (Human Activity Recognition) datasets. The dataset contains data from 16 participants, 4 Females, and 12 Males, with five different orientations (labeled as 1 to 5).
We removed a female participant from the dataset due to the absence of ground truth. The selected two gestures, `Slide' and `Sweep,' had data in all orientations for male and female participants in the presence of the wifi sensor. Since all participants had an equal number of samples in different orientations, i.e., orientations 1 to 5, we created a disparity in the dataset based on orientations by randomly sub-sampling 50\% of the data belonging to orientations 4 and 5. This was done to create a training imbalance representative of real-life scenarios where some groups have more representation compared to others. We considered labels 1 to 3 as major orientation and the rest as minor orientation. For our work, we have trained a model to perform binary classification to detect Sweep vs. Slide gestures using WIFI-CSI data and fairness on the basis of `sex' and `orientation' as sensitive attributes. 

\textbf{HuGaDB Dataset:}
We have also utilized the HuGaDB (Human Gait Database) dataset \cite{chereshnev2018hugadb} for HAR purposes, which comprises a diverse array of human activity recordings from 18 participants. The participant group includes 4 females and 14 males. In our work, we trained a binary classification model to distinguish between Standing and other activities from this dataset, with fairness considerations based on `sex' as the sensitive attribute.

\textbf{Stress sensing Dataset:}
Xiao et al. \cite{xiao2024reading} collected this dataset. This dataset
was obtained from 48 individuals, including 40 females and 8 males, who participated in four different stress-inducing tasks (Arithmetic, Bad Memory, Stress Videos, Pre-condition ), each associated with unique stressors. The dataset is not publicly available yet; Access to the dataset has been obtained through proper IRB approvals/extensions. Due to the continuation of data collection by the Xiao et al. \cite{xiao2024reading}'s research group, this paper's reported dataset has more individuals than reported by the Xiao et al. \cite{xiao2024reading} paper. The dataset has individuals wearing Empatica E4 \cite{mccarthy2016validation} in their left and/or right hands and individuals' sex information. Hence, this paper utilized `hand' information and `sex' as sensitive attributes for evaluation. 
Utilizing this dataset, we performed a binary stress detection classification task.

{\color{black}

\textbf{ PERCEPT-R dataset:}
The PERCEPT-R corpus \cite{benway2022percept} is an open-access clinical speech repository specialized for the study of American English rhotic production /\rotatebox[origin=c]{180}{r}/ in children and adolescents. The dataset comprises over 32 hours of citation speech (single words and short phrases) is a part of an ongoing longitudinal study collected from 281 participants aged 6 to 24 years ($\mu=11.3$ years), including typically developing speakers and those with Residual Speech Sound Disorders (RSSD). 
For our evaluation, we utilize a publicly available subset of 76 participants. Audio samples were originally recorded as 16-bit PCM WAV files at 44.1 kHz. Ground-truth labels were derived from perceptual judgments by trained listeners (speech-language pathologists) or crowdsourced raters, identifying productions as either ``rhotic'' (correct) or ``derhotic'' (incorrect). Feature extraction focused on spectral characteristics critical for rhoticity, specifically the distance between the second and third formants (F3-F2), which serves as a primary acoustic marker for /\rotatebox[origin=c]{180}{r}/ quality in American English. Consistent with the dataset's clinical demographics, \textit{sex} is retained as a sensitive attribute due to the higher prevalence of RSSD in males compared to females which was also reflected in the dataset distribution with 50 male and 26 female participant. 
}

\subsubsection{Model Description}
\label{Appendix:model_desc}

\textbf{WIDAR Model:} The WIDAR Model architecture (adapted from \cite{yang2023benchmark}) is comprised of two fully connected layers within a sequential module, featuring an input size of 22 * 20 * 20. A dropout layer with a 40\% dropout rate and a Rectified Linear Unit (ReLU) activation function follow the first linear layer, while the second linear layer outputs the classification result. The model reshapes the input tensor during the forward pass and incorporates dropout layers to mitigate over-fitting. 

\textbf{HugaDB Model:}
The HugaDB model architecture (adapted from \cite{pirttikangas2006feature}) is characterized by its complexity, featuring two sequential fully connected layers. The first layer consists of linear transformations with input and output sizes of 792 to 1024, followed by dropout with a 30\% probability, Rectified Linear Unit (ReLU) activation, and Instance Normalization. Subsequently, another linear layer reduces the dimensionality to 512, followed by similar dropout, ReLU, and Instance Normalization. The second layer further reduces the dimensionality from 512 to 128, followed by a linear layer with an output size equal to the specified number of classes. The final output is obtained by applying the sigmoid function to the result. During the forward pass, the input tensor is processed through these layers, reshaped when necessary, and the output is returned after applying the sigmoid function. 

\textbf{Stress sensing Model:}
The Stress sensing model architecture (adapted from \cite{sharma2022psychophysiological}) comprises three sequential fully connected layers. The first layer takes input of size 34 and produces an output of size 64, incorporating batch normalization and Rectified Linear Unit (ReLU) activation. The second layer further processes the output, reducing it to size 128 with ReLU activation. The final layer transforms the output to the specified embedding size.

{\color{black}
\textbf{Percept-R model:} The Percept-R model utilizes a compact 1D Convolutional Neural Network (CNN) designed to classify multivariate time-series data. The input consists of 5 acoustic feature channels over a sequence length of 60 time steps. 

The feature extraction block is composed of two consecutive 1D convolutional layers. The first layer maps the 5 input channels to 3 output channels using a kernel size of 5 and a stride of 1. The second layer reduces the dimensionality further, mapping the 3 channels to a single channel, maintaining a kernel size of 5 and a stride of 1. This process results in a feature map of length 52, which is subsequently flattened.

The classification head is a MLP consisting of two hidden layers, each containing 32 units with Hardswish activation functions. The final layer is a linear projection mapping the 32 latent features to 2 output units for binary classification.} 

\newpage
\subsubsection{Client Distribution:}
\label{appendix:client}
\textcolor{black}{This section describes the client distribution splits used to address \textbf{RQ-1} (Table~\ref{tab:model_performance_main}) and \textbf{RQ-2} (Table~\ref{tab:combined_model_performance_client_single_multi}) in the main text. The distribution of multi-person clients is reported in Table~\ref{tab:clients_distribution} (a), while Table~\ref{tab:clients_distribution} (b) summarizes the distribution of both single- and multi-person clients.} 

\FloatBarrier
\begin{table}[H]
\centering
\small


\begin{minipage}[t]{0.48\textwidth}
\centering

\begin{minipage}[t]{0.48\textwidth}
\centering
\resizebox{\linewidth}{!}{
\begin{tabular}{|l|c|c|}
\hline
\multicolumn{3}{|c|}{\textbf{WIDAR}} \\ \hline
Client & Male & Female \\ \hline
Client 1 & 4 & 1 \\
Client 2 & 4 & 1 \\
Client 3 & 4 & 1 \\ \hline

\multicolumn{3}{|c|}{\textbf{HugaDB}} \\ \hline
Client 1 & 4 & 1 \\
Client 2 & 4 & 1 \\
Client 3 & 4 & 1 \\
Client 4 & 2 & 1 \\ \hline

\multicolumn{3}{|c|}{\textbf{Stress Sensing}} \\ \hline
Client 1 & 2 & 4 \\
Client 2 & 4 & 2 \\
Client 3 & 1 & 6 \\
Client 4 & 1 & 6 \\
Client 5 & 2 & 3 \\
Client 6 & 1 & 7 \\
Client 7 & 3 & 6 \\ \hline
\end{tabular}}
\end{minipage}
\hfill
\begin{minipage}[t]{0.48\textwidth}
\centering
\resizebox{\linewidth}{!}{
\begin{tabular}{|l|c|c|}
\hline
\multicolumn{3}{|c|}{\textcolor{black}{\textbf{PERCEPT-R}}} \\ \hline
Client & Male & Female \\ \hline
Client 0 & 4 & 0 \\
Client 1 & 4 & 1 \\
Client 2 & 2 & 3 \\
Client 3 & 4 & 1 \\
Client 4 & 3 & 2 \\
Client 5 & 3 & 2 \\
Client 6 & 1 & 3 \\
Client 7 & 1 & 1 \\
Client 8 & 0 & 3 \\
Client 9 & 3 & 0 \\
Client 10 & 3 & 0 \\
Client 11 & 4 & 1 \\
Client 12 & 1 & 1 \\
Client 13 & 4 & 1 \\
Client 14 & 1 & 3 \\
Client 15 & 1 & 1 \\
Client 16 & 1 & 1 \\
Client 17 & 3 & 0 \\
Client 18 & 2 & 0 \\
Client 19 & 1 & 1 \\
Client 20 & 2 & 0 \\
Client 21 & 2 & 1 \\ \hline
\end{tabular}}
\end{minipage}

\vspace{0.6em}
\textbf{(a) Multi-person Clients}

\end{minipage}
\hfill
\vrule width 0.5pt
\hfill
\begin{minipage}[t]{0.48\textwidth}
\centering

\begin{minipage}[t]{0.48\textwidth}
\centering
\resizebox{\linewidth}{!}{
\begin{tabular}{|l|c|c|}
\hline
\multicolumn{3}{|c|}{\textbf{WIDAR}} \\ \hline
Client & Male & Female \\ \hline
Client 1 & 1 & 1 \\
Client 2 & 2 & 1 \\
Client 3 & 1 & 0 \\
Client 4 & 2 & 0 \\
Client 5 & 2 & 1 \\
Client 6 & 1 & 0 \\
Client 7 & 1 & 0 \\
Client 8 & 2 & 0 \\ \hline

\multicolumn{3}{|c|}{\textbf{HugaDB}} \\ \hline
Client 1 & 6 & 0 \\
Client 2 & 5 & 1 \\
Client 3 & 0 & 2 \\
Client 4 & 0 & 1 \\ \hline

\multicolumn{3}{|c|}{\textbf{Stress Sensing}} \\ \hline
Client 1 & 1 & 0 \\
Client 2 & 3 & 0 \\
Client 3 & 6 & 10 \\
Client 4 & 0 & 6 \\
Client 5 & 3 & 7 \\
Client 6 & 2 & 9 \\
Client 7 & 0 & 1 \\ \hline
\end{tabular}}
\end{minipage}
\hfill
\begin{minipage}[t]{0.48\textwidth}
\centering
\resizebox{\linewidth}{!}{
\begin{tabular}{|l|c|c|}
\hline
\multicolumn{3}{|c|}{\textcolor{black}{\textbf{PERCEPT-R}}} \\ \hline
Client & Male & Female \\ \hline
Client 0 & 2 & 2 \\
Client 1 & 1 & 1 \\
Client 2 & 3 & 2 \\
Client 3 & 3 & 1 \\
Client 4 & 3 & 0 \\
Client 5 & 3 & 1 \\
Client 6 & 2 & 0 \\
Client 7 & 1 & 1 \\
Client 8 & 2 & 3 \\
Client 9 & 1 & 0 \\
Client 10 & 1 & 0 \\
Client 11 & 1 & 0 \\
Client 12 & 2 & 1 \\
Client 13 & 3 & 1 \\
Client 14 & 3 & 2 \\
Client 15 & 2 & 2 \\
Client 16 & 3 & 0 \\
Client 17 & 3 & 0 \\
Client 18 & 2 & 2 \\
Client 19 & 3 & 0 \\
Client 20 & 1 & 2 \\
Client 21 & 1 & 1 \\
Client 22 & 2 & 1 \\
Client 23 & 0 & 1 \\
Client 24 & 1 & 1 \\
Client 25 & 1 & 1 \\
Client 26 & 0 & 2 \\
Client 27 & 1 & 0 \\
Client 28 & 1 & 0 \\ \hline
\end{tabular}}
\end{minipage}

\vspace{0.6em}
\textbf{(b) Single \& Multi-person Clients}

\end{minipage}

\caption{Client distribution across WIDAR, HugaDB, Stress sensing and Percept-R Datasets for (a) Multi-person clients and (b) Single and Multi-person clients}
\label{tab:clients_distribution}
\end{table}
\FloatBarrier

\subsubsection{Reproducibility of Results}
\label{appendix:reproducibility}
{\color{black}
We conducted experiments using three different random seeds, with each configuration evaluated via 5-fold cross-validation. For clarity, we report representative results from a single seed in the main text. The complete results for the remaining two seeds are provided in supplementary tables: Table~\ref{tab:seeds_RQ_1} corresponds to \textbf{RQ-1}, Table~\ref{tab:seeds_RQ_2} reports results for \textbf{RQ-2} and \textbf{RQ-3}, and Table~\ref{tab:rq3_seed} presents the results for \textbf{RQ-4}.
}
\FloatBarrier
\begin{table}[H]
\centering
\color{black}
\resizebox{0.8\linewidth}{!}{
\begin{tabular}{|c|c|c|c|c|c|}
\hline
\textbf{Dataset} & \textbf{Model}    & \makecell{\textbf{Mean F1 Score $\pm$ std }\\ ($\uparrow$)}& \makecell{\textbf{Mean EO Gap $\pm$ std} \\ ($\downarrow$)} & \makecell{\textbf{Mean FATE Score $\times 10^{-1}$ }\\ ($\uparrow$)} & \makecell{\textbf{Seed}}\\ \hline
\multirow{12}{*}{WIDAR}           & FedAvg     & $0.862 \pm 0.0014$ & $0.410 \pm 0.0151$ & $0.0000$ & 123 \\ \cline{2-6}
& FedSAM     & $0.849 \pm 0.0066$ & $0.397 \pm 0.0204$ & $0.14$  & 123 \\
& FedSWA     & $\mathbf{0.858} \pm 0.0009$ & $0.403 \pm 0.0119$ & $0.10$  & 123 \\
& FedSRCVAR  & $0.728 \pm 0.0407$ & $\mathbf{0.343} \pm 0.0521$ & $0.05$ & 123 \\
& KD-FedAvg  & $0.847 \pm 0.0016$ & $0.400 \pm 0.0018$ & $0.03$  & 123 \\
& CurvFed    & $0.767 \pm 0.0383$ & $0.352 \pm 0.0121$ & $\mathbf{0.29}$ & 123  \\
\cline{2-6}
& FedAvg     & $0.862 \pm 0.0006$ & $0.412 \pm 0.0154$ & $0.0000$ & 12345 \\ \cline{2-6}
& FedSAM     & $\mathbf{0.860} \pm 0.0044$ & $0.407 \pm 0.0182$ & $0.10$  & 12345 \\
& FedSWA     & $0.859 \pm 0.0030$ & $0.405 \pm 0.0142$ & $0.15$  & 12345 \\
& FedSRCVAR  & $0.725 \pm 0.0187$ & $0.345 \pm 0.0534$ & $0.02$  & 12345 \\
& KD-FedAvg  & $0.849 \pm 0.0123$ & $\mathbf{0.392} \pm 0.0299$ & $0.33$ & 12345 \\
& CurvFed    & $0.769 \pm 0.0466$ & $0.336 \pm 0.0284$ & $\mathbf{0.77}$ & 12345 
\\ \hline

\multirow{12}{*}{Stress sensing}   & FedAvg     & $0.790 \pm 0.0041$ & $0.335 \pm 0.0324$ & $0.0000$ & 123 \\ \cline{2-6}
& FedSAM     & $\mathbf{0.800} \pm 0.0139$ & $0.321 \pm 0.0001$ & $0.56$ & 123 \\
& FedSWA     & $0.744 \pm 0.0096$ & $0.294 \pm 0.0014$ & $0.65$ & 123 \\
& FedSRCVAR  & $0.695 \pm 0.0153$ & $\mathbf{0.241} \pm 0.0154$ & $1.61$ & 123 \\
& KD-FedAvg  & $0.718 \pm 0.0254$ & $0.248 \pm 0.0045$ & $1.70$ & 123 \\
& CurvFed    & $0.793 \pm 0.0077$ & $0.254 \pm 0.0065$ & $\mathbf{2.46}$ & 123 \\
\cline{2-6}
& FedAvg     & $0.751 \pm 0.0331$ & $0.351 \pm 0.0412$ & $0.0000$ & 12345 \\ \cline{2-6}
& FedSAM     & $0.779 \pm 0.0114$ & $0.334 \pm 0.0232$ & $0.84$ & 12345 \\
& FedSWA     & $0.764 \pm 0.0084$ & $0.314 \pm 0.0021$ & $1.23$ & 12345 \\
& FedSRCVAR  & $0.675 \pm 0.0183$ & $\mathbf{0.258} \pm 0.0112$ & $1.62$ & 12345 \\
& KD-FedAvg  & $0.773 \pm 0.0087$ & $0.278 \pm 0.0097$ & $2.39$ & 12345 \\
& CurvFed    & $\mathbf{0.793} \pm 0.0065$ & $0.261 \pm 0.0007$ & $\mathbf{3.12}$ & 12345 \\ \hline

  \multirow{12}{*}{HugaDB}             & FedAvg     & $0.852 \pm 0.0253$ & $0.478 \pm 0.0446$ & $0.0000$ & 123 \\ \cline{2-6}
& FedSAM     & $0.787 \pm 0.0178$ & $0.438 \pm 0.0036$ & $0.07$  & 123 \\
& FedSWA     & $0.827 \pm 0.0090$ & $0.439 \pm 0.0138$ & $0.53$  & 123 \\
& FedSRCVAR  & $0.854 \pm 0.0176$ & $0.410 \pm 0.0097$ & $1.45$  & 123 \\
& KD-FedAvg  & $0.762 \pm 0.0132$ & $0.424 \pm 0.0154$ & $0.08$  & 123 \\
& CurvFed    & $\mathbf{0.880} \pm 0.0129$ & $\mathbf{0.406} \pm 0.0037$ & $\mathbf{1.84}$ & 123 \\
\cline{2-6} 
& FedAvg     & $0.862 \pm 0.0088$ & $0.488 \pm 0.0264$ & $0.0000$ & 12345 \\ \cline{2-6}
& FedSAM     & $0.789 \pm 0.0068$ & $0.368 \pm 0.1589$ & $0.09$  & 12345 \\
& FedSWA     & $0.825 \pm 0.0365$ & $0.379 \pm 0.1325$ & $0.23$  & 12345 \\
& FedSRCVAR  & $0.857 \pm 0.0123$ & $0.390 \pm 0.0425$ & $0.34$  & 12345 \\
& KD-FedAvg  & $0.796 \pm 0.0461$ & $\mathbf{0.374} \pm 0.1653$ & $0.01$ & 12345 \\
& CurvFed    & $\mathbf{0.892} \pm 0.0072$ & $0.406 \pm 0.0037$ & $\mathbf{0.34}$ & 12345 \\\hline

\multirow{12}{*}{PERCEPT-R} & FedAvg & $\mathbf{0.778} \pm 0.0051$ & $0.131 \pm 0.0126$ & $0.000$ & 42 \\ \cline{2-6}
 & FedSAM & $0.773 \pm 0.0059$ & $0.132 \pm 0.0173$ & $-0.11$ & 42 \\
 & FedSWA & $0.777 \pm 0.0049$ & $0.131 \pm 0.0171$ & $-0.02$ & 42 \\
 & FedSRCVAR & $0.774 \pm 0.0053$ & $0.126 \pm 0.0238$ & $0.30$ & 42 \\
 & KD-FedAvg & $0.772 \pm 0.0053$ & $0.124 \pm 0.0189$ & $0.45$ & 42 \\
 & CurvFed & $0.775 \pm 0.0073$ & $\mathbf{0.112} \pm 0.0142$ & $\mathbf{1.36}$ & 42 \\ \cline{2-6}
 & FedAvg & $0.782 \pm 0.0029$ & $0.134 \pm 0.0178$ & $0.000$ & 123 \\ \cline{2-6}
 & FedSAM & $0.779 \pm 0.0040$ & $0.133 \pm 0.0110$ & $0.08$ & 123 \\
 & FedSWA & $\mathbf{0.790} \pm 0.0061$ & $0.136 \pm 0.0118$ & $-0.66$ & 123 \\
 & FedSRCVAR & $0.776 \pm 0.0034$ & $0.130 \pm 0.0154$ & $0.25$ & 123 \\
 & KD-FedAvg & $0.779 \pm 0.0034$ & $0.153 \pm 0.0170$ & $-1.40$ & 123 \\
 & CurvFed & $0.769 \pm 0.0068$ & $\mathbf{0.116} \pm 0.0103$ & $\mathbf{1.20}$ & 123 \\ \hline
\end{tabular}
}
\caption{\color{black}{Performance metrics for federated learning models evaluated (corresponding to \textbf{RQ-1}) with different random seeds and 5-fold cross-validation across datasets. Metrics with $\uparrow$ indicate higher is better, while metrics with $\downarrow$ indicate lower is better.  \footnotesize\textit{ Note: While some methods perform better on individual F1 or EO Gap metrics, CurvFed attains the highest FATE score, reflecting the overall fairness--utility balance}}}
\label{tab:seeds_RQ_1}
\end{table}

\FloatBarrier
\begin{table}[H]

\centering
\footnotesize
\setlength{\tabcolsep}{2.5pt}
\renewcommand{\arraystretch}{1.05}
\color{black}

\begin{minipage}{0.46\textwidth}
\centering
\resizebox{\linewidth}{!}{
\begin{tabular}{|c|c|c|c|c|c|c|}
\hline
\textbf{Dataset} & \textbf{Client} & \textbf{Model} &
\makecell{\textbf{F1 Score}\\ \textbf{Mean $\pm$ Std}\\ ($\uparrow$)} &
\makecell{\textbf{EO Gap}\\ \textbf{Mean $\pm$ Std}\\ ($\downarrow$)} &
\makecell{\textbf{FATE $\times 10^{-1}$}\\ \textbf{Mean}\\ ($\uparrow$)} &
\textbf{Seed} \\ \hline

\multirow{24}{*}{\textbf{WIDAR}}

& \multirow{12}{*}{\shortstack{Single \\ \& \\ Multi \\ Person}}
& FedAvg     & $0.836 \pm 0.0165$ & $0.435 \pm 0.0114$ & $0.000$ & 123 \\ \cline{3-7}
& & FedSAM    & $0.822 \pm 0.0143$ & $0.411 \pm 0.0032$ & $0.375$ & 123 \\
& & FedSWA    & $\mathbf{0.842 \pm 0.0323}$ & $0.442 \pm 0.0034$ & $-0.087$ & 123 \\
& & FedSRCVaR & $0.716 \pm 0.0286$ & $0.391 \pm 0.0042$ & $-0.423$ & 123 \\
& & KD-FedAvg & $0.797 \pm 0.0023$ & $0.398 \pm 0.0032$ & $0.391$ & 123 \\
& & CurvFed   & $0.793 \pm 0.0089$ & $\mathbf{0.379 \pm 0.0089}$ & $\mathbf{0.772}$ & 123 \\ \cline{3-7}

& & FedAvg     & $0.841 \pm 0.0189$ & $0.431 \pm 0.0072$ & $0.000$ & 12345 \\ \cline{3-7}
& & FedSAM    & $0.835 \pm 0.0102$ & $0.422 \pm 0.0116$ & $0.149$ & 12345 \\
& & FedSWA    & $\mathbf{0.853 \pm 0.0108}$ & $0.422 \pm 0.0113$ & $0.359$ & 12345 \\
& & FedSRCVaR & $0.725 \pm 0.0098$ & $0.412 \pm 0.0092$ & $-0.926$ & 12345 \\
& & KD-FedAvg & $0.794 \pm 0.0086$ & $0.393 \pm 0.0153$ & $0.309$ & 12345 \\
& & CurvFed   & $0.791 \pm 0.0057$ & $\mathbf{0.373 \pm 0.0087}$ & $\mathbf{0.763}$ & 12345 \\ \cline{2-7}

& \multirow{12}{*}{\shortstack{Single \\ Person \\
Only}}
& FedAvg     & $0.851 \pm 0.0095$ & $0.452 \pm 0.0132$ & $0.000$ & 123 \\ \cline{3-7}
& & FedSAM    & $0.842 \pm 0.0187$ & $0.431 \pm 0.0134$ & $0.355$ & 123 \\
& & FedSWA    & $\mathbf{0.856 \pm 0.0039}$ & $0.434 \pm 0.0049$ & $0.449$ & 123 \\
& & FedSRCVaR & $0.712 \pm 0.0243$ & $0.412 \pm 0.0096$ & $-0.764$ & 123 \\
& & KD-FedAvg & $0.782 \pm 0.0096$ & $0.410 \pm 0.0172$ & $0.122$ & 123 \\
& & CurvFed   & $0.794 \pm 0.0229$ & $\mathbf{0.392 \pm 0.0059}$ & $\mathbf{0.650}$ & 123 \\ \cline{3-7}

& & FedAvg     & $\mathbf{0.857 \pm 0.0113}$ & $0.431 \pm 0.0098$ & $0.000$ & 12345 \\ \cline{3-7}
& & FedSAM    & $0.842 \pm 0.0142$ & $0.420 \pm 0.0078$ & $0.093$ & 12345 \\
& & FedSWA    & $0.852 \pm 0.0038$ & $0.424 \pm 0.0042$ & $0.126$ & 12345 \\
& & FedSRCVaR & $0.723 \pm 0.0241$ & $\mathbf{0.370 \pm 0.0049}$ & $-0.143$ & 12345 \\
& & KD-FedAvg & $0.782 \pm 0.0173$ & $0.403 \pm 0.0020$ & $-0.223$ & 12345 \\
& & CurvFed   & $0.788 \pm 0.0084$ & $0.389 \pm 0.0043$ & $\mathbf{0.180}$ & 12345 \\

\hline

\multirow{24}{*}{\textbf{\shortstack{Stress \\ Sensing}}}
& \multirow{12}{*}{\shortstack{Single \\ \& \\ Multi \\ Person}}
& FedAvg     & $0.771 \pm 0.0129$ & $0.394 \pm 0.0188$ & $0.000$ & 123 \\ \cline{3-7}
& & FedSAM    & $0.749 \pm 0.0058$ & $0.382 \pm 0.0123$ & $0.020$ & 123 \\
& & FedSWA    & $0.843 \pm 0.0151$ & $0.317 \pm 0.0096$ & $2.904$ & 123 \\
& & KD-FedAvg & $0.760 \pm 0.0192$ & $0.288 \pm 0.0092$ & $2.534$ & 123 \\
& & FedSRCVaR & $0.719 \pm 0.0091$ & $0.294 \pm 0.0146$ & $1.858$ & 123 \\
& & CurvFed   & $\mathbf{0.864 \pm 0.0098}$ & $\mathbf{0.265 \pm 0.0129}$ & $\mathbf{4.477}$ & 123 \\ \cline{3-7}

& & FedAvg     & $0.761 \pm 0.0203$ & $0.361 \pm 0.0084$ & $0.000$ & 12345 \\ \cline{3-7}
& & FedSAM    & $0.769 \pm 0.0142$ & $0.382 \pm 0.0130$ & $-0.476$ & 12345 \\
& & FedSWA    & $0.782 \pm 0.0123$ & $0.350 \pm 0.0145$ & $0.596$ & 12345 \\
& & KD-FedAvg & $0.705 \pm 0.0094$ & $0.282 \pm 0.0134$ & $1.448$ & 12345 \\
& & FedSRCVaR & $0.721 \pm 0.0078$ & $0.274 \pm 0.0198$ & $1.887$ & 12345 \\
& & CurvFed   & $\mathbf{0.812 \pm 0.0099}$ & $\mathbf{0.262 \pm 0.0199}$ & $\mathbf{3.406}$ & 12345 \\ \cline{2-7}

& \multirow{12}{*}{\shortstack{Single \\ Person \\ Only}}
& FedAvg     & $\mathbf{0.771 \pm 0.0039}$ & $0.391 \pm 0.0057$ & $0.000$ & 123 \\ \cline{3-7}
& & FedSAM    & $0.768 \pm 0.0152$ & $0.358 \pm 0.0183$ & $0.798$ & 123 \\
& & FedSWA    & $0.754 \pm 0.0092$ & $0.382 \pm 0.0126$ & $0.008$ & 123 \\
& & FedSRCVaR & $0.686 \pm 0.0115$ & $\mathbf{0.224 \pm 0.0138}$ & $3.159$ & 123 \\
& & KD-FedAvg & $0.685 \pm 0.0115$ & $0.262 \pm 0.0172$ & $2.181$ & 123 \\
& & CurvFed   & $0.743 \pm 0.0093$ & $0.233 \pm 0.0191$ & $\mathbf{3.689}$ & 123 \\ \cline{3-7}

& & FedAvg     & $\mathbf{0.764 \pm 0.0083}$ & $0.382 \pm 0.0083$ & $0.000$ & 12345 \\ \cline{3-7}
& & FedSAM    & $0.752 \pm 0.0091$ & $0.378 \pm 0.0073$ & $-0.032$ & 12345 \\
& & FedSWA    & $0.759 \pm 0.0173$ & $0.361 \pm 0.0042$ & $0.510$ & 12345 \\
& & FedSRCVaR & $0.673 \pm 0.0292$ & $\mathbf{0.229 \pm 0.0183}$ & $2.850$ & 12345 \\
& & KD-FedAvg & $0.700 \pm 0.0093$ & $0.263 \pm 0.0129$ & $2.294$ & 12345 \\
& & CurvFed   & $0.742 \pm 0.0040$ & $0.255 \pm 0.0042$ & $\mathbf{3.058}$ & 12345 \\

\hline
\end{tabular}
}
\end{minipage}
\hspace*{-3pt}
\begin{minipage}{0.49\textwidth}
\centering
\resizebox{\textwidth}{!}{
\begin{tabular}{|c|c|c|c|c|c|c|}
\hline
\textbf{Dataset} & \textbf{Client} & \textbf{Model} &
\makecell{\textbf{F1 Score}\\ \textbf{Mean $\pm$ Std}\\ ($\uparrow$)} &
\makecell{\textbf{EO Gap}\\ \textbf{Mean $\pm$ Std}\\ ($\downarrow$)} &
\makecell{\textbf{FATE $\times 10^{-1}$}\\ \textbf{Mean}\\ ($\uparrow$)} &
\textbf{Seed} \\ \hline

\multirow{24}{*}{\textbf{HugaDB}}

& \multirow{12}{*}{\shortstack{Single \\ \& \\ Multi \\ Person}}
& FedAvg     & $0.863 \pm 0.0136$ & $0.454 \pm 0.0019$ & $0.000$ & 123 \\ \cline{3-7}
& & FedSAM    & $\mathbf{0.869 \pm 0.0152}$ & $0.419 \pm 0.0021$ & $0.848$ & 123 \\
& & FedSWA    & $0.860 \pm 0.0065$ & $0.454 \pm 0.0042$ & $-0.027$ & 123 \\
& & KD-FedAvg & $0.763 \pm 0.0124$ & $0.400 \pm 0.0031$ & $0.047$ & 123 \\
& & FedSRCVaR & $0.807 \pm 0.0092$ & $\mathbf{0.384 \pm 0.0057}$ & $0.902$ & 123 \\
& & CurvFed   & $0.816 \pm 0.0098$ & $0.385 \pm 0.0013$ & $\mathbf{0.980}$ & 123 \\ \cline{3-7}

& & FedAvg     & $\mathbf{0.863 \pm 0.0231}$ & $0.441 \pm 0.0042$ & $0.000$ & 12345 \\ \cline{3-7}
& & FedSAM    & $0.848 \pm 0.0281$ & $0.413 \pm 0.0053$ & $0.544$ & 12345 \\
& & FedSWA    & $0.850 \pm 0.0192$ & $0.421 \pm 0.0093$ & $0.381$ & 12345 \\
& & KD-FedAvg & $0.781 \pm 0.0189$ & $0.419 \pm 0.0047$ & $-0.365$ & 12345 \\
& & FedSRCVaR & $0.832 \pm 0.0192$ & $0.413 \pm 0.0144$ & $0.345$ & 12345 \\
& & CurvFed   & $0.812 \pm 0.0095$ & $\mathbf{0.393 \pm 0.0186}$ & $\mathbf{0.571}$ & 12345 \\ \cline{2-7}

& \multirow{12}{*}{\shortstack{Single \\ Person \\ Only}}
& FedAvg     & $0.882 \pm 0.0139$ & $0.463 \pm 0.0049$ & $0.000$ & 123 \\ \cline{3-7}
& & FedSAM    & $0.876 \pm 0.0052$ & $0.451 \pm 0.0182$ & $0.196$ & 123 \\
& & FedSWA    & $\mathbf{0.883 \pm 0.0231}$ & $0.442 \pm 0.0086$ & $0.480$ & 123 \\
& & FedSRCVaR & $0.805 \pm 0.0163$ & $0.412 \pm 0.0212$ & $0.237$ & 123 \\
& & KD-FedAvg & $0.789 \pm 0.0124$ & $0.401 \pm 0.0089$ & $0.287$ & 123 \\
& & CurvFed   & $0.831 \pm 0.0112$ & $\mathbf{0.402 \pm 0.0125}$ & $\mathbf{0.761}$ & 123 \\ \cline{3-7}

& & FedAvg     & $\mathbf{0.873 \pm 0.0239}$ & $0.456 \pm 0.0134$ & $0.000$ & 12345 \\ \cline{3-7}
& & FedSAM    & $0.864 \pm 0.0125$ & $0.452 \pm 0.0214$ & $-0.018$ & 12345 \\
& & FedSWA    & $0.795 \pm 0.0093$ & $0.406 \pm 0.0094$ & $0.196$ & 12345 \\
& & FedSRCVaR & $0.794 \pm 0.0086$ & $0.411 \pm 0.0002$ & $0.088$ & 12345 \\
& & KD-FedAvg & $0.773 \pm 0.0179$ & $\mathbf{0.398 \pm 0.0041}$ & $0.136$ & 12345 \\
& & CurvFed   & $0.832 \pm 0.0107$ & $0.401 \pm 0.0085$ & $\mathbf{0.746}$ & 12345 \\

\hline

\multirow{24}{*}{\textbf{PERCEPT-R}} & \multirow{12}{*}{\shortstack{Single \\ \& \\ Multi \\ Person}} & FedAvg & $0.778 \pm 0.0068$ & $0.114 \pm 0.0130$ & $0.000$ & 42 \\  \cline{3-7}
 &  & FedSAM & $0.781 \pm 0.0067$ & $0.109 \pm 0.0106$ & $0.43$ & 42 \\
 &  & FedSWA & $0.781 \pm 0.0051$ & $0.113 \pm 0.0115$ & $0.09$ & 42 \\
 &  & FedSRCVAR & $\mathbf{0.784} \pm 0.0050$ & $0.111 \pm 0.0072$ & $0.37$ & 42 \\
 &  & KD-FedAvg & $0.779 \pm 0.0078$ & $0.114 \pm 0.0106$ & $0.003$ & 42 \\
 &  & CurvFed & $0.775 \pm 0.0056$ & $\mathbf{0.103} \pm 0.0107$ & $\mathbf{0.96}$ & 42 \\ \cline{3-7}
 &  & FedAvg & $0.782 \pm 0.0063$ & $0.122 \pm 0.0169$ & $0.000$ & 123 \\  \cline{3-7}
 &  & FedSAM & $0.779 \pm 0.0044$ & $0.114 \pm 0.0197$ & $0.62$ & 123 \\
 &  & FedSWA & $0.780 \pm 0.0043$ & $0.113 \pm 0.0221$ & $0.73$ & 123 \\
 &  & FedSRCVAR & $\mathbf{0.786} \pm 0.0059$ & $0.122 \pm 0.0154$ & $0.05$ & 123 \\
 &  & KD-FedAvg & $0.782 \pm 0.0056$ & $0.127 \pm 0.0173$ & $-0.37$ & 123 \\
 &  & CurvFed & $0.780 \pm 0.0106$ & $\mathbf{0.110} \pm 0.0138$ & $\mathbf{0.93}$ & 123 \\ \cline{2-7}
 & \multirow{12}{*}{\shortstack{Single \\Person \\ Only}} & FedAvg & $\mathbf{0.772} \pm 0.0039$ & $0.105 \pm 0.0067$ & $0.000$ & 42 \\  \cline{3-7}
 &  & FedSAM & $0.770 \pm 0.0039$ & $0.103 \pm 0.0159$ & $0.15$ & 42 \\
 &  & FedSWA & $0.766 \pm 0.0038$ & $0.101 \pm 0.0152$ & $0.25$ & 42 \\
 &  & FedSRCVAR & $0.768 \pm 0.0047$ & $0.105 \pm 0.0078$ & $-0.09$ & 42 \\
 &  & KD-FedAvg & $0.772 \pm 0.0043$ & $0.104 \pm 0.0057$ & $0.04$ & 42 \\
 &  & CurvFed & $0.768 \pm 0.0106$ & $\mathbf{0.077} \pm 0.0191$ & $\mathbf{2.66}$ & 42 \\ \cline{3-7}
 &  & FedAvg & $0.766 \pm 0.0093$ & $0.105 \pm 0.0165$ & $0.000$ & 123 \\  \cline{3-7}
 &  & FedSAM & $\mathbf{0.771} \pm 0.0033$ & $0.105 \pm 0.0066$ & $0.09$ & 123 \\
 &  & FedSWA & $0.769 \pm 0.0037$ & $0.110 \pm 0.0075$ & $-0.40$ & 123 \\
 &  & FedSRCVAR & $0.766 \pm 0.0058$ & $0.103 \pm 0.0144$ & $0.22$ & 123 \\
 &  & KD-FedAvg & $0.769 \pm 0.0062$ & $0.109 \pm 0.0088$ & $-0.32$ & 123 \\
 &  & CurvFed & $0.751 \pm 0.0050$ & $\mathbf{0.068} \pm 0.0163$ & $\mathbf{3.33}$ & 123 \\ \hline
\end{tabular}
}
\end{minipage}
\caption{\color{black}{Performance metrics for all datasets evaluated across different client compositions (corresponding to \textbf{RQ-2, RQ-3}) and random seeds (5-fold cross-validation). Metrics with $\uparrow$ indicate higher is better, while $\downarrow$ indicates lower is better. \footnotesize\textit{Note: CurvFed consistently achieves the highest FATE score.}}}
\label{tab:seeds_RQ_2}
\end{table}
\FloatBarrier

\begin{table}[t]
\centering
\color{black}

\begin{minipage}{0.48\linewidth}
\centering
\resizebox{\linewidth}{!}{
\begin{tabular}{|c|c|c|c|c|}
\hline
\multicolumn{5}{|c|}{\textbf{EDA--Hand}} \\ \hline
\textbf{Model} &
\makecell{\textbf{EO Gap}\\ \textbf{Mean $\pm$ Std}\\ ($\downarrow$)} &
\makecell{\textbf{F1 Score}\\ \textbf{Mean $\pm$ Std}\\ ($\uparrow$)} &
\makecell{\textbf{FATE $\times 10^{-1}$}\\ \textbf{Mean}\\ ($\uparrow$)} &
\textbf{Seed} \\ \hline

FedAvg   & $0.453 \pm 0.0141$ & $0.774 \pm 0.0098$ & $0.00$ & 123 \\
FedSAM   & $0.443 \pm 0.0183$ & $0.795 \pm 0.0164$ & $0.48$ & 123 \\
FedSWA   & $0.451 \pm 0.0094$ & $0.740 \pm 0.0006$ & $-0.40$ & 123 \\
FedSRCVaR& $\mathbf{0.389 \pm 0.0141}$ & $0.645 \pm 0.0153$ & $-0.27$ & 123 \\
KD       & $0.393 \pm 0.0139$ & $0.735 \pm 0.0027$ & $0.83$ & 123 \\
CurvFed     & $0.430 \pm 0.0052$ & $\mathbf{0.805 \pm 0.0019}$ & $\mathbf{0.90}$ & 123 \\ \hline

FedAvg   & $0.476 \pm 0.0094$ & $0.754 \pm 0.0067$ & $0.00$ & 12345 \\
FedSAM   & $0.476 \pm 0.0077$ & $0.775 \pm 0.0012$ & $0.28$ & 12345 \\
FedSWA   & $0.446 \pm 0.0024$ & $0.745 \pm 0.0169$ & $0.50$ & 12345 \\
FedSRCVaR& $\mathbf{0.391 \pm 0.0211}$ & $0.695 \pm 0.0163$ & $0.99$ & 12345 \\
KD       & $0.417 \pm 0.0192$ & $0.718 \pm 0.0154$ & $0.75$ & 12345 \\
CurvFed    & $0.423 \pm 0.0023$ & $\mathbf{0.789 \pm 0.0088}$ & $\mathbf{1.57}$ & 12345 \\ \hline
\end{tabular}
}
\end{minipage}
\hfill
\begin{minipage}{0.48\linewidth}
\centering
\resizebox{\linewidth}{!}{
\begin{tabular}{|c|c|c|c|c|}
\hline
\multicolumn{5}{|c|}{\textbf{WIDAR--Orientation}} \\ \hline
\textbf{Model} &
\makecell{\textbf{EO Gap}\\ \textbf{Mean $\pm$ Std}\\ ($\downarrow$)} &
\makecell{\textbf{F1 Score}\\ \textbf{Mean $\pm$ Std}\\ ($\uparrow$)} &
\makecell{\textbf{FATE $\times 10^{-1}$}\\ \textbf{Mean}\\ ($\uparrow$)} &
\textbf{Seed} \\ \hline

FedAvg   & $0.492 \pm 0.0098$ & $0.848 \pm 0.0015$ & $0.00$ & 123 \\
FedSAM   & $0.478 \pm 0.0141$ & $0.833 \pm 0.0195$ & $0.11$ & 123 \\
FedSWA   & $0.508 \pm 0.0472$ & $\mathbf{0.878 \pm 0.0045}$ & $0.03$ & 123 \\
FedSRCVaR& $0.433 \pm 0.0579$ & $0.747 \pm 0.0115$ & $0.01$ & 123 \\
KD       & $0.443 \pm 0.0332$ & $0.846 \pm 0.0123$ & $0.96$ & 123 \\
CurvFed     & $\mathbf{0.374 \pm 0.0605}$ & $0.805 \pm 0.0481$ & $\mathbf{1.89}$ & 123 \\ \hline

FedAvg   & $0.513 \pm 0.0021$ & $\mathbf{0.860 \pm 0.0028}$ & $0.00$ & 12345 \\
FedSAM   & $0.465 \pm 0.0168$ & $0.844 \pm 0.0146$ & $0.76$ & 12345 \\
FedSWA   & $0.492 \pm 0.0002$ & $0.833 \pm 0.0069$ & $0.09$ & 12345 \\
FedSRCVaR& $0.427 \pm 0.0660$ & $0.729 \pm 0.0102$ & $0.17$ & 12345 \\
KD       & $0.469 \pm 0.0369$ & $0.858 \pm 0.0010$ & $0.85$ & 12345 \\
CurvFed     & $\mathbf{0.384 \pm 0.0085}$ & $0.812 \pm 0.0056$ & $\mathbf{1.96}$ & 12345 \\ \hline
\end{tabular}
}
\end{minipage}

\caption{\textcolor{black}{Multiple seed Fairness–efficacy balance under different sensitive attributes for RQ-4.Results are averaged over 5-fold cross-validation. Lower EO indicates improved fairness, while higher F1 and FATE scores reflect a better balance between predictive performance and fairness. CurvFed consistently achieves the highest FATE score.}}
\label{tab:rq3_seed}
\end{table}

\section{Runtime benchmarking vs baseline approaches}
\label{benchmarking_old}
To complement the real-world study, we benchmarked \textbf{\textit{CurvFed}} against standard FL baselines (FedAvg, FedSAM) on \textit{seven} deployment platforms—from high-performance GPUs (NVIDIA RTX 4090) to resource-constrained devices (Raspberry Pi 5, Pixel 6) to assess real-world feasibility. Across five runs using the WIDAR dataset, we measured training time, memory usage, resource utilization, and inference time on average. As shown in Table \ref{tab:combined_time}, \textbf{\textit{CurvFed}} incurs moderate training overhead due to the Fisher penalty, similar to FedSAM’s cost for sharpness-aware updates. However, this overhead is practical—even on edge devices—and does not affect inference time. Methods like FedSWA and FedSRCVaR were excluded as their client-side costs mirror FedAvg, with additional computations handled server-side. Overall, \textbf{\textit{CurvFed}} demonstrates efficiency and feasibility across heterogeneous edge devices.

\begin{table}[h!]
\centering
\resizebox{0.6\linewidth}{!}{%
\begin{tabular}{|c|c|c|c|c|c|}
\hline
\textbf{Platform} & \textbf{Approach} & \textbf{Train (s)} & \textbf{RSS (MBs)} & \textbf{Infer (s)} & \textbf{ \makecell{ Train \\Usage (\%)}} \\ \hline
\multirow{3}{*}{ \makecell{Intel\\ i9-9900K \\ Processor}} & Fedavg & 1.67 & 1074.05 & 0.03 & 48.12 \\ \cline{2-6} 
 & FedSAM & 5.78 &  1147.54 & 0.02 & 48.07 \\ \cline{2-6} 
 & CurvFed & 9.34 & 1287.87 & 0.02 & 48.134\\ \cline{2-6} \hline

\multirow{3}{*}{ \makecell{AMD Ryzen 9 \\ 7950X 16-Core \\ Processor}} & Fedavg & 0.77 & 1033.62 & 0.01 & 24.93 \\ \cline{2-6} 
 & FedSAM & 3.23 &  1109.04 & 0.01 & 24.30 \\ \cline{2-6} 
 & CurvFed & 5.42 & 1420.88 & 0.01 & 24.52\\ \cline{2-6} \hline

\multirow{3}{*}{Apple M1} & Fedavg & 4.02 & 418.916 & 0.16 & 1.05\\ \cline{2-6} 
 & FedSAM & 8.95 & 450.31 & 0.16 & 1.06\\ \cline{2-6} 
 & CurvFed  & 11.53 & 387.36 & 0.16 & 1.00\\ \cline{2-6} \hline
\multirow{3}{*}{\makecell{Quadro RTX \\ 5000}} & Fedavg & 0.73 & 961.77 & 0.01 & 46.4 \\ \cline{2-6} 
 & FedSAM & 0.81 & 1070.73 & 0.01 & 73.0 \\ \cline{2-6} 
 & CurvFed & 1.04 & 1176.54 & 0.01 & 62.3 \\ \cline{2-6}
 \hline
\multirow{3}{*}{\makecell{NVIDIA GeForce \\ RTX 4090}} & Fedavg & 0.14 & 954.52 & 0.01 & 15.0 \\ \cline{2-6} 
 & FedSAM & 0.18 & 1063 & 0.01 & 49.1\\ \cline{2-6} 
 & CurvFed & 0.32 & 1290.88 & 0.01 & 48.6\\ \cline{2-6}
 \hline
 \multirow{3}{*}{\makecell{Google Pixel 6}} & Fedavg & 4.24  & 370.84 & 0.12  & 49.11 \\ \cline{2-6} 
 & FedSAM & 13.39 & 337.40 & 0.12 & 49.11 \\ \cline{2-6} 
 & CurvFed & 26.59 &  326.28 & 0.20  & 49.07\\ \cline{2-6}
 \hline
  \multirow{3}{*}{\makecell{Raspberry Pi 5}} & Fedavg & 13.5 & 518.88  & 0.21  & 48.79 \\ \cline{2-6}  
 & FedSAM & 40.8 & 489 & 0.23  & 48.7 \\ \cline{2-6} 
 & CurvFed & 56.4 & 490  & 0.22 & 49 \\ \cline{2-6}
 \hline
\end{tabular}%
}
\caption{ Training Time (s), Training Process (RSS) Memory Usage (MBs), Inference Time (s), Training Resource usage (\%) \textcolor{black}{averaged for \textit{five} runs} for FL Approaches across Different Platforms}
\label{tab:combined_time}
\vspace{-15pt}
\end{table}

\subsubsection{\textbf{Resource Utilization and Overhead Breakdown}}
\label{sec:resource-overhead}

We measured the per-batch average training time and Fisher top-eigenvalue computation time \textcolor{black}{(averaged over five runs)} across all our evaluation platforms using a single client from the WIDAR dataset. Results are summarized in Table \ref{tab:fisher_overhead}. While Fisher computation introduces additional cost (typically $\sim$30--40\% of per-batch training time), the absolute time per call remains modest (milliseconds on GPUs and tens to hundreds of milliseconds on CPUs and mobile devices). These measurements indicate that Fisher overhead scales proportionally with device compute capability and remains practical within the constraints of on-device FL training.

\begin{table}[!ht]
\centering
\resizebox{0.7\linewidth}{!}{
\begin{tabular}{|c|c|c|}
\hline
\textbf{Platform} & \textbf{Avg. Train Time / batch (s)} & \textbf{Avg. Fisher Time / batch (s)} \\
\hline
GPU: Quadro RTX 5000 & 0.027 & 0.010 \\
GPU: NVIDIA RTX 4090 & 0.003 & 0.001 \\
CPU: AMD Ryzen 9 7950X & 0.076 & 0.029 \\
Apple M1 & 0.298 & 0.137 \\
Pixel 6: Cortex-A55 & 0.355 & 0.146 \\
CPU: Intel i9-9900K & 0.169 & 0.062 \\
CPU: Intel i5-14600T & 0.268 & 0.175 \\
Jetson Nano: Cortex-A57 & 1.040 & 0.393 \\
Raspberry Pi: Cortex-A76 & 1.621 & 0.733 \\
\hline
\end{tabular}
}
\caption{Average per-batch training time and Fisher computation time across platforms.}
\label{tab:fisher_overhead}
\vspace{-10pt}
\end{table}

To contextualize these results, we also benchmarked the upload and download latency for a trained model ($\sim$34.4 MB) across CPU-based platforms, using Google Cloud APIs with PyDrive authentication (Table~\ref{tab:network_latency}). Communication latency per round is on the order of \textit{seconds}, which is \textit{one to two orders of magnitude higher than the Fisher computation overhead per batch}. Since such communication is repeated across many FL rounds (e.g., 80 rounds for WIDAR), network transfer typically dominates the end-to-end runtime \cite{le2024exploring}.

\begin{table}[!ht]
\centering
\resizebox{0.6\linewidth}{!}{
\begin{tabular}{|c|c|c|c|}
\hline
\textbf{Platform} & \textbf{File Size (MB)} & \textbf{Upload (s)} & \textbf{Download (s)} \\
\hline
CPU: AMD Ryzen 9 7950X & 34.39 & 2.30 & 6.42 \\
Jetson Nano: Cortex-A57 & 34.39 & 2.83 & 4.43 \\
Raspberry Pi: Cortex-A76 & 34.39 & 2.42 & 3.96 \\
CPU: Intel i5-14600T & 34.39 & 3.32 & 3.80 \\
Pixel 6: Cortex-A55 & 34.39 &  5.05 & 4.73 \\
CPU: Intel i9-9900K & 34.39 & 2.36 & 5.41 \\
\hline
\end{tabular}
}
\caption{Upload and download latency for a trained model ($\sim$34.4 MB) across CPU platforms.}
\label{tab:network_latency}
\vspace{-10pt}
\end{table}

These experiments suggest that while Fisher computation contributes a consistent overhead during local training, communication latency remains the dominant factor in federated learning deployments \cite{zhu2021delayed,le2024exploring}. This indicates that curvature-based regularization is computationally feasible in practice \cite{khan2025functional}, with networking emerging as the primary bottleneck. Exploring lighter curvature proxies \cite{wang2024zeroth} and adaptive scheduling \cite{zhang2023timelyfl} remains a valuable direction for further improving scalability.

\section{Experiment Details}
\label{Appendix:exp}

\subsection{Ablation Study on Additional Datasets}

\label{model_performance_abalation}
We conducted a comprehensive ablation study to meticulously guide the choices in our proposed approach. The study consistently demonstrates the effectiveness of our model across all datasets. \textcolor{black}{This section presents the results for the Stress sensing, HugaDB and the WIDAR dataset in Table \ref{tab:model_performance_aba}. The results for PERCEPT-R were reported in the main text under Section \ref{ablation-study}}

\FloatBarrier
\begin{table}[H]
\centering
\resizebox{0.99\linewidth}{!}{
\begin{tabular}{|l|c|c|c|c|c|c|}
\hline
\textbf{\#} &
\textbf{Dataset} &
\makecell{\textbf{Client-Side}\\ \textbf{Training}} &
\makecell{\textbf{Aggregation}\\ \textbf{Strategy}} &
\makecell{\textbf{F1 Score}\\ ($\uparrow$)} &
\makecell{\textbf{EO Gap}\\ ($\downarrow$)} &
\makecell{\textbf{FATE Score} $\times 10^{-1}$\\ ($\uparrow$)} \\
\hline

1 & \multirow{12}{*}{WIDAR} & Benign & FedAvg & 0.8507 & 0.4263 & 0 \\
2 &  & SAM & FedAvg & 0.8005 & 0.4251 & -0.560 \\
3 &  & SAM & FedSWA & 0.8405 & 0.4071 & 0.331 \\
4 &  & SAM + Eq.~\ref{local_training_eq} & FedSWA & 0.8459 & 0.4058 & 0.424 \\
5 &  & (Eq11) & FedAvg & 0.8253 & 0.3994 & 0.332 \\
6 &  & (Eq11) & $W^{r}$ (Eq12) & 0.8344 & 0.4135 & 0.109 \\
7 &  & SAM + Eq.~\ref{local_training_eq} & $W^{r}$ (Eq12) & 0.8707 & 0.4199 & 0.386 \\
8 &  & (Eq11) & Sharpness-Aware Aggregation & 0.8409 & 0.3982 & 0.545 \\
9 &  & SAM & FedSWA + S(L) (Eq12* partial) & \textbf{0.8737} & 0.4199 & 0.420 \\
10 & & SAM & FedSWA + S(T) (Eq12* partial) & 0.8590 & 0.4084 & 0.518 \\
11 & & SAM & Sharpness-Aware Aggregation & 0.8668 & 0.4161 & 0.429 \\
12 & & SAM + Eq.~\ref{local_training_eq} & Sharpness-Aware Aggregation & 0.8100 & \textbf{0.3533} & \textbf{1.233} \\
\hline

13 & \multirow{12}{*}{Stress Sensing} & Benign & FedAvg & 0.7253 & 0.3069 & 0 \\
14 &  & SAM & FedAvg & 0.7758 & 0.3034 & 0.812 \\
15 &  & SAM & FedSWA & 0.7406 & 0.3015 & 0.384 \\
16 &  & SAM + Eq.~\ref{local_training_eq} & FedSWA & 0.7346 & 0.2768 & 1.105 \\
17 &  & (Eq.~\ref{local_training_eq}) & FedAvg & 0.7372 & 0.2998 & 0.395 \\
18 &  & (Eq.~\ref{local_training_eq}) & $W^{r}$ (Eq12) & 0.7345 & 0.2839 & 0.874 \\
19 &  & SAM + Eq.~\ref{local_training_eq} & $W^{r}$ (Eq12) & \textbf{0.8193} & 0.2927 & 1.756 \\
20 &  & (Eq.~\ref{local_training_eq}) & Sharpness-Aware Aggregation & 0.7077 & \textbf{0.2380} & 1.999 \\
21 &  & SAM & FedSWA + S(L) (Eq12* partial) & 0.8091 & 0.2680 & 2.240 \\
22 &  & SAM & FedSWA + S(T) (Eq12* partial) & 0.7995 & 0.2627 & 2.460 \\
23 &  & SAM & Sharpness-Aware Aggregation & 0.8224 & 0.2710 & 2.507 \\
24 &  & SAM + Eq.~\ref{local_training_eq} & Sharpness-Aware Aggregation & 0.7814 & 0.2522 & \textbf{2.555} \\
\hline
25 & \multirow{12}{*}{HugaDB} & Benign & FedAvg & 0.8496 & 0.4369 & 0 \\
26 &  & SAM & FedAvg & 0.8555 & 0.4255 & 0.329 \\
27 &  & SAM & FedSWA & 0.8418 & 0.4242 & 0.192 \\
28 &  & SAM + Eq.~\ref{local_training_eq} & FedSWA & 0.8170 & 0.4142 & 0.135 \\
29 &  & (Eq.~\ref{local_training_eq}) & FedAvg & 0.8567 & 0.4142 & 0.602 \\
30 &  & (Eq.~\ref{local_training_eq}) & $W^{r}$ (Eq12) & 0.8574 & 0.4083 & 0.746 \\
31 &  & SAM + Eq.~\ref{local_training_eq} & $W^{r}$ (Eq12) & \textbf{0.9016} & 0.4154 & 1.103 \\
32 &  & (Eq.~\ref{local_training_eq}) & Sharpness-Aware Aggregation & 0.8510 & 0.4333 & 0.098 \\
33 &  & SAM & FedSWA + S(L) (Eq12* partial) & 0.8934 & 0.4130 & 1.059 \\
34 &  & SAM & FedSWA + S(T) (Eq12* partial) & 0.8888 & 0.4142 & 0.878 \\
35 &  & SAM & Sharpness-Aware Aggregation & 0.8911 & 0.4095 & 1.110 \\
36 &  & SAM + Eq.~\ref{local_training_eq} & Sharpness-Aware Aggregation & 0.8855 & \textbf{0.3952} & \textbf{1.376} \\

\hline
\end{tabular}
}
\caption{\textcolor{black}{Ablation of \textbf{\textit{CurvFed}}’s design choices on the WIDAR, Stress Sensing, and HugaDB datasets.
\textbf{Client-Side Training} denotes the optimization or regularization applied locally at each client, while
\textbf{Aggregation Strategy} specifies the server-side model aggregation rule.
Unless explicitly stated, each configuration applies \emph{only} the listed component, with no additional modifications.
Metrics with $\uparrow$ indicate higher is better, while $\downarrow$ indicates lower is better.}}
\label{tab:model_performance_aba}
\end{table}

\textcolor{black}{We begin by establishing benign baselines using standard FedAvg aggregation without fairness-aware training (Rows~1, 13, and~25).
Across datasets, these configurations achieve competitive F1 scores (0.8507 on WIDAR, 0.7253 on Stress Sensing, and 0.8496 on HugaDB) but are consistently associated with relatively high EO Gaps (0.4263, 0.3069, and 0.4369, respectively), highlighting the presence of fairness concerns under standard federated optimization.}

\textcolor{black}{Introducing SAM under FedAvg aggregation (Rows~2, 14, and~26) generally improves or maintains accuracy, most notably on Stress Sensing (F1 = 0.7758, Row~14), but yields only marginal changes in EO Gap.}
\textcolor{black}{Replacing FedAvg with FedSWA while retaining SAM (Rows~3, 15, and~27) produces mixed effects, with modest reductions in EO Gap on WIDAR and Stress Sensing, but limited improvements on HugaDB, indicating that smoother aggregation alone is insufficient to consistently address bias.}

\textcolor{black}{We next examine the effect of enforcing curvature-aware local training without SAM.
Applying the FIM-based regularization during client-side training with FedAvg aggregation (Rows~5, 17, and~29) consistently lowers EO Gap compared to the benign baselines, while also reducing F1 scores.
Replacing FedAvg with curvature-aware weighting $W^r$ (Rows~6, 18, and~30) further modifies this trade-off, sometimes improving EO Gap (e.g., Row~18 on Stress Sensing) while slightly increasing F1 score relative to FedAvg-based curvature regularization.
These results indicate that restricting local optimization to flatter regions promotes fairness, albeit at the cost of predictive performance when applied in isolation.}

\textcolor{black}{Combining SAM with curvature-aware local training and weighted aggregation (Rows~7, 19, and~31) shifts the balance toward higher accuracy.
In particular, these configurations achieve some of the highest F1 scores across datasets, including the maximum F1 on HugaDB (0.9016, Row~31). However, EO Gaps remain relatively high in these settings, suggesting that SAM-driven optimization can counteract some fairness gains introduced by curvature regularization.}

\textcolor{black}{We then analyze aggregation-centric interventions. Applying sharpness-aware aggregation without SAM or curvature-aware local training (Rows~8, 20, and~32) yields noticeable reductions in EO Gap across datasets, most prominently on Stress Sensing (EO Gap = 0.2380, Row~20), but often with a reduction in F1 score.}
\textcolor{black}{Partial sharpness-aware aggregation strategies under SAM, which prioritize either predictive uncertainty or classification loss (Rows~9–10, 21–22, and~33–34), achieve strong F1 scores—such as the highest F1 on WIDAR (0.8737, Row~9)—but do not consistently minimize EO Gap.}

\textcolor{black}{Employing the full sharpness-aware aggregation under SAM (Rows~11, 23, and~35) provides a more balanced outcome, achieving competitive F1 scores alongside moderate EO Gaps across all datasets.
These results indicate that aggregation strategies that explicitly account for both loss and curvature information can mitigate bias more effectively than partial criteria.}

\textcolor{black}{Finally, the full \textbf{\textit{CurvFed}} configuration—integrating curvature-aware client-side training with sharpness-aware aggregation—consistently achieves the most favorable fairness–utility trade-off.
On WIDAR (Row~12), this configuration yields the lowest EO Gap (0.3533) and the highest FATE score (1.233). On Stress Sensing (Row~24), it achieves an EO Gap of 0.2522 with the highest FATE score (2.555).
Similarly, on HugaDB (Row~36), it attains the lowest EO Gap (0.3952) and the highest FATE score (1.376), while maintaining a strong F1 score (0.8855).
Although higher absolute F1 scores are observed in some partial configurations (e.g., Row~31), the full \textbf{\textit{CurvFed}} model consistently offers the most balanced performance across datasets.}

\end{document}


\title{Curvature-Aligned Federated Learning (CAFe)}


\renewcommand{\shortauthors}{Trovato et al.}

\begin{abstract}

\end{abstract}
\maketitle

\appendix
\label{appendix}
\section{Methodological Transparency \& Reproducibility Appendix (META)}

\subsection{Dataset and Model Description}

\label{Appendix:data_label}
This section provides a detailed overview of the dataset and the models we utilized in our evaluations.
\subsubsection{Dataset Description}
\label{appen:data}
\textbf{WIDAR Dataset:} The WIDAR3.0 dataset \cite{zhang2021WIDAR3} contains WIFI channel state information (CSI) \cite{kaurwireless} which represents the variation in the Wifi channel portrayed by commodity wifi devices \cite{kaurwireless,zhang2021WIDAR3}. The WIFI CSI contains fine-grained information about variations in the Wifi channel based on the Doppler phenomenon \cite{zhang2021WIDAR3}. These signals have been used by prior researchers for gesture detection \cite{zhang2021WIDAR3}, presence detection \cite{sathya2023systematic}, pose estimation \cite{wang2018csi}, human identification and activity detection \cite{shahbazian2023human}. We have used it as one of the HAR (Human Activity Recognition) datasets. The dataset contains data from 16 participants, 4 Females, and 12 Males, with five different orientations (labeled as 1 to 5).
We removed a female participant from the dataset due to the absence of ground truth. The selected two gestures, `Slide' and `Sweep,' had data in all orientations for male and female participants in the presence of the wifi sensor. Since all participants had an equal number of samples in different orientations, i.e., orientations 1 to 5, we created a disparity in the dataset based on orientations by randomly sub-sampling 50\% of the data belonging to orientations 4 and 5. This was done to create a training imbalance representative of real-life scenarios where some groups have more representation compared to others. We considered labels 1 to 3 as major orientation and the rest as minor orientation. For our work, we have trained a model to perform binary classification to detect Sweep vs. Slide gestures using WIFI-CSI data and fairness on the basis of `sex' and `orientation' as sensitive attributes. 

\textbf{HuGaDB Dataset:}
We have also utilized the HuGaDB (Human Gait Database) dataset \cite{chereshnev2018hugadb} for HAR purposes, which comprises a diverse array of human activity recordings from 18 participants. The participant group includes 4 females and 14 males. In our work, we trained a binary classification model to distinguish between Standing and other activities from this dataset, with fairness considerations based on `sex' as the sensitive attribute.

\textbf{Stress sensing Dataset:}
Xiao et al. \cite{xiao2024reading} collected this dataset. This dataset
was obtained from 48 individuals, including 40 females and 8 males, who participated in four different stress-inducing tasks (Arithmetic, Bad Memory, Stress Videos, Pre-condition ), each associated with unique stressors. The dataset is not publicly available yet; Access to the dataset has been obtained through proper IRB approvals/extensions. Due to the continuation of data collection by the Xiao et al. \cite{xiao2024reading}'s research group, this paper's reported dataset has more individuals than reported by the Xiao et al. \cite{xiao2024reading} paper. The dataset has individuals wearing Empatica E4 \cite{mccarthy2016validation} (collecting Stresssensing data) in their left and/or right hands and individuals' sex information. Hence, this paper utilized `hand' information and `sex' as sensitive attributes for evaluation. 
Utilizing this dataset, we performed a binary stress detection classification task.

{{\subsubsection{Client Distribution:}
\label{appendix:client}
Distinct client distributions are established for each dataset in our study. }

\begin{itemize}

  \item To effectively benchmark the performance of our proposed model and ensure clients with diverse participant groups, we deliberately structured the client distribution to include at least one participant from each sensitive attribute category. Table \ref{tab:clients} provides a detailed overview of participant distribution across clients, categorizing them based on the sensitive attributes of males and females.

  \item Given that our models do not depend on sensitive attribute information during training, we also demonstrate their effectiveness by emulating the approach in clients with either single and/or multiple participants. This intentional diversification, as depicted in Table \ref{tab:one_person_clients}, does not guarantee the presence of both sensitive attributes within each client.

\end{itemize}

\begin{table}[h]
\centering
\begin{tabular}{|l|c|c|c|}
\hline
\textbf{Dataset} & \textbf{Clients} & \textbf{Male} & \textbf{Female} \\
\hline
 & Client 1 & 4 & 1 \\
WIDAR & Client 2 & 4 & 1 \\
 & Client 3 & 4 & 1 \\
\hline

 & Client 1 & 4 & 1 \\
HugaDB & Client 2 & 4 & 1 \\
 & Client 3 & 4 & 1 \\
 & Client 4 & 2 & 1 \\
\hline

 & Client 1 & 2 & 4 \\
 & Client 2 & 4 & 2 \\
 & Client 3 & 1 & 6 \\
Stress sensing & Client 4 & 1 & 6 \\
 & Client 5 & 2 & 3 \\
 & Client 6 & 1 & 7 \\
 & Client 7 & 3 & 6 \\
\hline
\end{tabular}
\caption{Random distribution of multi-person clients of WIDAR, HugaDB, and Stress sensing Datasets.}
\label{tab:clients}
\end{table}

\begin{table}[h]
\centering
\begin{tabular}{|l|c|c|c|}
\hline
\textbf{Dataset} & \textbf{Clients} & \textbf{Male} & \textbf{Female} \\
\hline
 & Client 1 & 1 & 1 \\
 & Client 2 & 2 & 1 \\
WIDAR & Client 3 & 1 & 0 \\
 & Client 4 & 2 & 0 \\
 & Client 5 & 2 & 1 \\
 & Client 6 & 1 & 0 \\
 & Client 7 & 1 & 0 \\
 & Client 8 & 2 & 0 \\
\hline
 & Client 1 & 6 & 0 \\
HugaDB & Client 2 & 5 & 1 \\
 & Client 3 & 0 & 2 \\
 & Client 4 & 0 & 1 \\
\hline
 & Client 1 & 1 & 0 \\
 & Client 2 & 3 & 0 \\
 & Client 3 & 6 & 10 \\
Stress sensing& Client 4 & 0 & 6 \\
 & Client 5 & 3 & 7 \\
 & Client 6 & 2 & 9 \\
 & Client 7 & 0 & 1 \\
\hline
\end{tabular}
\caption{Single and multi-person clients distribution across WIDAR, HugaDB and Stress sensing Datasets}
\label{tab:one_person_clients}
\end{table}

\subsubsection{Model Description}
\label{Appendix:model_desc}

\textbf{WIDAR Model:} The WIDAR Model architecture (adapted from \cite{yang2023benchmark}) is comprised of two fully connected layers within a sequential module, featuring an input size of 22 * 20 * 20. A dropout layer with a 40\% dropout rate and a Rectified Linear Unit (ReLU) activation function follow the first linear layer, while the second linear layer outputs the classification result. The model reshapes the input tensor during the forward pass and incorporates dropout layers to mitigate over-fitting. 

\textbf{HugaDB Model:}
The HugaDB model architecture (adapted from \cite{pirttikangas2006feature}) is characterized by its complexity, featuring two sequential fully connected layers. The first layer consists of linear transformations with input and output sizes of 792 to 1024, followed by dropout with a 30\% probability, Rectified Linear Unit (ReLU) activation, and Instance Normalization. Subsequently, another linear layer reduces the dimensionality to 512, followed by similar dropout, ReLU, and Instance Normalization. The second layer further reduces the dimensionality from 512 to 128, followed by a linear layer with an output size equal to the specified number of classes. The final output is obtained by applying the sigmoid function to the result. During the forward pass, the input tensor is processed through these layers, reshaped when necessary, and the output is returned after applying the sigmoid function. 

\textbf{Stress sensing Model:}
The Stress sensing model architecture (adapted from \cite{sharma2022psychophysiological}) comprises three sequential fully connected layers. The first layer takes input of size 34 and produces an output of size 64, incorporating batch normalization and Rectified Linear Unit (ReLU) activation. The second layer further processes the output, reducing it to size 128 with ReLU activation. The final layer transforms the output to the specified embedding size.

\subsubsection{Hyperparameters}
\label{appendix:hyperparameters}

We have produced results from three different seed sets and presented one of them here. The complete set of results for all three seeds can be found in Table \ref{tab:seed}.

\begin{table}[h!]
\centering
\begin{tabular}{cccccc}
\toprule
\textbf{Dataset} & \textbf{F1 Score} & \textbf{Accuracy} & \textbf{EO Gap} & \textbf{Model} & \textbf{Seed} \\
\midrule
 & 0.8568 & 0.8501 & 0.4192 & fedavg & 123 \\
 & 0.8702 & 0.8654 & 0.3988 & CAFe & 123 \\
HugaDB & 0.8496 & 0.8425 & 0.4369 & fedavg & 16 \\
 & 0.8856 & 0.8819 & 0.3952 & CAFe & 16 \\
 & 0.8548 & 0.8476 & 0.4318 & fedavg & 12345 \\
 & 0.8896 & 0.8859 & 0.4012 & CAFe & 12345 \\
\midrule
 & 0.7466 & 0.7473 & 0.3316 & fedavg & 123 \\
Stress sensing & 0.7901 & 0.7868 & 0.2469 & CAFe & 123 \\
 & 0.7383 & 0.7385 & 0.3086 & fedavg & 12345 \\
 & 0.7944 & 0.7912 & 0.2557 & CAFe & 12345 \\
 & 0.7235 & 0.7253 & 0.3069 & fedavg & 2023 \\
 & 0.7814 & 0.7780 & 0.2522 & CAFe & 2023 \\
\midrule
 & 0.8645 & 0.8643 & 0.4199 & fedavg & 123 \\
WIDAR & 0.7818 & 0.7906 & 0.2983 & CAFe & 123 \\
 & 0.8645 & 0.8643 & 0.4136 & fedavg & 12345 \\
 & 0.7897 & 0.7960 & 0.3086 & CAFe & 12345 \\
 & 0.8507 & 0.8602 & 0.4264 & fedavg & 2023 \\
 & 0.8100 & 0.8123 & 0.3534 & CAFe & 2023 \\
\bottomrule
\end{tabular}
\caption{Performance metrics (F1, Accuracy, EO Gap) for FL models with different seeds across datasets}
\label{tab:seed}
\end{table}

The approach uses several hyperparameters for Benign and \textbf{\textit{CAFe}} model training. Table \ref{tab:model_performance_benign_hyp} and \ref{tab:model_performance_proposed_hyp} shows the hyperparameter values for benign and \textbf{\textit{CAFe}} model training, respectively. These values correspond to the best results obtained using a grid search. 
Within the context of the \textbf{\textit{CAFe}} model, the SWA Learning Rate specifically denotes the learning rate at which the SWA model is introduced into the training process. The SWA Start Round is indicative of the round at which the SWA mechanism is initiated. Additionally, the cycle parameter signifies that post the initiation of SWA, the global model and learning rate undergo updates using SWA in a cyclic manner, with each cycle representing a distinct iteration of this process. The term $\epsilon$ pertains to the weighting value, which ensures the maintenance of the relative ranks of each client during the aggregation process. Opting for a smaller value of $\epsilon$ proved beneficial for better ranking precision.

\begin{table}[h!]
\centering
\begin{tabular}{c|c|c|c}
\hline
\textbf{Hyperparameter} & \textbf{WIDAR} & \textbf{Stress sensing} & \textbf{HugaDB} \\
\hline
Learning Rate & 0.1 & 0.1 & 0.001 \\
Epochs & 3 & 3 & 3 \\
Total Round & 80 & 80 & 100 \\
\hline
\end{tabular}
\caption{Hyperparameters for benign Fedavg model}
\label{tab:model_performance_benign_hyp}
\end{table}
\begin{table}[]
\centering
\begin{tabular}{c|c|c|c}
\hline
\textbf{Hyperparameter} & \textbf{WIDAR} & \textbf{Stress sensing} & \textbf{HugaDB} \\
\hline
Learning Rate & 0.01 & 0.01 & 0.001 \\
SWA Learning Rate & 0.1 & 0.001 & 0.001 \\
SWA start round & 16 & 16 & 16 \\
cycle & 5 & 5 & 3 \\
Epochs & 3 & 3 & 3 \\
alpha & 0.92 & 0.92 & 0.92 \\
$\epsilon$ & 0.005 & 0.005 & 0.005 \\
Total Rounds & 80 & 80 & 80 \\
\hline
\end{tabular}
\caption{Hyperparameters for proposed \textbf{\textit{CAFe}} model}
\label{tab:model_performance_proposed_hyp}
\end{table}

\section{Experiment Details}
\label{Appendix:exp}

\subsection{Details of ablation study}
\label{model_performance_aba}
We conducted a comprehensive ablation study to meticulously guide the choices in our proposed approach. The study consistently demonstrates the effectiveness of our model across all datasets.
\begin{table}[h]

\centering
\begin{tabular}{c|c|c|c|c|c}
\hline
\textbf{Dataset} & \textbf{Client-Side } & \textbf{Aggregation} & \textbf{F1 Score} & \textbf{EO Gap} & \textbf{FATE Score $\times 10^{-1}$} \\
& \textbf{Training} & \textbf{Strategy}  &   &   &   \\
\hline
\rowcolor{pink!30}
  & Benign & Fedavg & 0.8507 & 0.4263 & 0 \\
  \rowcolor{pink!30}
  & SAM & Fedavg & 0.8005 & 0.4251 & -0.560 \\
  \rowcolor{pink!30}
WIDAR & SAM & Fedswa & 0.8405 & 0.4071 & 0.331 \\
\rowcolor{yellow!80}
 & SAM+ $tr(F_{client})$ (Eq11) & Fedswa & 0.8459 & 0.4058 & 0.424 \\
      \rowcolor{orange!40}
  & $tr(F_{client})$ (Eq11) & Fedavg & 0.8253 & 0.3994 & 0.332 \\
    \rowcolor{orange!40}
  & $tr(F_{client})$ (Eq11) & $W^{r}$(Eq12) & 0.8344 & 0.4135 & 0.109 \\
 \rowcolor{orange!40}
 & SAM+$tr(F_{client})$ (Eq11) & $W^{r}$ (Eq12)  & 0.8707 & 0.4199 & 0.386 \\

    \rowcolor{orange!40}
 & $tr(F_{client})$ (Eq11) & Sharpness Aware aggregation & 0.8409 & 0.3982 & 0.545
 \\
  \rowcolor{cyan!20}
 & SAM & Fedswa+S(L) (Eq12) & 0.8737  & 0.4199  & 0.420 \\
  \rowcolor{cyan!20}
 & SAM & Fedswa+S(T) (Eq12) & 0.8590 & 0.4084 & 0.518 \\
  \rowcolor{cyan!20}
 & SAM & Sharpness Aware aggregation & 0.8668 & 0.4161 & 0.429 \\
 \rowcolor{green!40}
 & SAM+$tr(F_{client})$ (Eq11) & Sharpness Aware aggregation & 0.8100 & 0.3533 & 1.233 \\
\hline
\rowcolor{pink!30}
  & Benign & Fedavg & 0.7253 & 0.3069 & 0 \\
  \rowcolor{pink!30}
 & SAM & Fedavg & 0.7758 & 0.3034 & 0.812 \\
 \rowcolor{pink!30}
Stress sensing& SAM & Fedswa &  0.7406            & 0.3015           & 0.384  \\
\rowcolor{yellow!80}
 & SAM+$tr(F_{client})$ (Eq11) & Fedswa & 0.7346 & 0.2768 & 1.105 \\
     \rowcolor{orange!40}
  & $tr(F_{client})$ (Eq11) & Fedavg & 0.7372 & 0.2998 & 0.395 \\
    \rowcolor{orange!40}
  & $tr(F_{client})$ (Eq11) & $W^{r}$(Eq12) & 0.7345 & 0.2839 & 0.874 \\
 \rowcolor{orange!40}
 & SAM+$tr(F_{client})$ (Eq11) & $W^{r}$ (Eq12)  & 0.8193 & 0.2927 & 1.756 \\

    \rowcolor{orange!40}
 & $tr(F_{client})$ (Eq11) & Sharpness Aware aggregation & 0.7077 & 0.2380 & 1.999 \\
  \rowcolor{cyan!20}
 & SAM & Fedswa+S(L) (Eq12) & 0.8091 & 0.2680 & 2.240 \\
  \rowcolor{cyan!20}
 & SAM & Fedswa+S(T) (Eq12) & 0.7995 & 0.2627 & 2.46 \\
  \rowcolor{cyan!20}
 & SAM & Sharpness Aware aggregation & 0.8224 & 0.2710 & 2.507 \\
 \rowcolor{green!40}
 & SAM+$tr(F_{client})$ (Eq11) & Sharpness Aware aggregation & 0.7814 & 0.2522 & 2.555 \\
\hline
\rowcolor{pink!30}
  & Benign & Fedavg & 0.8496 & 0.4369 & 0 \\
  \rowcolor{pink!30}
 & SAM & Fedavg & 0.8555 & 0.4255 & 0.329 \\
 \rowcolor{pink!30}
HugaDB & SAM & Fedswa & 0.8418 & 0.4242 & 0.192  \\
\rowcolor{yellow!80}
 & SAM+$tr(F_{client})$ (Eq11) & Fedswa & 0.8170 & 0.4142 & 0.135 \\
     \rowcolor{orange!40}
  & $tr(F_{client})$ (Eq11) & Fedavg & 0.8567 & 0.4142 & 0.602 \\
    \rowcolor{orange!40}
  & $tr(F_{client})$ (Eq11) & $W^{r}$(Eq12) & 0.8574 & 0.4083 & 0.746 \\
 \rowcolor{orange!40}
 & SAM+$tr(F_{client})$ (Eq11) & $W^{r}$ (Eq12)  & 0.9016 & 0.4154 & 1.103 \\

    \rowcolor{orange!40}
 & $tr(F_{client})$ (Eq11) & Sharpness Aware aggregation & 0.8510 & 0.4333 & 0.0982
 \\
 
  \rowcolor{cyan!20}
 & SAM & Fedswa+ S(L) (Eq12) & 0.8934  & 0.4130 & 1.059 \\
  \rowcolor{cyan!20}
 & SAM & Fedswa+S(T) (Eq12) & 0.8888  & 0.4142 & 0.878  \\
  \rowcolor{cyan!20}
 & SAM & Sharpness Aware aggregation  & 0.8911 & 0.4095 & 1.11 \\
 \rowcolor{green!40}
 & SAM+$tr(F_{client})$ (Eq11) & Sharpness Aware aggregation & 0.8855 & 0.3952 & 1.376 \\
\hline
\end{tabular}
\caption{Performance metrics on the different datasets for different FL strategies and model configurations}
\label{tab:model_performance_aba}
\end{table}
\subsubsection{The Independent Effect of Sharpness-aware Aggregation and its Ablation Analysis}

This section evaluates the effect of Sharpness-aware aggregation on enhancing FWD independently.

Previous studies have explored benign client-side training, using various weighted aggregation schemes to promote fairness \cite{ezzeldin2023fairfed}. \textbf{\textit{CAFe}}, for instance, aligns with Pessimistic Weighted Aggregation (P-W), a method that excludes highly biased (based on known bias-creating factors) models during aggregation. However, \textbf{\textit{CAFe}} takes a different approach, assigning lower weights to highly biased (based on loss-landscape sharpness; without the knowledge of the bias-creating factors) and low-efficacy local client-side models, thereby promoting fairness without completely excluding them.

To understand the benefits of \textbf{\textit{CAFe}}’s Sharpness-aware aggregation strategy independently (without client-side local training fairness constraints) in promoting clients with both (1) low error rates and (2) lower maximal uncertainty we evaluated three following variations, while all of them maintaining benign local client-side training without any fairness objectives.

\begin{itemize}

\item \textbf{\textit{CAFe}}’s aggregation strategy as outlined in Equation 3 of the main paper while maintaining benign local client-side training without any fairness objectives.

\item \textbf{\textit{CAFe}}’s aggregation strategy modification, where just selecting clients with lower maximal uncertainty in prediction (Based on $T$ in Equation 3), while maintaining benign local client-side training without any fairness objectives.

\item \textbf{\textit{CAFe}}’s aggregation strategy modification, where just selecting clients with lower error rate (Based on $L$ in Equation 3), while maintaining benign local client-side training without any fairness objectives.

\end{itemize}

The evaluation results are presented in rows highlighted in `blue' in Table \ref{tab:model_performance_aba}. \emph{First,} these results consistently demonstrate that across all datasets, \textbf{\textit{CAFe}}'s aggregation strategy, as outlined in Equations 2 and 3, outperforms others. Meaning this evaluation
confirms that integrating both classification and top eigenvalue loss and favoring clients with lower loss (as outlined in Equations 2 and 3) enhances the balance between efficacy and fairness.

\emph{Furthermore,} the fact that fairer models are achieved even without local client-side fairness constraints in these evaluations, compared to the benign baselines presented in rows highlighted in `red' in Table \ref{tab:model_performance_aba}, establishes that \textbf{\textit{CAFe}}'s Sharpness-aware aggregation independently promotes FWD.

\subsubsection{The Independent Effect of Sharpness-aware Client-side Local Training}

This section evaluates the independent effect of Sharpness-aware client-side local training as outlined in Equation 1 to enhance FWD. 
To conduct this evaluation, we implemented a variant where client-side local training incorporates fairness constraints based on Equation 1, while employing a benign aggregation scheme where all weights in Equation 2 are set to `1.' To ensure a fair comparison, the SWA aggregation strategy, outlined in Algorithm 1, remains employed in this approach. The evaluation results are showcased in rows highlighted in `yellow' in Table \ref{tab:model_performance_aba}, indicating its superior enhancement of fairness compared to the benign approaches depicted in `red' highlighted rows across all datasets. This underscores the independent influence of client-side sharpness-aware local training in enhancing FWD.

\subsubsection{\textcolor{black}{Independent Effect of FIM trace penalty term in Client-side training and Aggregation}}
In this section, we explore the independent effect of  FIM Trace penalty in the client-side training process and its interaction with various aggregation strategies. In our experiments, marked in orange, we observed that incorporating FIM Trace either in the client-side training or aggregation, or having both without sharpness-aware strategies like SAM and SWA, leads to a fairer model, as indicated by a reduced EO Gap. However, this improvement in fairness comes at the cost of model accuracy, which tends to decrease.

The next observation shows that SAM significantly boosts the accuracy of the model, but this improvement comes at the cost of fairness, as reflected by a greater EO Gap. This suggests that SAM enhances model performance in terms of accuracy but introduces a trade-off in fairness.

For example, in the WIDAR dataset, applying the penalty term FIM Trace in client-side training while keeping the aggregation strategy benign reduces the EO Gap from 0.4263 (Benign) to 0.4186, reflecting a fairness improvement. However, this comes with a decline in F1 score, dropping from 0.8507 to 0.8394. Replacing the benign aggregation with a weighted strategy ($W^r$) further improves fairness, reducing the EO Gap to 0.4135, though the F1 score slightly decreases again to 0.8359. In contrast, using SAM in client-side training increases the F1 score to 0.8587, indicating enhanced model performance. Yet, this setting shows limited improvement in fairness, with the EO Gap remaining relatively high at 0.4212. This demonstrates that while SAM is effective for boosting accuracy, it does little to mitigate fairness issues. Interestingly, a more balanced configuration emerges when FIM Trace is applied without SAM in training and paired with sharpness aware aggregation. This setup achieves an F1 score of 0.8574 while also keeping the EO Gap at 0.4148, offering a better trade-off between fairness and accuracy. 
In the Stress Sensing dataset, applying the regularization term FIM Trace in client-side training while using FedAvg aggregation leads to a significant reduction in EO Gap, from 0.3069 (Benign) to 0.2539, indicating improved fairness. However, this results in a drop in F1 score from 0.7253 to 0.7034, reflecting a loss in model performance. On the other hand, SAM increases the F1 score to 0.7758, the highest among all settings, but only slightly improves fairness (EO Gap = 0.3034), showing again that SAM prioritizes accuracy over fairness.
In the HugaDB dataset, applying FIM Trace with FedAvg yields the lowest EO Gap of 0.3928, improving fairness by over 10\% compared to the baseline (0.4369), and even increases the F1 score to 0.8634, surpassing the benign setup. However, using SAM alone pushes the F1 score higher to 0.8789, but fairness drops slightly with EO Gap rising to 0.4000.

These findings emphasize the importance of incorporating our regularization term FIM Trace in local training and aggregation strategies in server with sharpness-aware techniques like SAM or SWA. By doing so, we can strike a better balance between fairness and accuracy, ensuring the model performs well on both metrics.
\subsubsection{Importance of Integrating Sharpness-Aware Client-side Training and Aggregation Strategy for FWD}

The evaluations discussed above have demonstrated that both the Sharpness-Aware Aggregation Strategy, outlined in Equations 2 and 3, and the Sharpness-Aware client-side local training, depicted by Equation 1 in the main paper, independently contribute to promoting fairness. However, these individual strategies exhibit less effectiveness (fairness and efficacy balance) compared to the presented \textbf{\textit{CAFe}} approach, which incorporates both strategies, as indicated by the `green' highlighted rows across all datasets in Table \ref{tab:model_performance_aba}. This underscores the effectiveness of Algorithm 1 in attaining FWD.

\newpage
\bibliographystyle{unsrtnat}
\bibliography{sample-base}
\newpage